\documentclass[11pt]{article}
\usepackage{coling2018}
\usepackage{times}
\usepackage{url}
\usepackage{latexsym}
\usepackage{amsmath}
\usepackage{amsfonts}
\usepackage{mathrsfs}
\usepackage{graphicx}
\usepackage{subcaption}
\usepackage{booktabs}
\usepackage{threeparttable}
\usepackage{diagbox}
\usepackage{booktabs}
\usepackage{amsmath,amssymb}
\usepackage{algorithm}
\usepackage{algpseudocode}
\usepackage[hidelinks]{hyperref}
\usepackage{url}

\usepackage{algorithm}
\usepackage{algorithmicx}
\usepackage{algpseudocode}
\usepackage{listings}
\usepackage{xcolor}
\usepackage{graphicx}
\usepackage{array}
\usepackage{tabularx}
\usepackage[most]{tcolorbox}
\usepackage{enumitem}
\usepackage{xcolor}

\definecolor{CriticBlue}{HTML}{2F5D8C}
\definecolor{CriticBackground}{HTML}{F4F7FA}

\usepackage{graphicx}

\title{Moonworks Lunara: Modeling Artistic Intelligence}

\author{ Yan Wang, Yanzu Wang, Maitreyee Joshi, Samiha Sadeka, \\\textbf{Partho Hassan, Reza Jarral, Sayeef Abdullah, Sabit Hassan}\\
  {\tt research@moonworks.ai} \\}

\date{}

\begin{document}

\maketitle
\vspace{-1cm}
\begin{figure}[!h]
    \centering
    \setlength{\tabcolsep}{2pt}
    \begin{tabular}{cccccc}
        \includegraphics[width=0.16\textwidth]{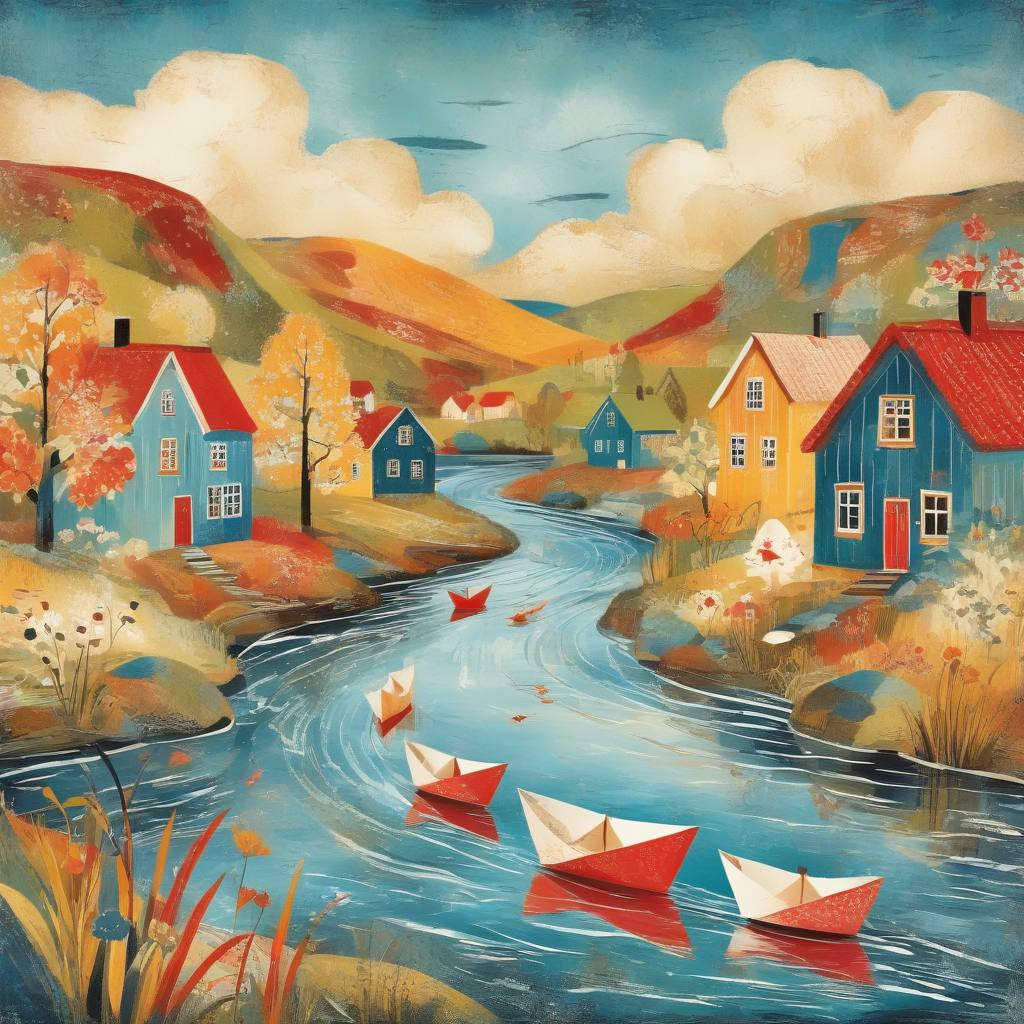} &
                \includegraphics[width=0.16\textwidth]{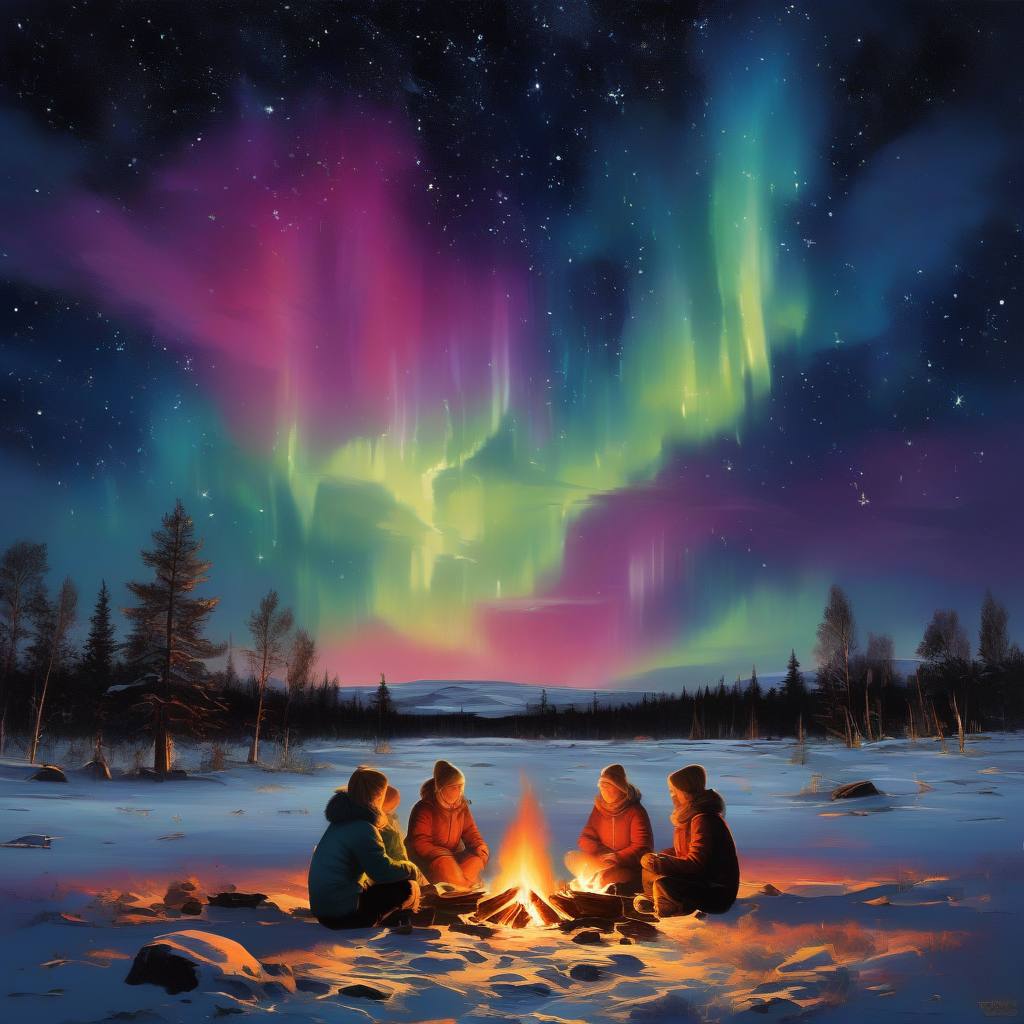} &

        \includegraphics[width=0.26\textwidth]{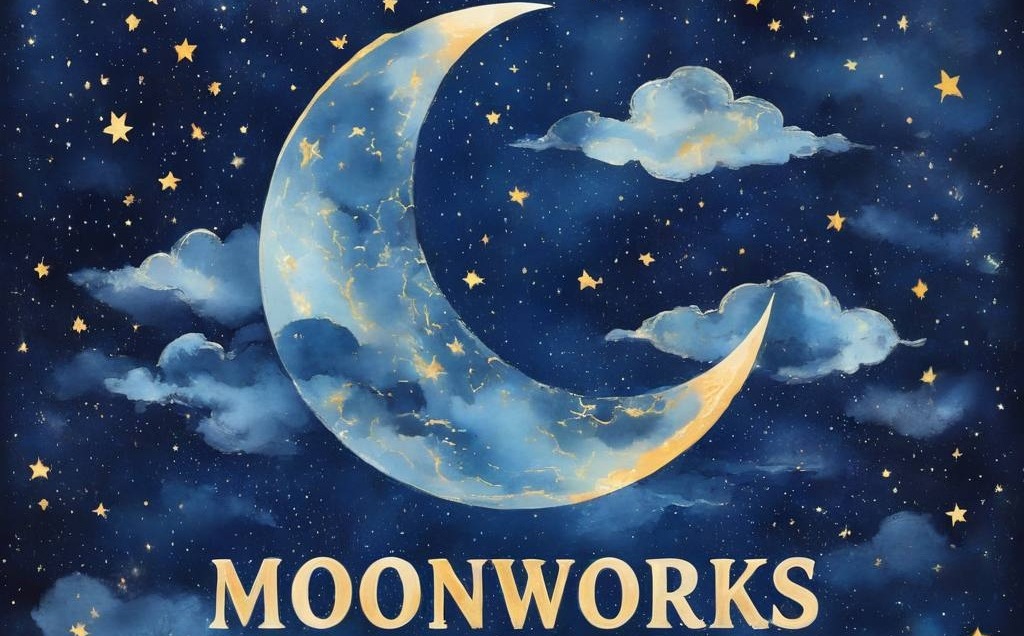}
                \includegraphics[width=0.16\textwidth]{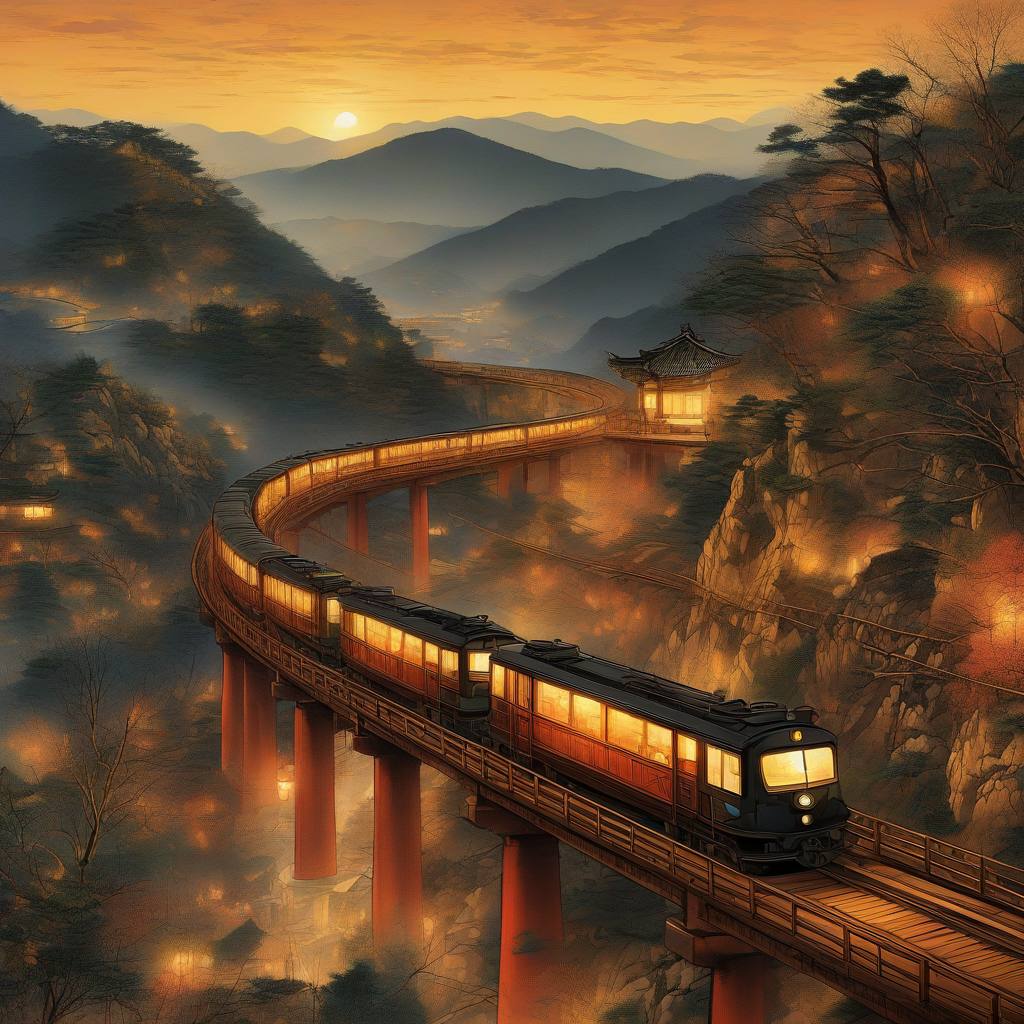} &

        \includegraphics[width=0.16\textwidth]{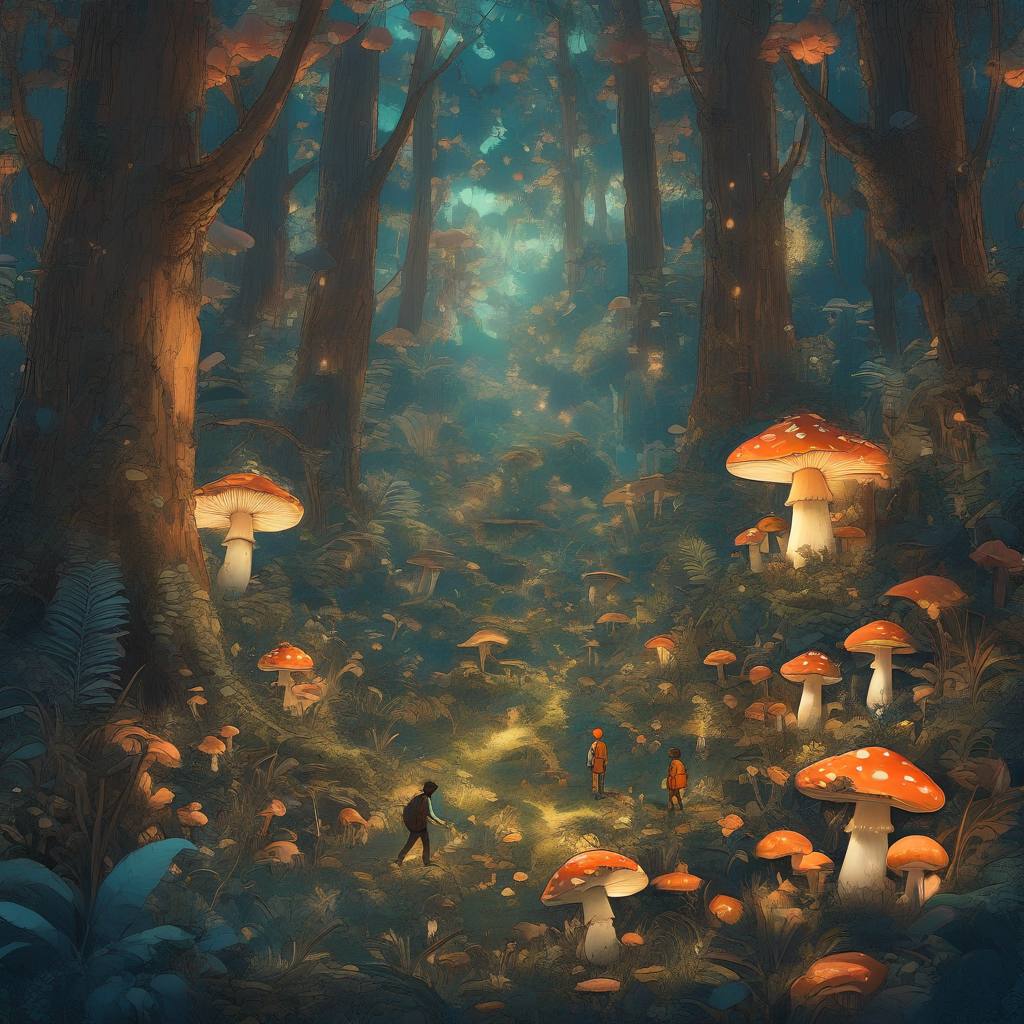}
    \end{tabular}
    \label{fig:style-banner}
\end{figure}
\begin{abstract}

We formulate \emph{Artistic Intelligence} as exploration driven world realization, leaving space for creative possibility while preserving the semantic, artistic, and compositional structure that must remain true. Moonworks Lunara, a text-to-image model, implements this framework with a novel Diffusion Mixture Transformer architecture. A new training algorithm iteratively evolves the data distribution through informative sample acquisition and targeted injection of human-created art. We benchmark Lunara against seven image-generation models, including FLUX.2-Klein-4B, Qwen-Image (20B), and GPT-Image-1-Mini. With GPT-5.6 Sol as evaluator, Lunara ranks first in \emph{Aesthetic Quality (8.473 vs. 8.457 GPT-Image-1-mini)}, second in \emph{Emotional Resonance}, and remains competitive in \emph{Content Integrity}. A blind human evaluation over the same evaluation set corroborates the automated metrics, ranking Lunara first. It also stays among the strongest models under conventional measures including CLIPScore and LAION Aesthetic Predictor. On GenEval, Lunara achieves competitive performance against a broader set of 16 models, including GPT Image 2 and Seedream 4.0. These results place Lunara at the frontier with Artistic Intelligence while maintaining a sub-10B active-parameter footprint and sub-10-second inference latency. Lunara advances the general visual intelligence frontier by shifting the question from whether models can get images right to how deeply they can interpret meaning and realize it as imaginative, expressive worlds.

\end{abstract}
\begin{figure*}[t]
\centering
\includegraphics[
    width=0.95\textwidth]{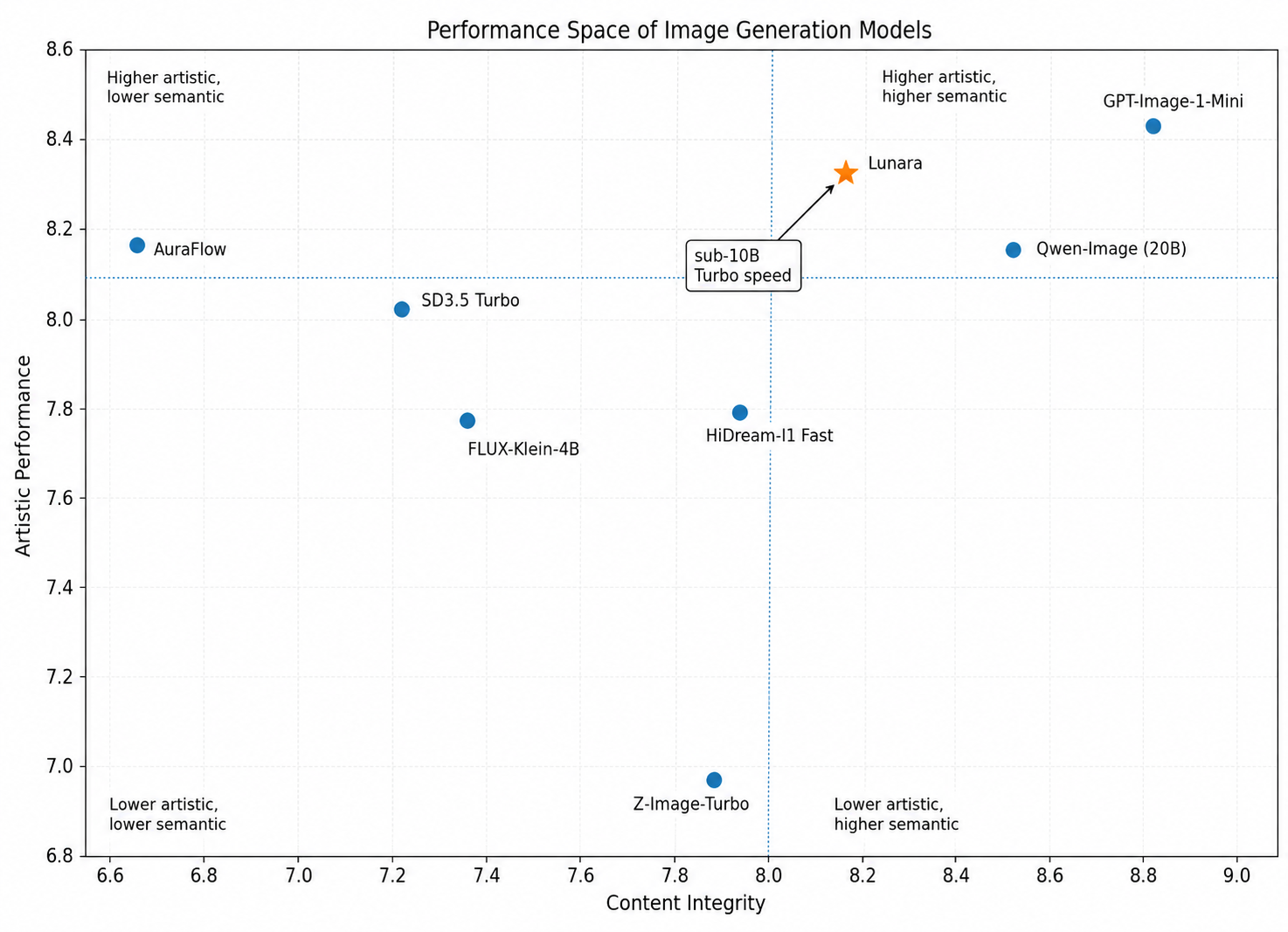}

\caption{Performance landscape across artistic performance and content integrity. Lunara occupies the artistic frontier, leading in aesthetic quality and emotional resonance while maintaining competitive content integrity at sub-10B active parameters and fast inference.}
\label{fig:Baseline_performance}
\end{figure*}

\begin{figure*}[]
    \centering
    \setlength{\tabcolsep}{2pt}
    \renewcommand{\arraystretch}{0}

    \begin{tabular}{cccc}
    \includegraphics[width=0.24\textwidth]{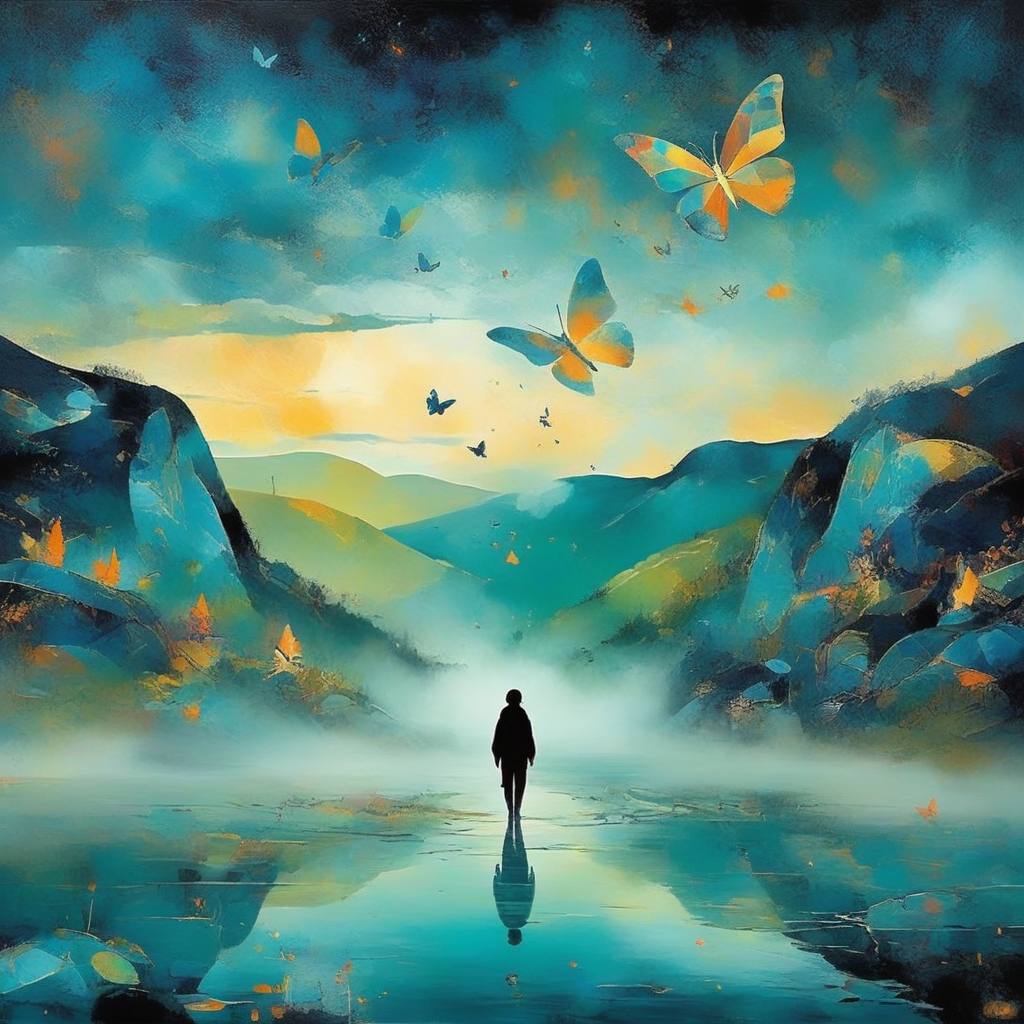} &
    \includegraphics[width=0.24\textwidth]{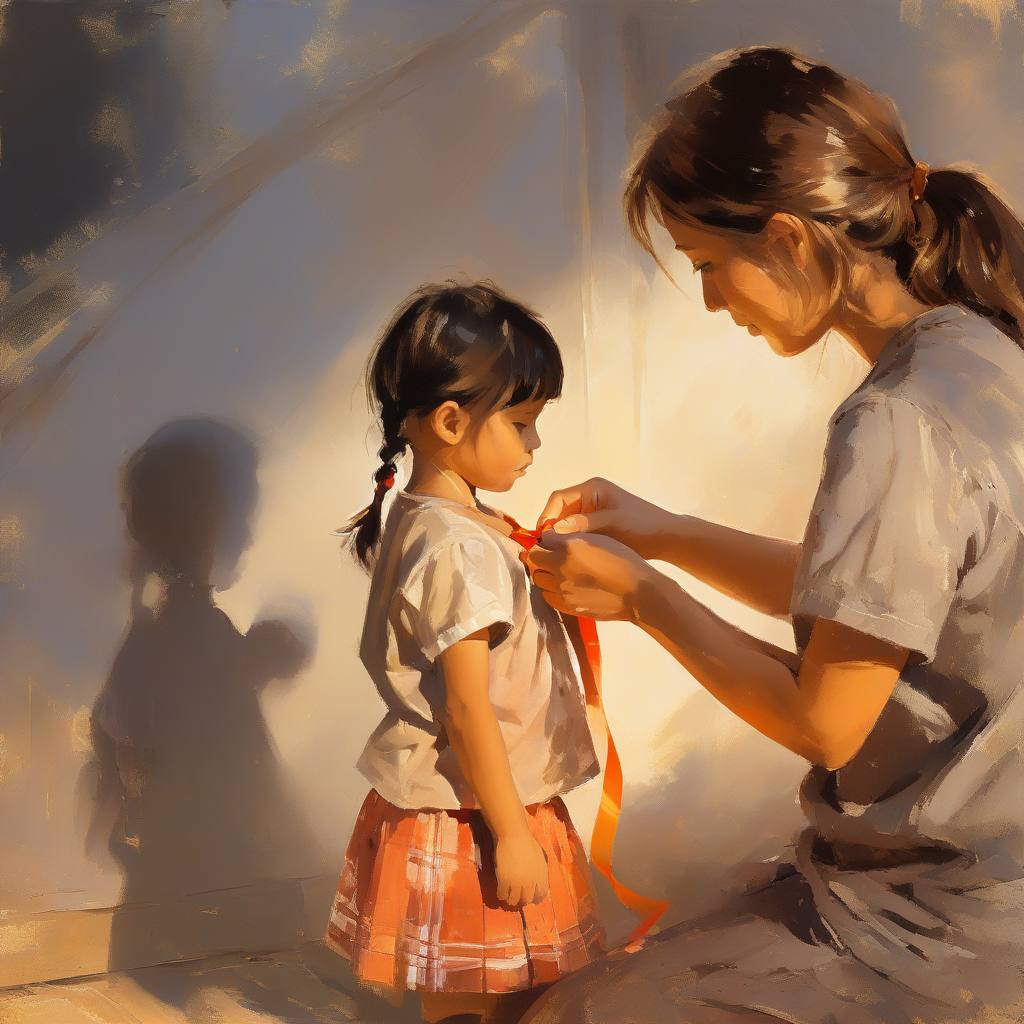} &
    \includegraphics[width=0.24\textwidth]{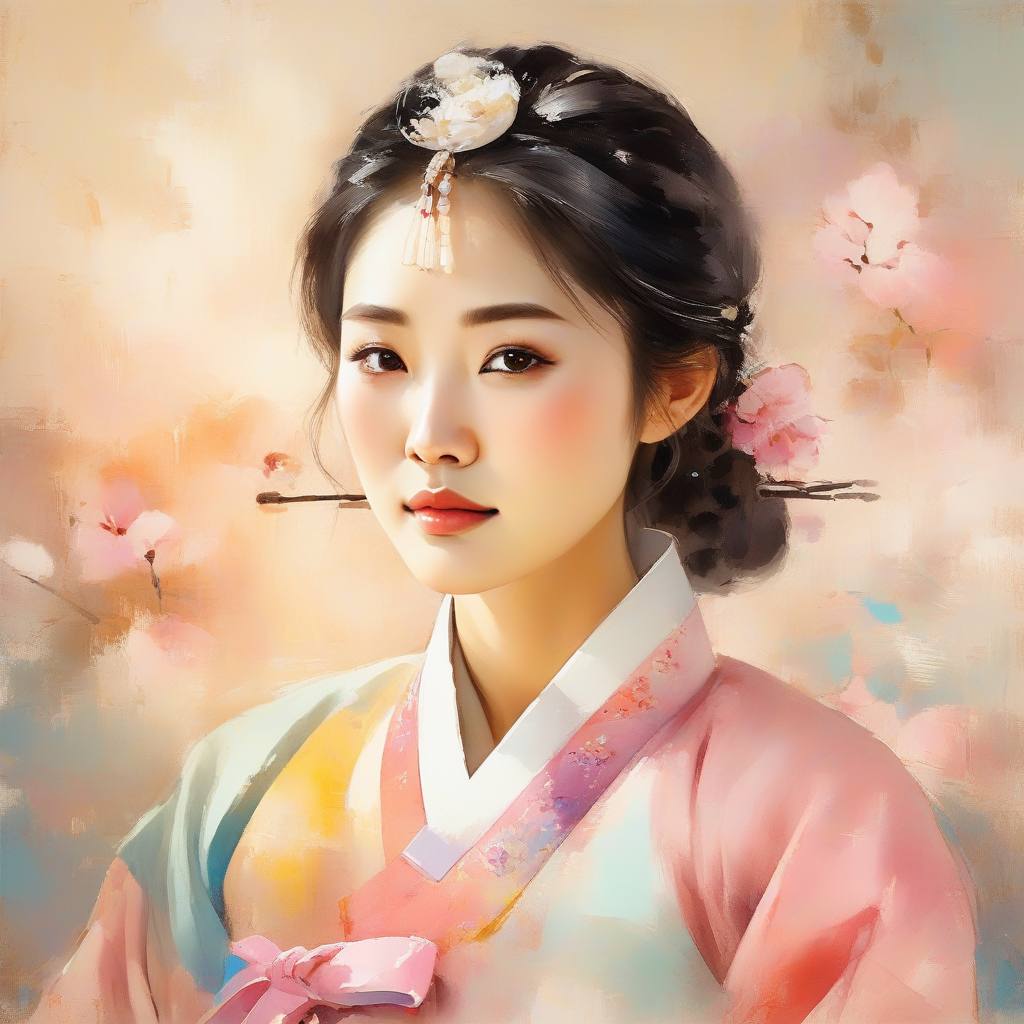} &
    \includegraphics[width=0.24\textwidth]{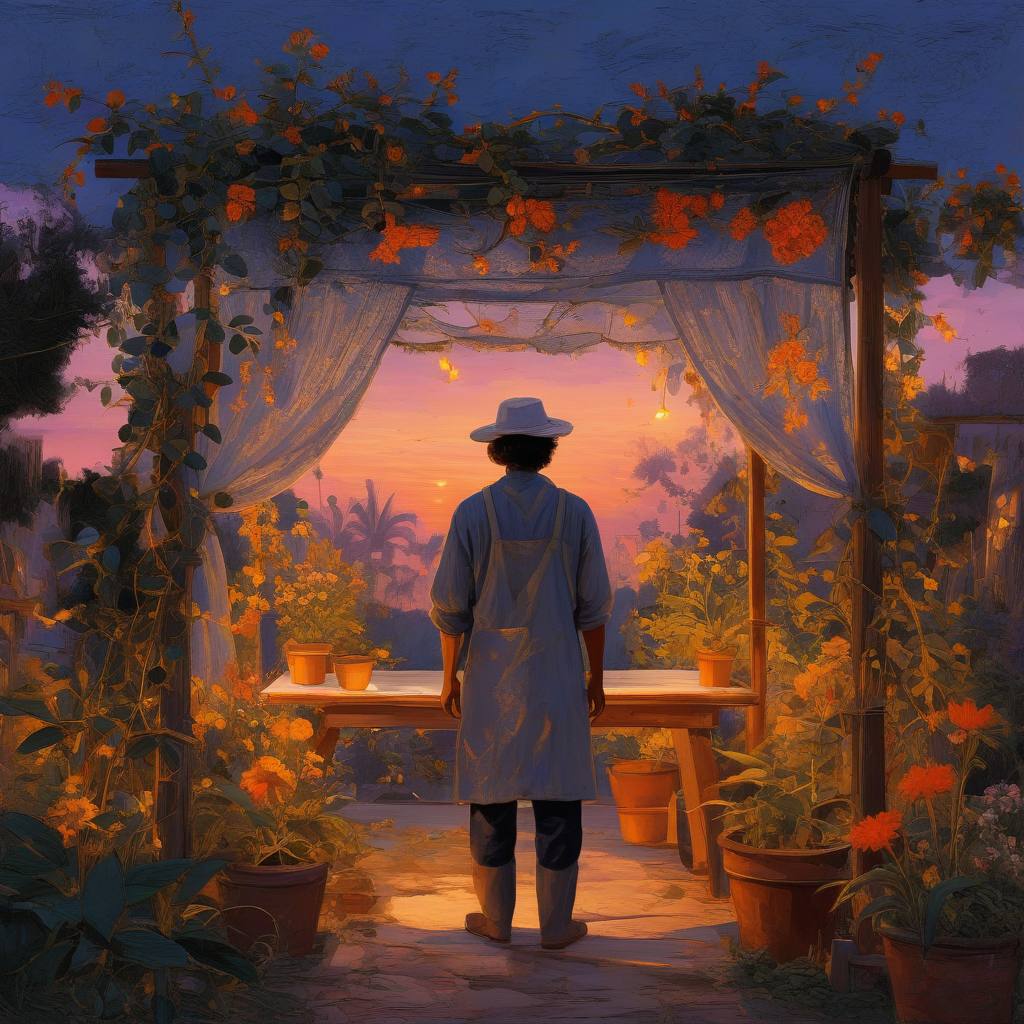} \\[4pt]

    \includegraphics[width=0.24\textwidth]{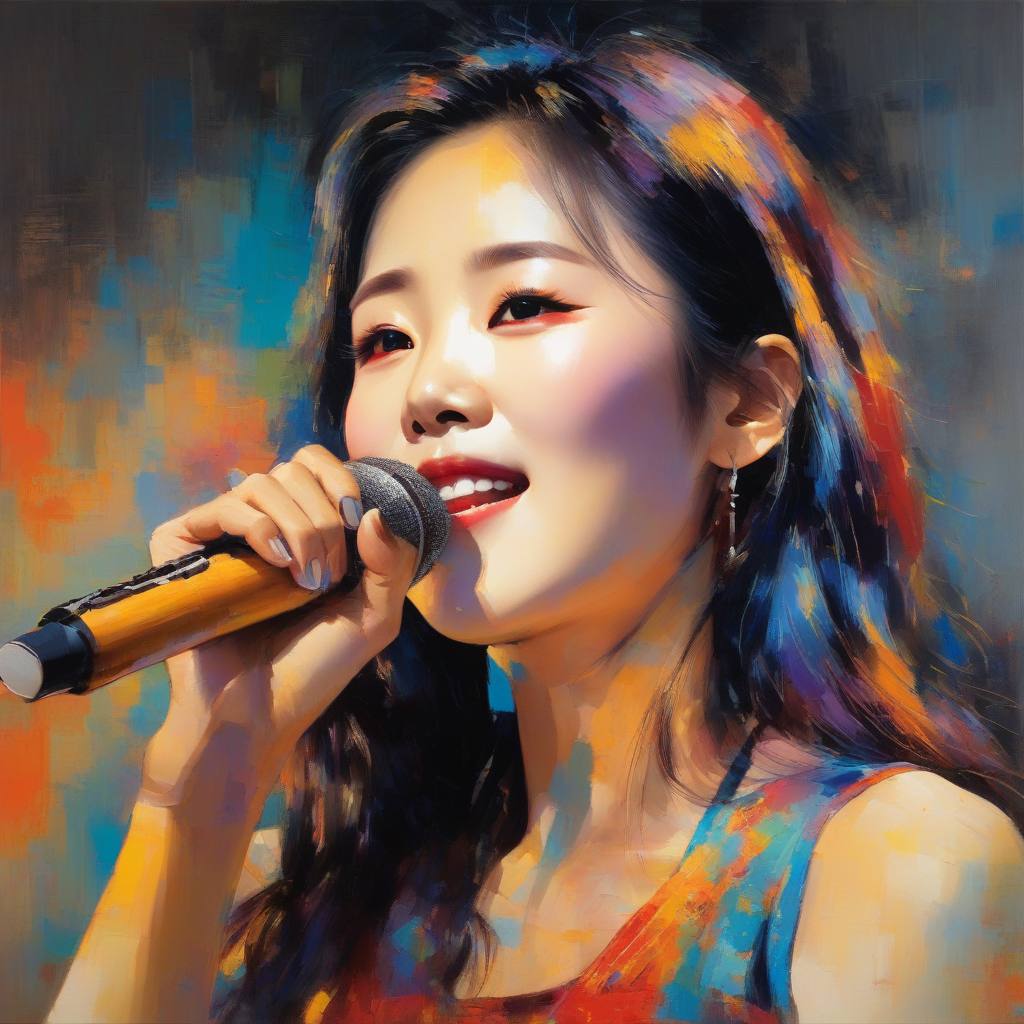} &
    \includegraphics[width=0.24\textwidth]{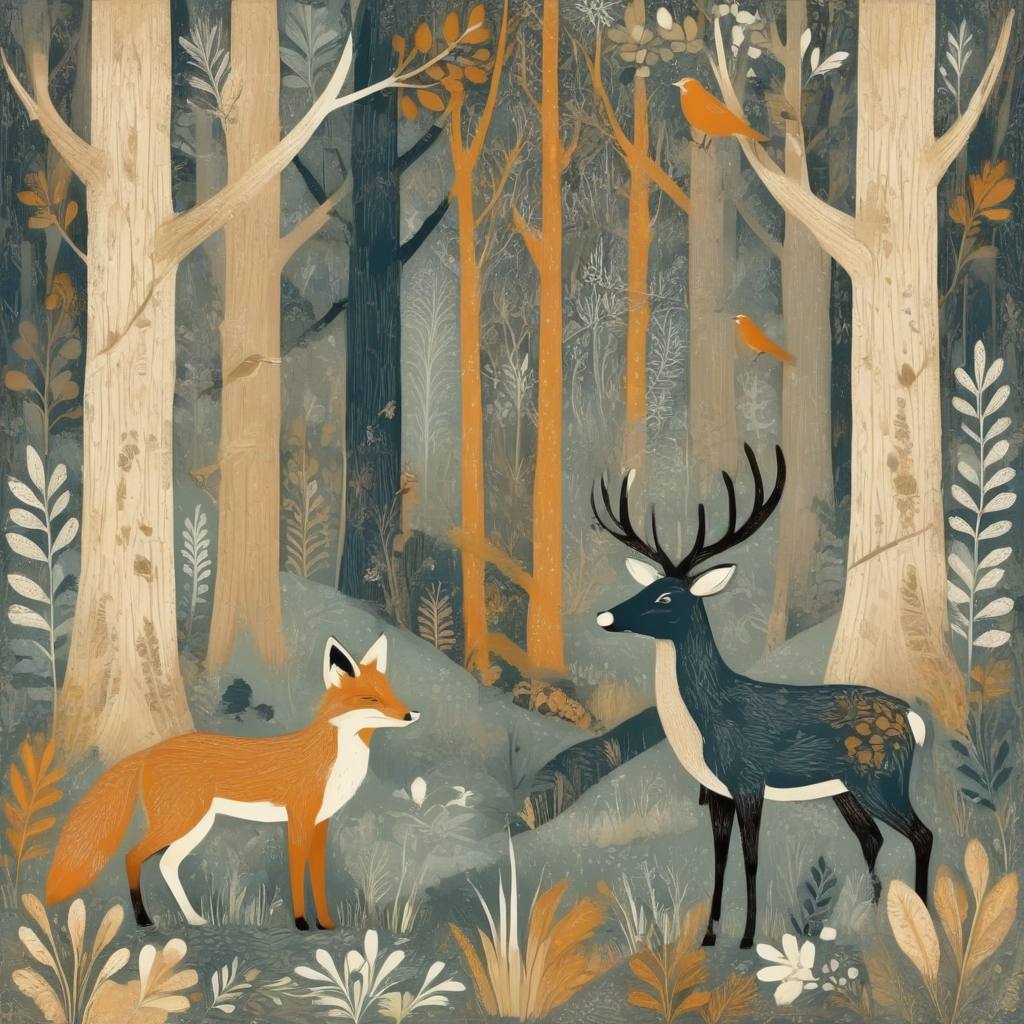} &
    \includegraphics[width=0.24\textwidth]{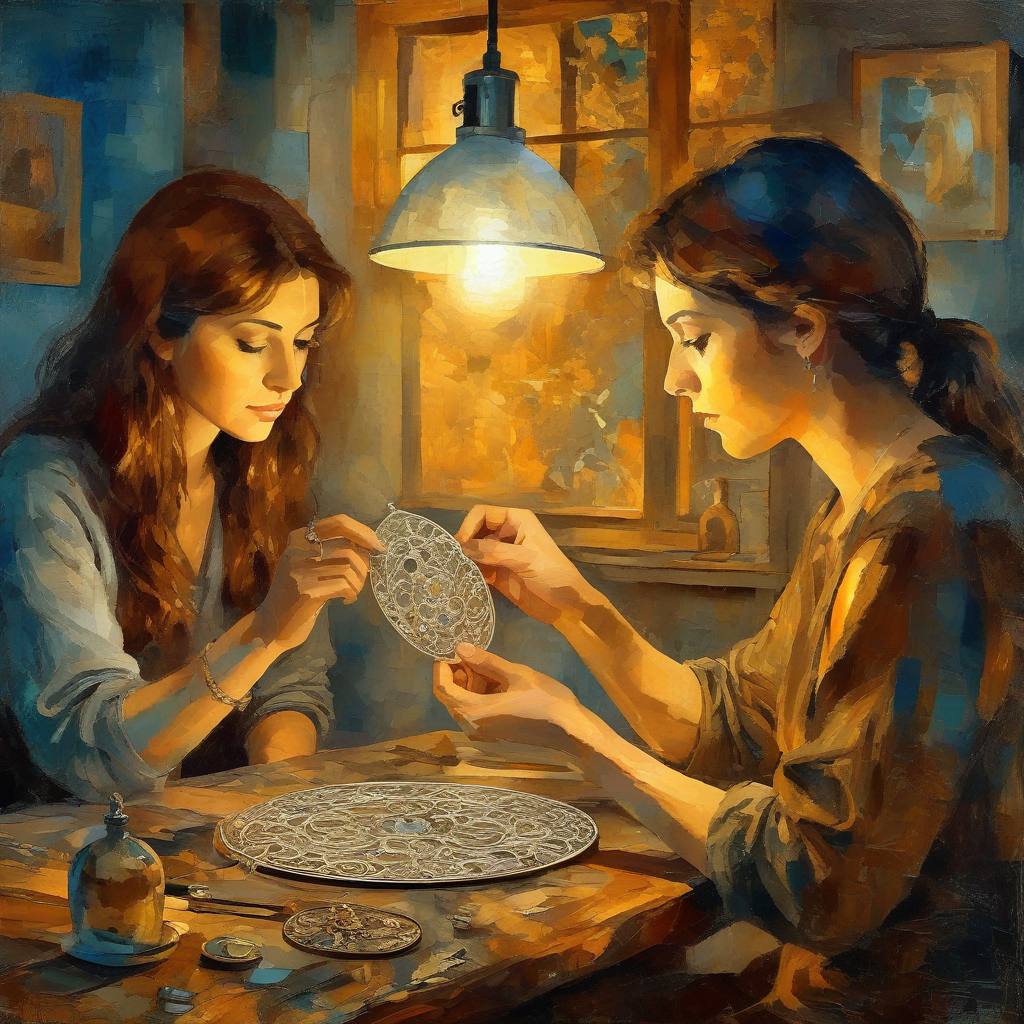} &
    \includegraphics[width=0.24\textwidth]{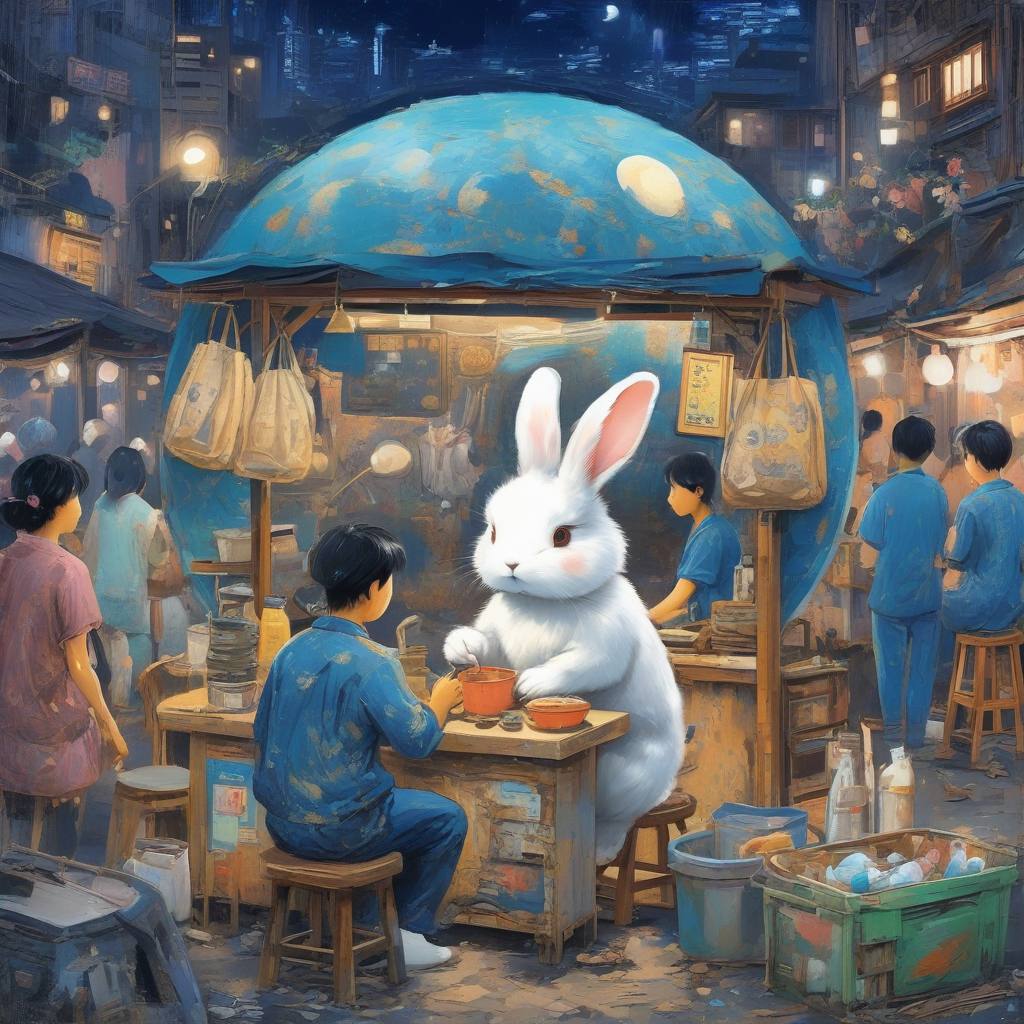} \\[4pt]

    \includegraphics[width=0.24\textwidth]{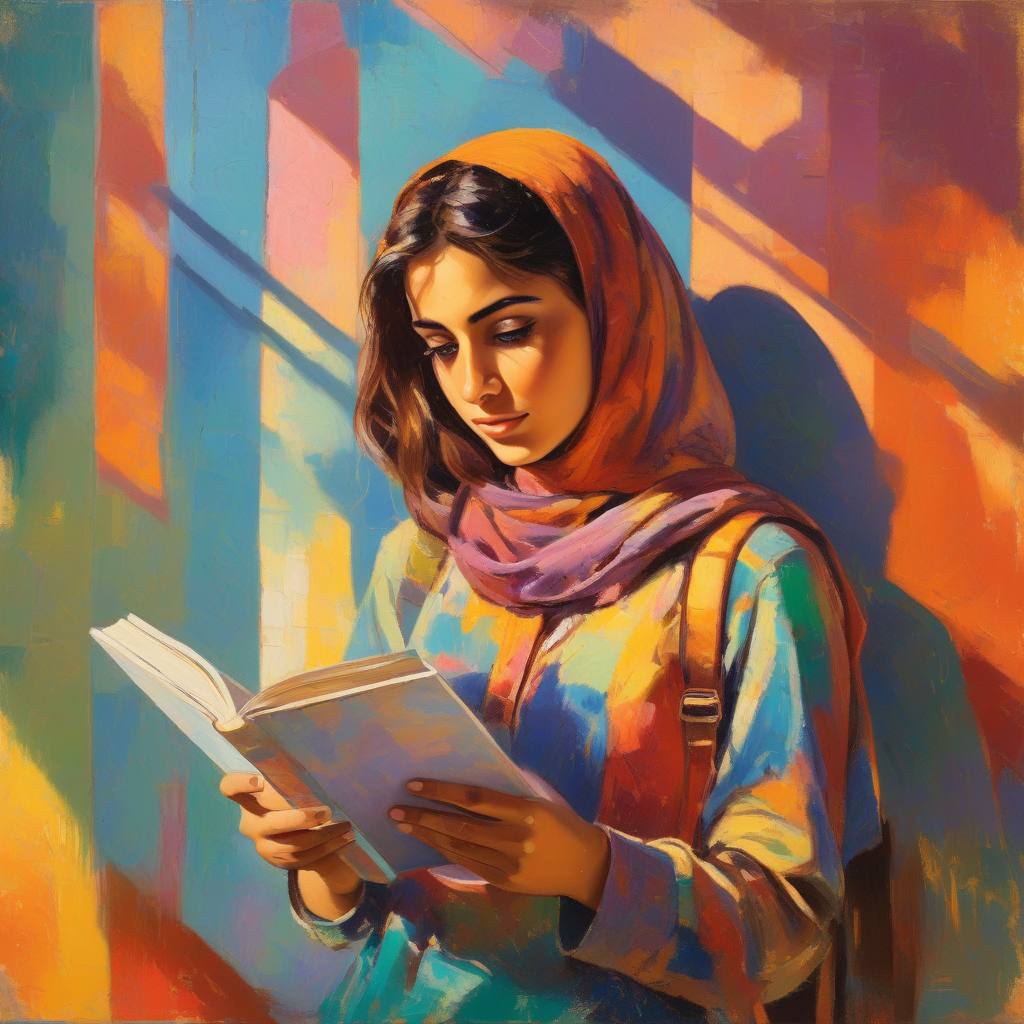} &
    \includegraphics[width=0.24\textwidth]{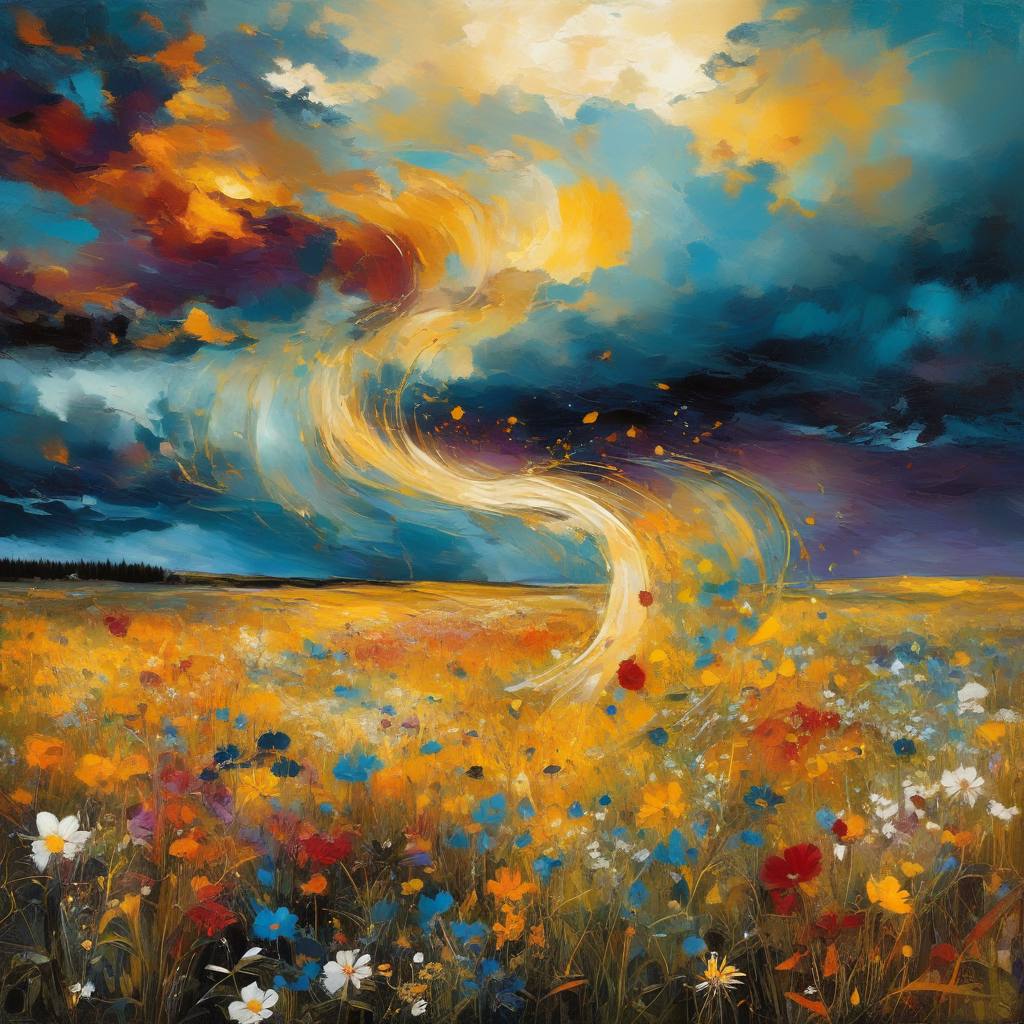} &
    \includegraphics[width=0.24\textwidth]{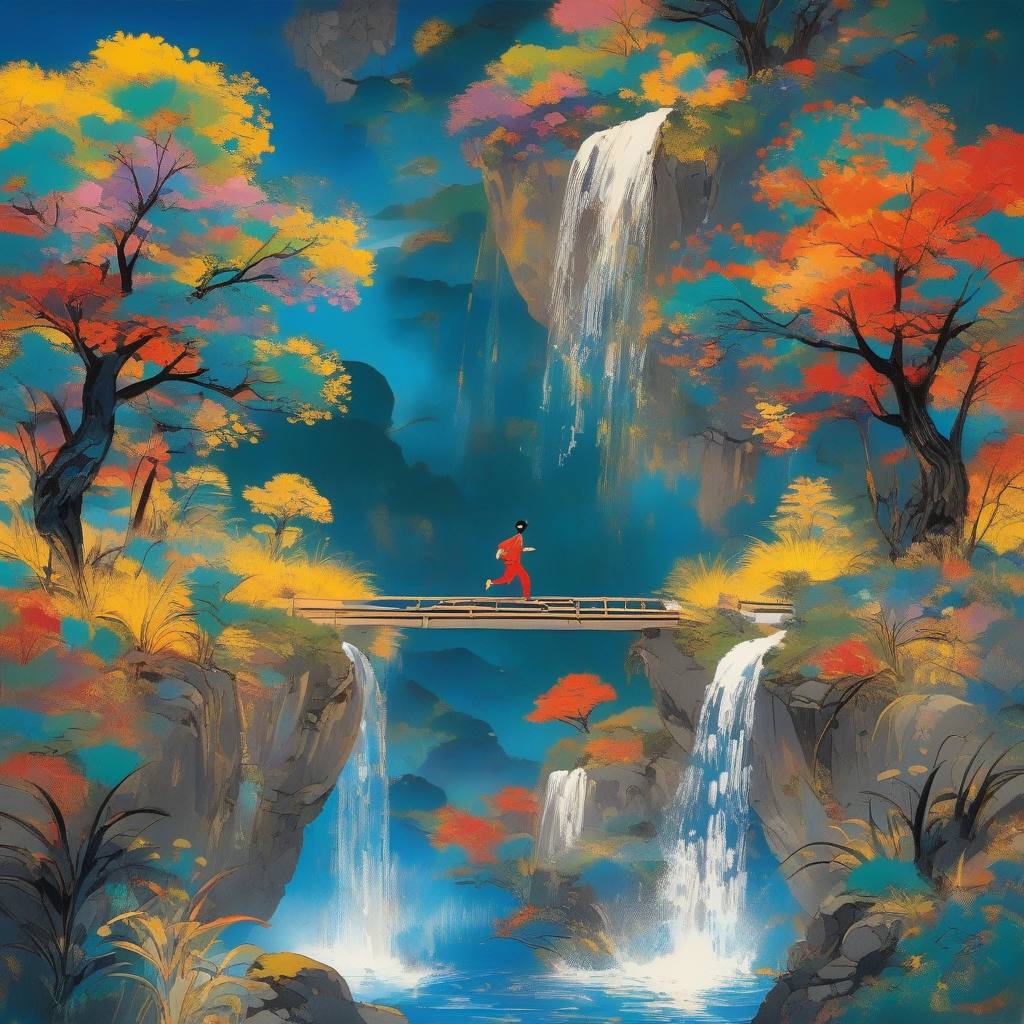} &
    \includegraphics[width=0.24\textwidth]{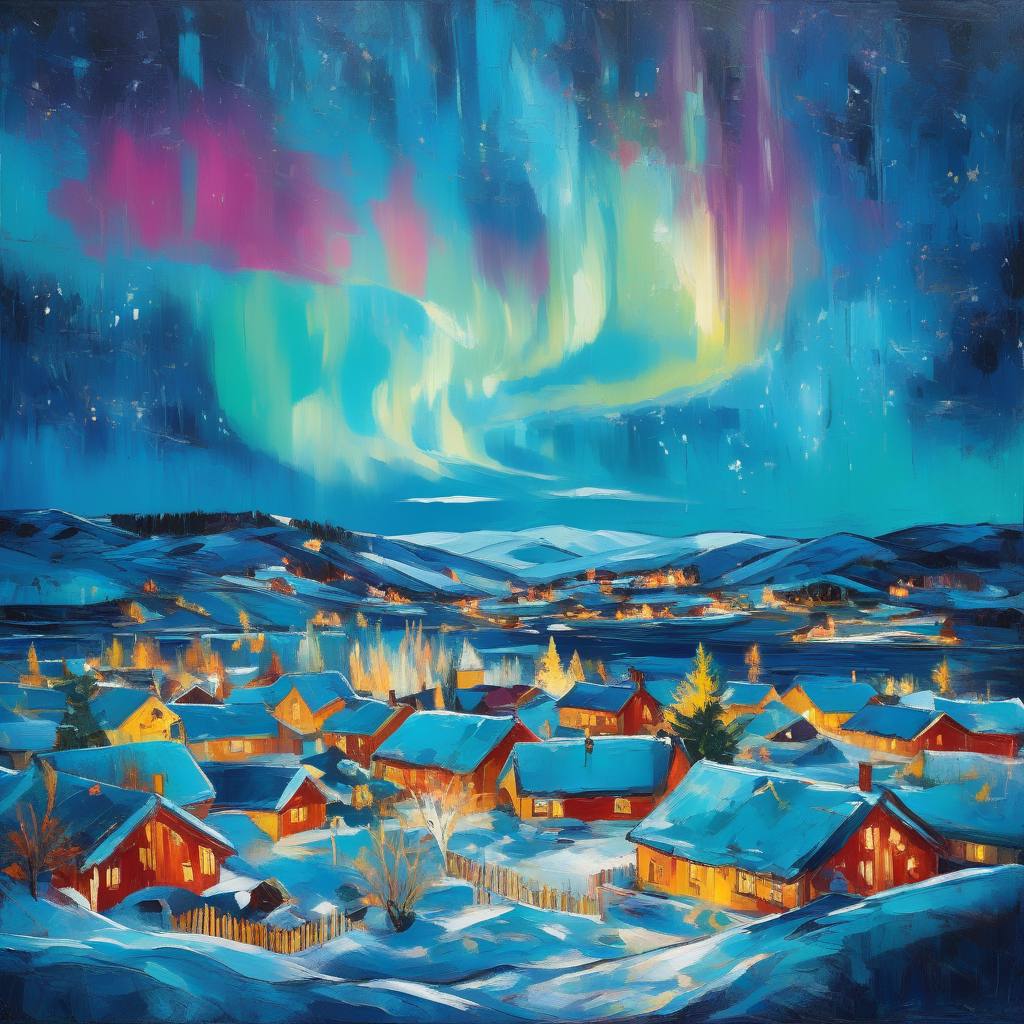} \\[4pt]

    \includegraphics[width=0.24\textwidth]{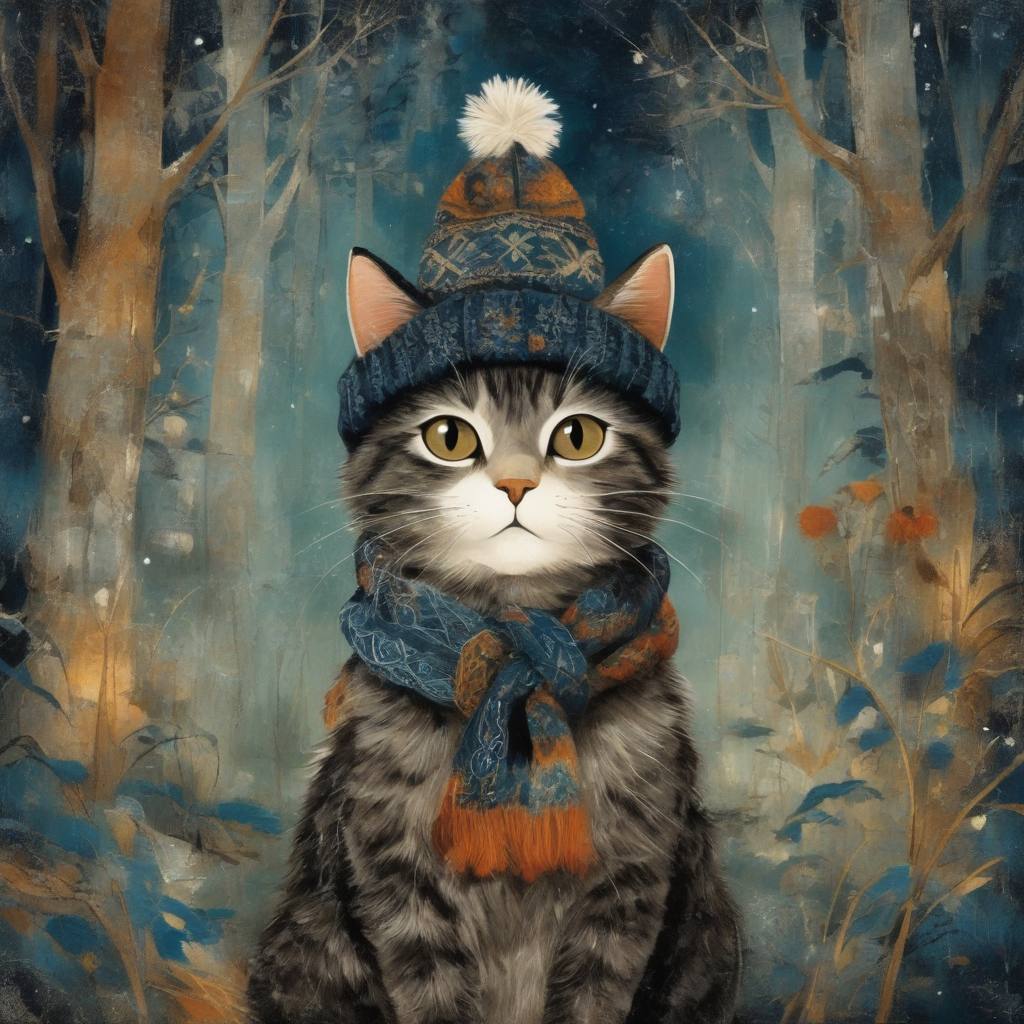} &
    \includegraphics[width=0.24\textwidth]{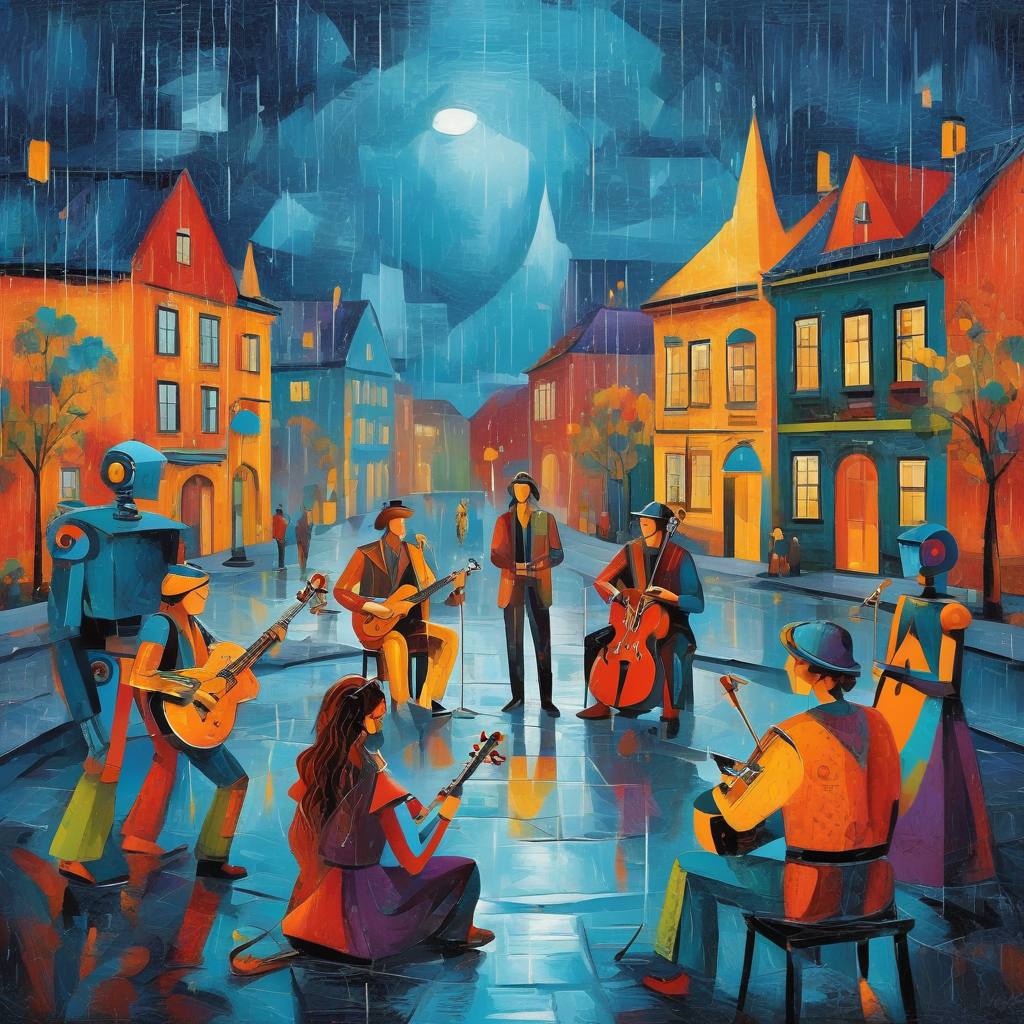} &
    \includegraphics[width=0.24\textwidth]{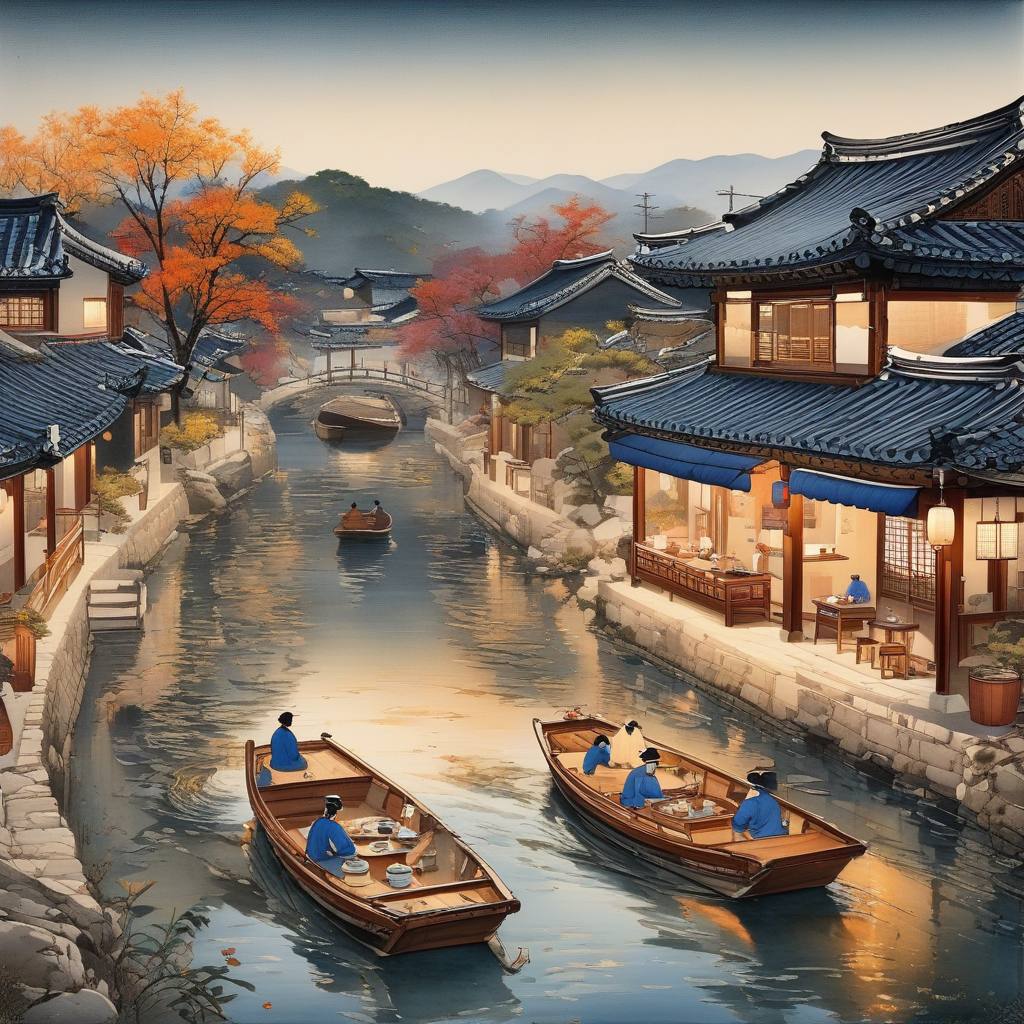} &
    \includegraphics[width=0.24\textwidth]{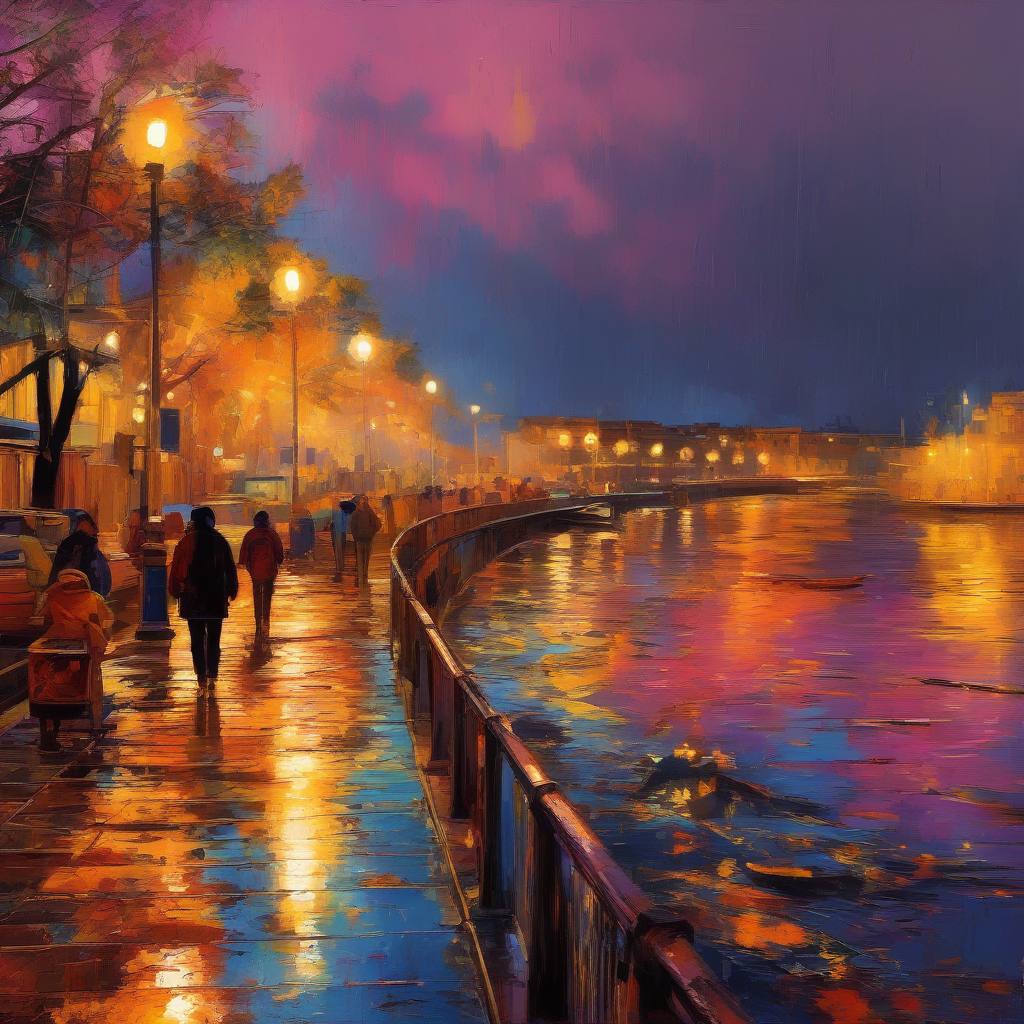} \\[4pt]

    \includegraphics[width=0.24\textwidth]{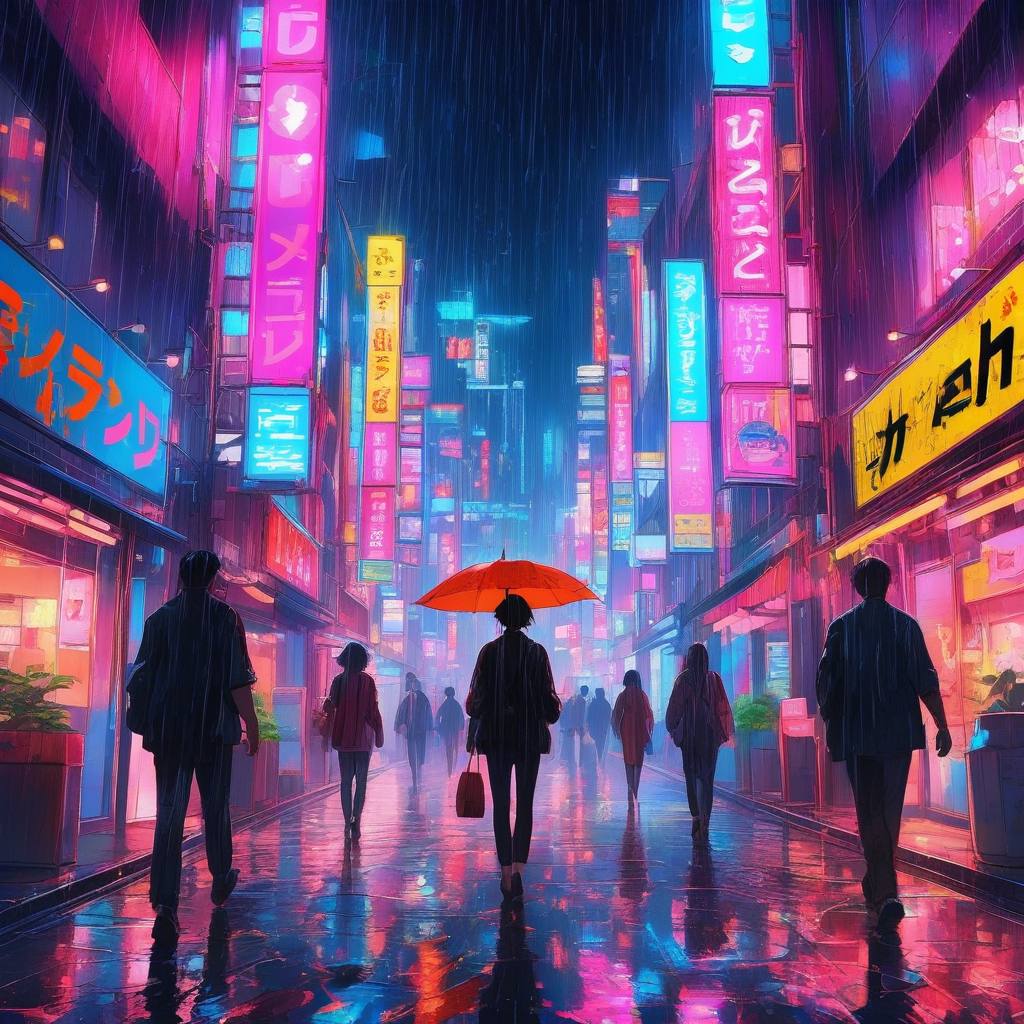} &
    \includegraphics[width=0.24\textwidth]{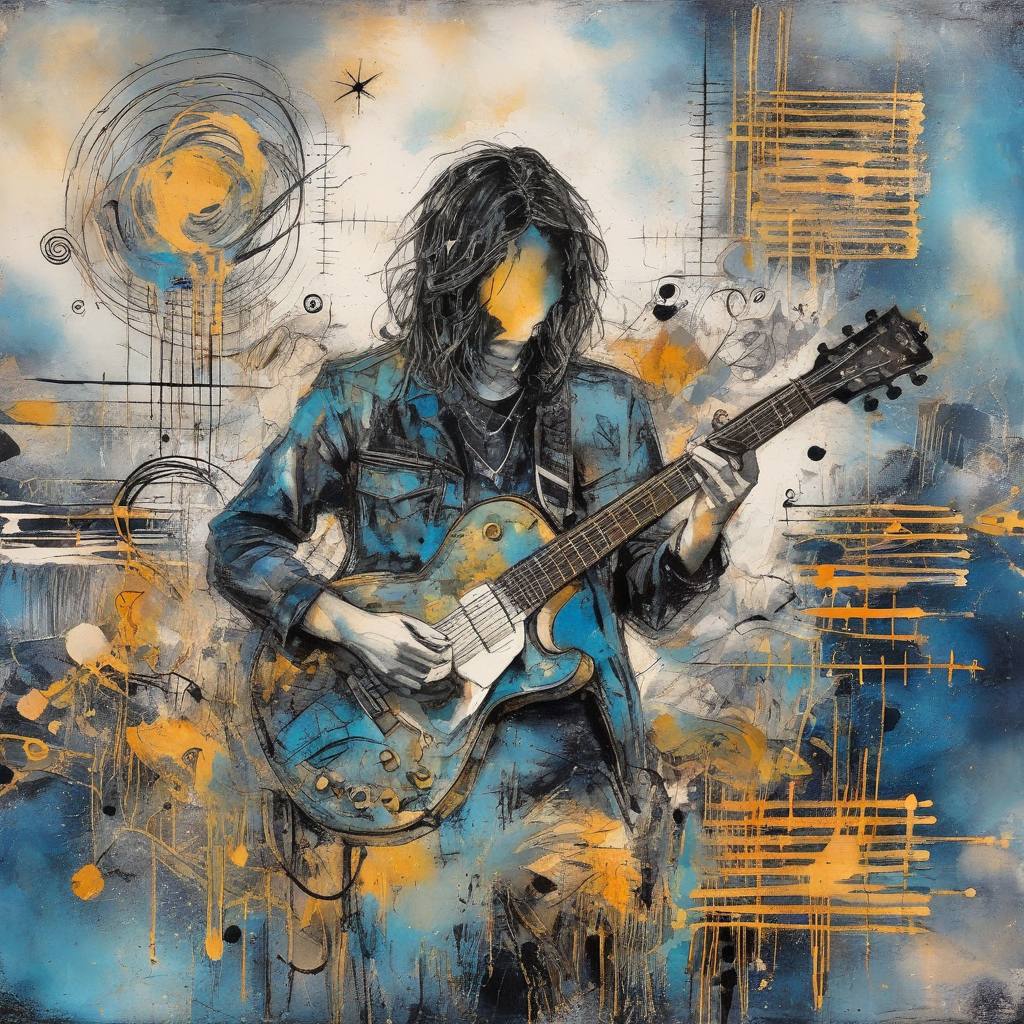} &
    \includegraphics[width=0.24\textwidth]{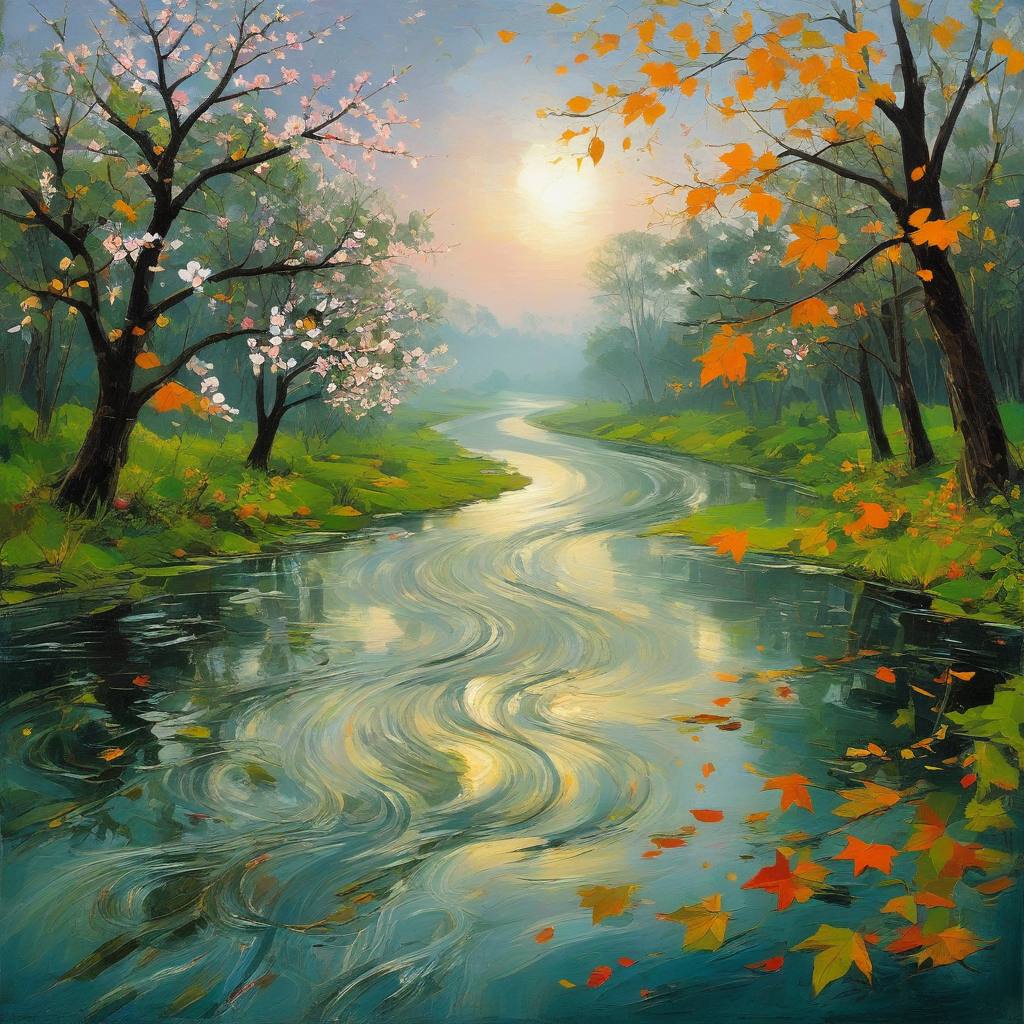} &
    \includegraphics[width=0.24\textwidth]{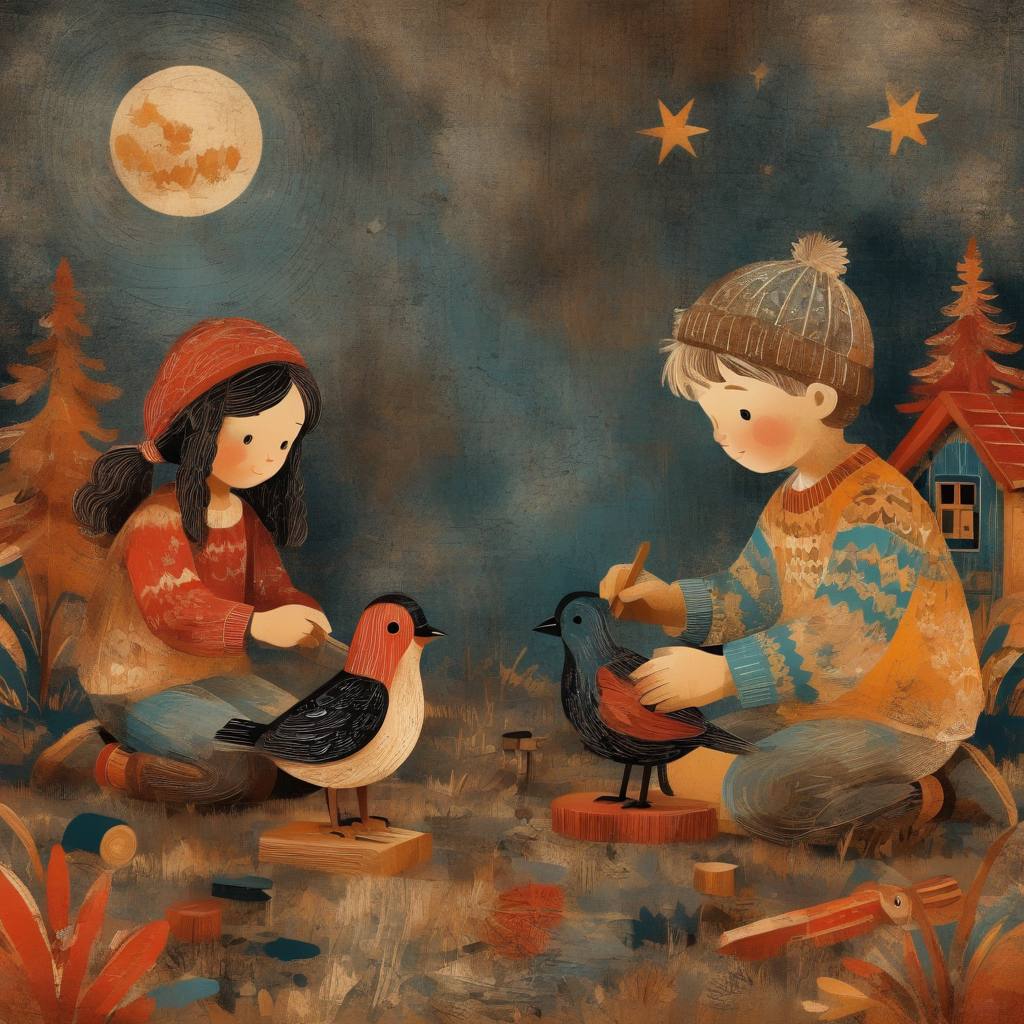}
\end{tabular}

    \caption{Examples generated by Lunara.}
    \label{fig:lunara-examples}
\end{figure*}

\begin{figure}[]
\centering

\begin{subfigure}[t]{\textwidth}
    \centering
    \begin{minipage}[t]{0.32\textwidth}
        \vspace{0pt}
        \centering
        \includegraphics[width=\linewidth]
        {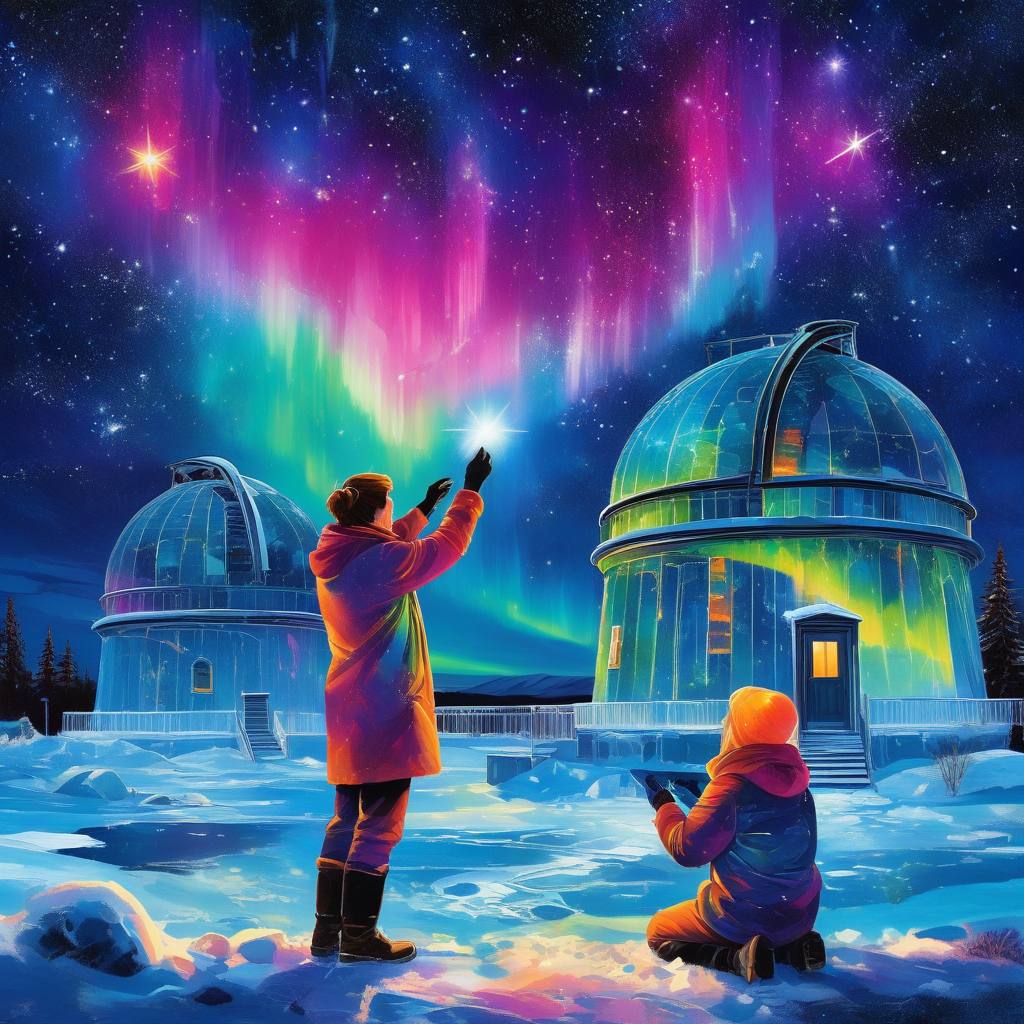}
        \vspace{-0.1em}
        {\scriptsize Lunara}
    \end{minipage}
    \hfill
    \begin{minipage}[t]{0.32\textwidth}
        \vspace{0pt}
        \centering
        \includegraphics[width=\linewidth]
        {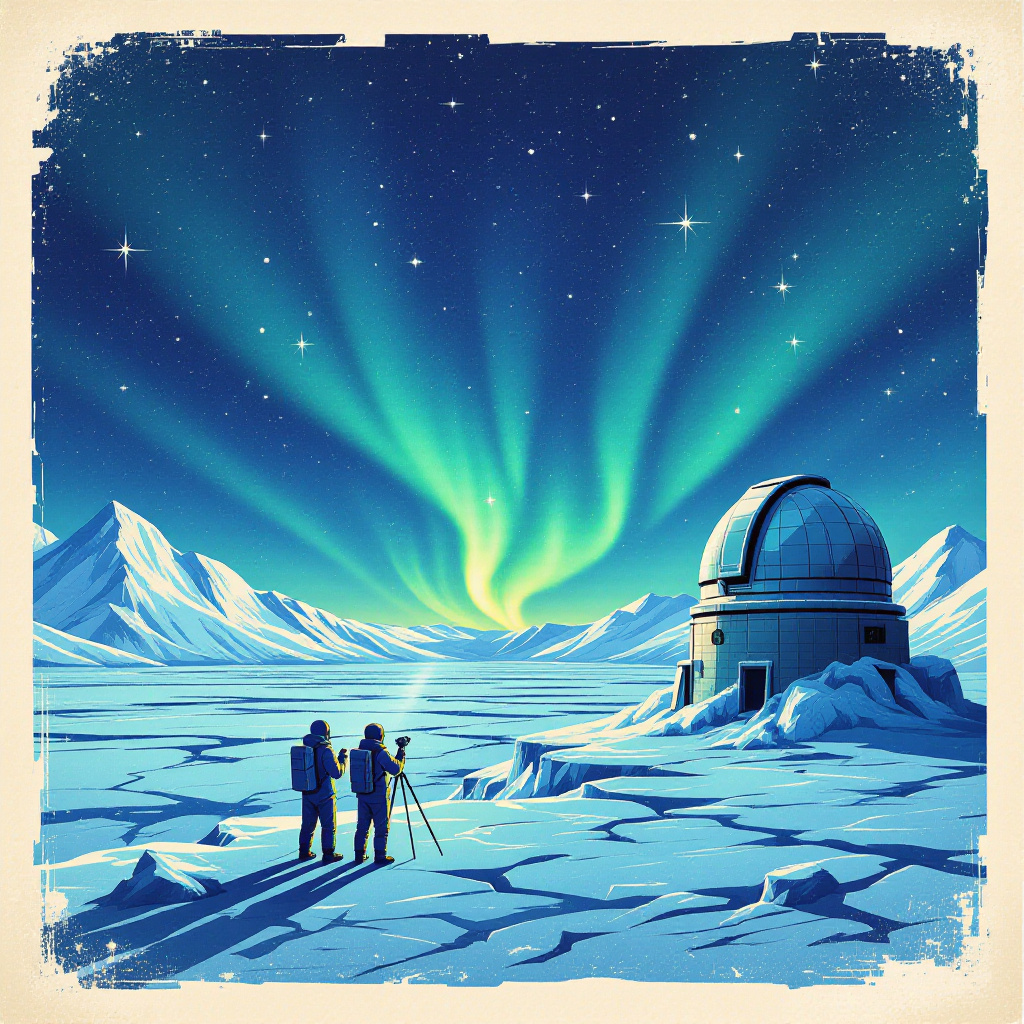}
        \vspace{-0.1em}
        {\scriptsize HiDream-I1}
    \end{minipage}
    \hfill
    \begin{minipage}[t]{0.32\textwidth}
        \vspace{0pt}
        \centering
        \includegraphics[width=\linewidth]
        {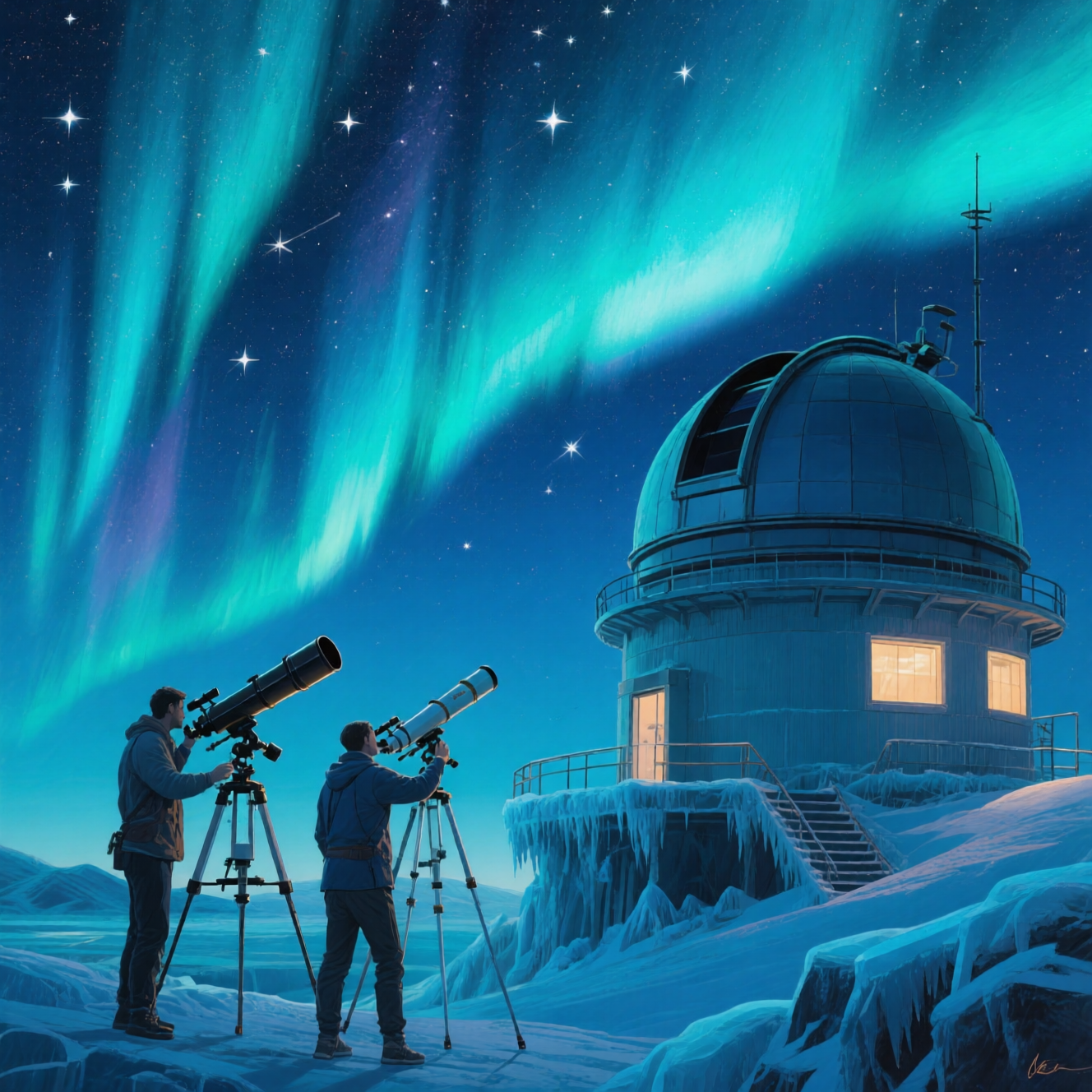}
        \vspace{-0.1em}
        {\scriptsize Qwen-Image}
    \end{minipage}

    \caption{Aurora art depicting two astronomers mapping stars beside a frozen observatory}
\end{subfigure}

\vspace{0.1em}
\begin{subfigure}[t]{\textwidth}
    \centering
    \begin{minipage}[t]{0.32\textwidth}
        \vspace{0pt}
        \centering
        \includegraphics[width=\linewidth]
        {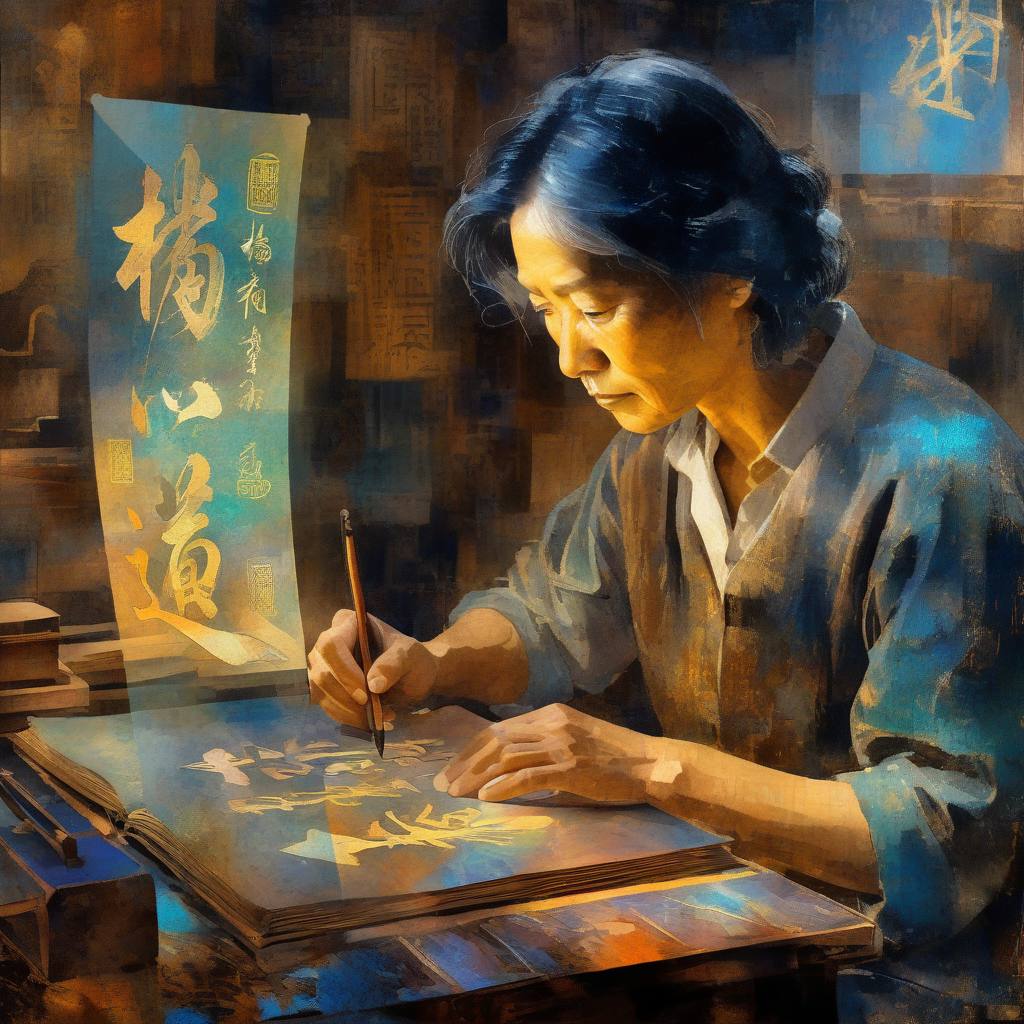}
        \vspace{-0.1em}
        {\scriptsize Lunara}
    \end{minipage}
    \hfill
    \begin{minipage}[t]{0.32\textwidth}
        \vspace{0pt}
        \centering
        \includegraphics[width=\linewidth]
        {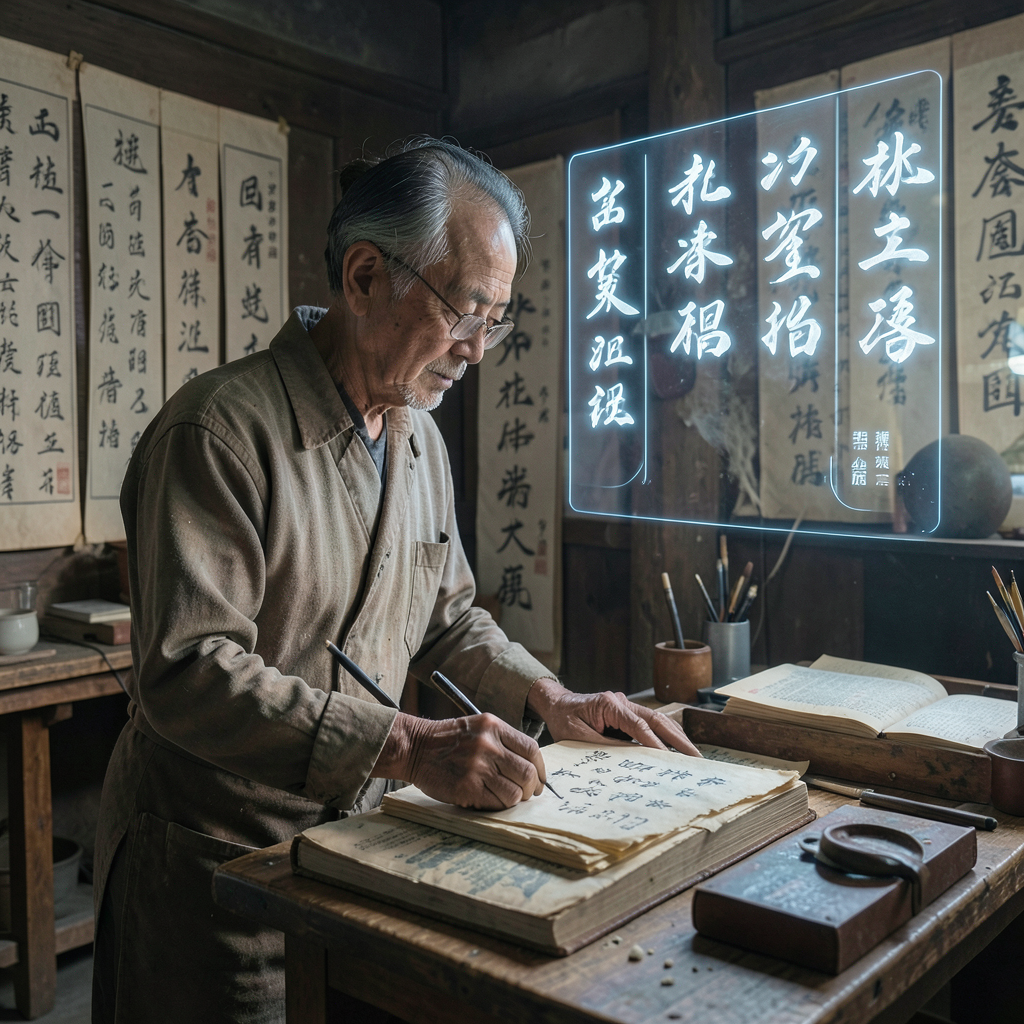}
        \vspace{-0.1em}
        {\scriptsize FLUX.2 Klein 4B Turbo}
    \end{minipage}
    \hfill
    \begin{minipage}[t]{0.32\textwidth}
        \vspace{0pt}
        \centering
        \includegraphics[width=\linewidth]
        {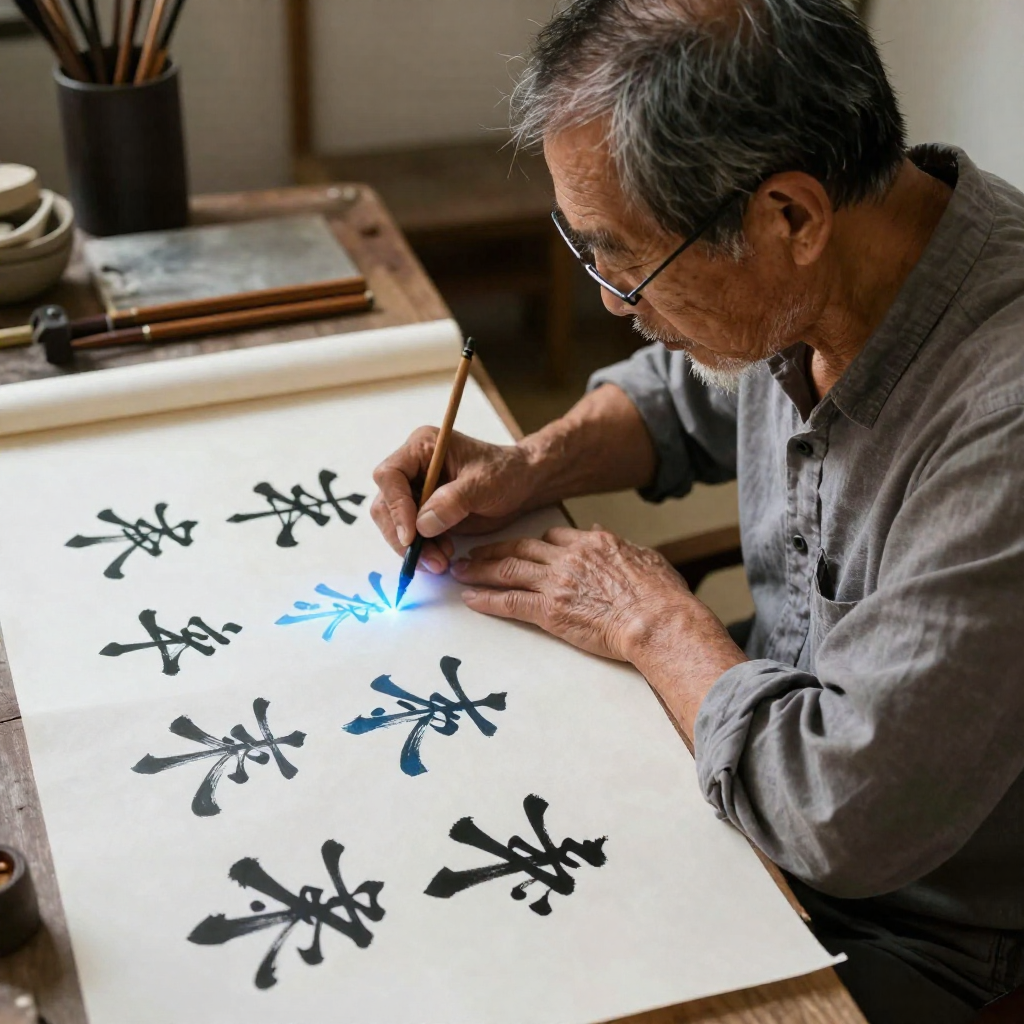}
        \vspace{-0.1em}
        {\scriptsize Z-Image-Turbo}
    \end{minipage}

    \caption{an old bookbinder working beside holographic calligraphy in East Asian digital art}
\end{subfigure}

\vspace{0.1em}






\begin{subfigure}[t]{\textwidth}
    \centering
    \begin{minipage}[t]{0.32\textwidth}
        \vspace{0pt}
        \centering
        \includegraphics[width=\linewidth]
        {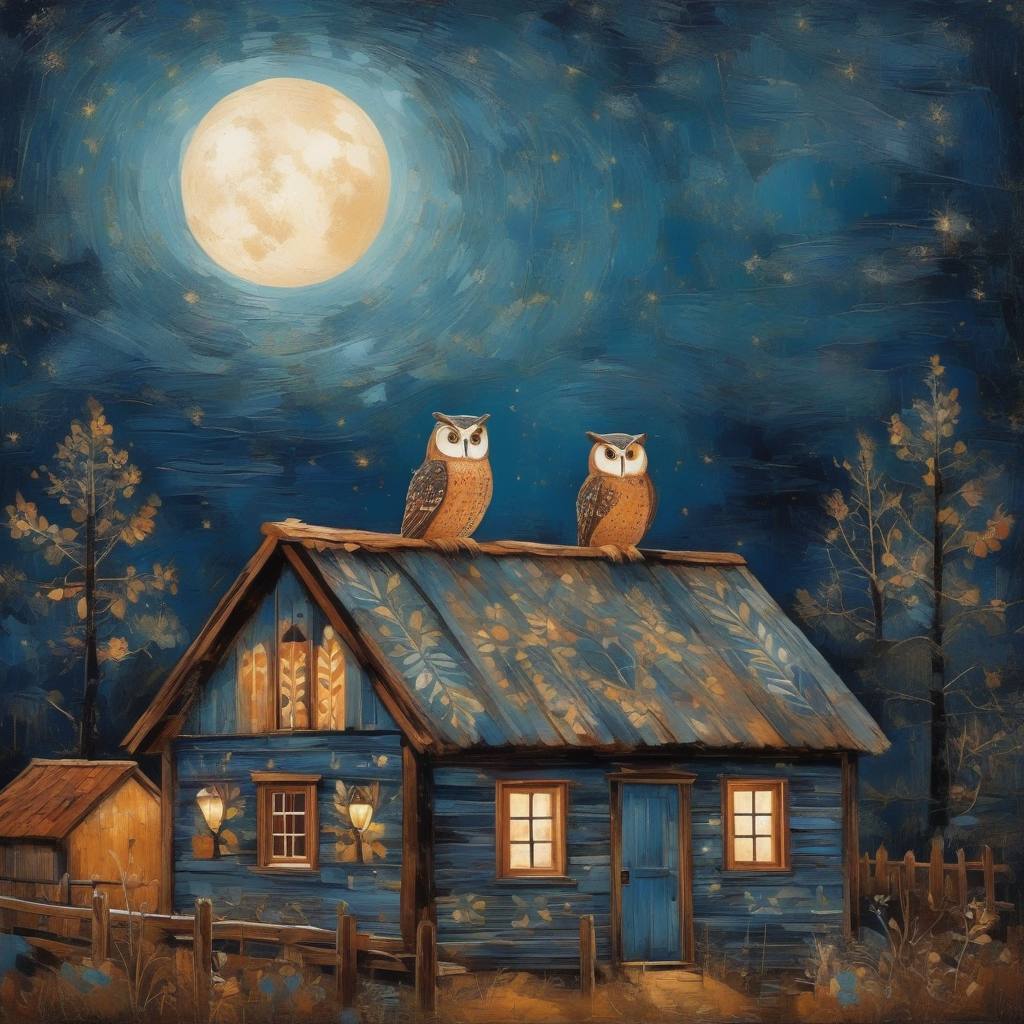}
        \vspace{-0.4em}
        {\scriptsize Lunara}
    \end{minipage}
    \hfill
    \begin{minipage}[t]{0.32\textwidth}
        \vspace{0pt}
        \centering
        \includegraphics[width=\linewidth]
        {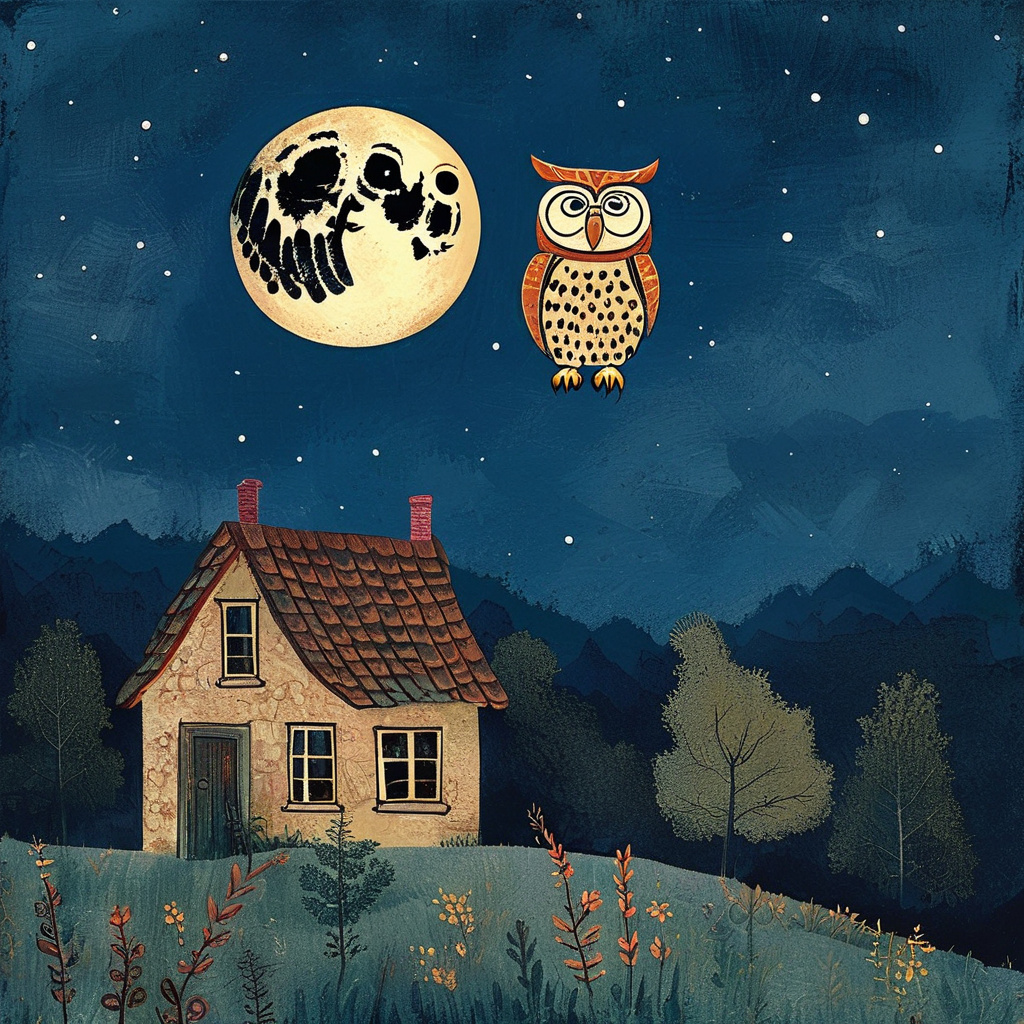}
        \vspace{-0.4em}
        {\scriptsize AuraFlow}
    \end{minipage}
    \hfill
    \begin{minipage}[t]{0.32\textwidth}
        \vspace{0pt}
        \centering
        \includegraphics[width=\linewidth]
        {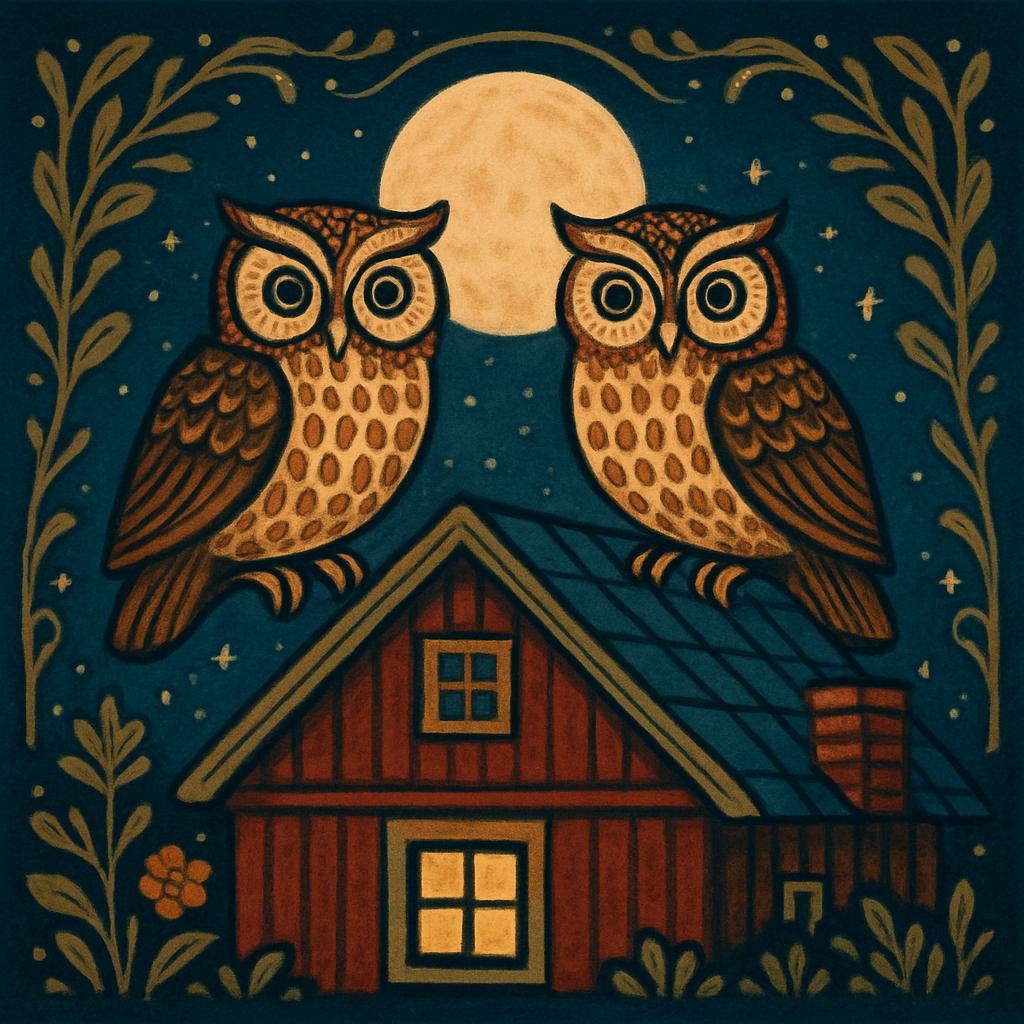}
        \vspace{-0.4em}
        {\scriptsize GPT-image-1 Mini}
    \end{minipage}

    \caption{A Nordic folk art scene of two owls resting above a moonlit farmhouse roof}
\end{subfigure}

\caption{Comparison of images generated by Lunara and other models. FLUX and Z-Image-Turbo often produce photorealistic outputs even when prompted for digital art.  GPT-image-1 Mini and  demonstrates strong aesthetic quality but AuraFlow lacks content integrity, while Lunara performs strongly across both aesthetic quality and content integrity.}

\label{fig:model-comp}

\end{figure}


\section{Introduction}



Since 2024, the frontier of image generation has been defined by the pursuit of images that are correct. Stable Diffusion 3 \cite{esser2024scaling} introduced better training formulation to yield predictable gains in fidelity and prompt adherence. Subsequent models decomposed correctness into more specific properties: rendered text, from spelling in Imagen 4 \cite{google2025imagen4} to multilingual typography in Qwen-Image-2.0 \cite{qwen2026qwenimage2} and multi-image composition in Seedream 4.0 \cite{bytedance2025seedream4}; world knowledge of depicted subjects in GPT Image 1 \cite{openai2025gptimage}; preservation under iterative editing in FLUX.1 Kontext \cite{blackforestlabs2025flux1kontext}; and identity across multiple references in FLUX.2 \cite{bfl2025flux2}. Recent models, including GPT Image 2 \cite{openai2026gptimage2} and Nano Banana 2 \cite{googledeepmind2026nanobanana2}, push correctness toward dense and structured compositions. As the field moves toward general visual intelligence, strong performance on conventional benchmarks can still diverge sharply under human judgment \cite{wu2023human,otani2023toward,jayasumana2024rethinking}. This suggests that visual correctness alone may not capture the full set of qualities people value in an image. Lunara advances the frontier through \emph{Artistic Intelligence}, the capacity to interpret meaning and construct imaginative and expressive worlds.

Following earlier releases of Lunara generated art and semantic variation datasets Lunara I \cite{wang2026moonworks} and Lunara II \cite{wang2026bmoonworks}, this paper formulates Lunara’s \emph{Artistic Intelligence} as two stages of computation that first establish semantic, artistic, and compositional constraints, then realize a visual world within them. Lunara implements this framework with a new Diffusion Mixture Transformer (DMT) architecture, drawing inspirations from Diffusion Transformers \cite{esser2024scaling,zheng2025diffusiont,zimage2026} and UNet \cite{ronneberger2015unet,rombach2022ldm,peebles2023scalable}. The DMT utilizes a unified encoder, Art Conception blocks and Compositional Attention blocks that transform input into an Artistic Frame, a contextualized representation that constrains latent refinement throughout the active Mixture; the resulting denoised latent is then decoded into pixel space.

Inspired by active learning principles \cite{Hassan2018interactive,hassan-alikhani-2023-calm,hassan-etal-2024-active,hassan-etal-2025-active,hassan2025coherence}, Moonworks has developed the CAT (composite active transfer) algorithm, a training paradigm for both pretraining and post-training where the model enters an iterative loop after a base training stage. At each iteration, the model acquires most informative examples and concentrates computation for the greatest information gain, evolving the training distribution. Lunara pushes this paradigm into generative image modeling, where the training corpus is iteratively constructed through targeted data acquisition, synthetic data expansion, and human creation and refinement. 

We benchmark Lunara against seven image-generation models spanning frontier, open-weight, and efficiency-oriented models. Across 1,000 prompts and 8,000 generated images, GPT 5.6-Sol \cite{openai2026gpt56sol} measures three observable dimensions of Artistic Intelligence as evaluator: \emph{Aesthetic Quality}, \emph{Emotional Resonance}, and \emph{Content Integrity}. The evaluation set, along with GPT 5.6 Sol scores is publicly released. \footnote{ \url{https://huggingface.co/datasets/moonworks/lunara-art-eval}}

Lunara ranks first in Aesthetic Quality and second in Emotional Resonance while remaining competitive in Content Integrity. As shown in Figure~\ref{fig:Baseline_performance}, Lunara sits near the top of both artistic performance (Aesthetic Quality and Emotional Resonance aggregated), and content integrity alongside GPT-Image-1 mini \cite{openai2025gptimage1mini} and Qwen-Image \cite{qwen2025qwenimage}. And it does so with fewer than 10B active parameters and sub-10-second wall-clock latency on 40GB A100 GPU for 1024X1024 resolution images. 

Further, we evaluate Lunara with conventional measures of prompt alignment and aesthetic preference. Lunara remains among the strongest models in both CLIPScore \cite{hessel2021clipscore} and the LAION improved aesthetic predictor \cite{schuhmann2022aesthetic}. On a four-category GenEval  \cite{ghosh2023geneval}, among 16 models, Lunara posts perfect single-object accuracy and remains among the top performers in dimensions including colors. 

Lastly, we validate the automated findings with a blind human-preference study over the earlier 1,000 prompt benchmark. Lunara’s performance holds across conventional benchmarks, our Artistic Intelligence measures, and human judgment. Six evaluators independently assess the same three dimensions, corroborating the automated results: Lunara ranks first in aesthetic quality, emotional resonance, and content integrity. Figure~\ref{fig:lunara-examples} shows representative Lunara's generations.

Together, human and machine evaluations place Lunara on the frontier of artistic intelligence at sub-10B active scale and fast inference. More fundamentally, these results establish Artistic Intelligence as a distinct axis of progress toward general visual intelligence, beyond scale and correctness alone.


\section{Artistic Intelligence Framework} 

We formalize \emph{Artistic Intelligence} as the ability to construct imaginative and expressive visual worlds while preserving what must remain true. Our framework decomposes task of image generation into first constructing an \textit{Artistic Frame} that imposes constraints and then constructing the output image through \textit{World Realization}. 

\subsection{Artistic Frame Formation}
We materialize \textit{Artistic Frame} as a set of constraints over generation, which is motivated by prior work showing that composition conditioning can preserve semantic, structural, and visual properties while retaining a large space of possible realizations \cite{pmlr-v202-huang23b}. 

\textbf{Semantic constraints} The model first constructs a semantic representation $H_{\mathrm{sem}}$, encoding the requested entities, attributes, and actions. Preserving such information is necessary because diffusion models can omit requested subjects or incorrectly bind attributes \cite{chefer2023}.

\textbf{Artistic constraints.}  The model constructs an artistic representation $H_{\mathrm{art}}$, describing how the content should be expressed through aesthetic, stylistic, and affective properties. This representation constrains artistic identity without prescribing a single realization, preserving expressive degrees of freedom within the desired artistic space \cite{wang2024instantstyle}.

\textbf{Spatial and structural constraints.} Text-to-image models often recover the right concepts but lose their spatial organization \cite{feng2023,gokhale2023benchmarking}. Inspired by structural conditioning for layout and geometry control \cite{zhang2023controlnet}, we introduce $H_{\mathrm{comp}}$ to encode the relationships that must remain invariant, including spatial organization, directional and intensity patterns. 

Together, the Artistic Frame is formed as \(F= \phi(H_{\mathrm{sem}}, H_{\mathrm{art}}, H_{\mathrm{comp}})\), the set of semantic, artistic, and compositional constraints that define what must remain true, while leaving the remaining degrees of freedom open to imaginative and expressive realization. 

\subsection{World Realization}

\emph{World Realization} is the process of turning visual intent into a complete and expressive world. Faithful object placement and surface aesthetics are necessary, but are not sufficient for Artistic Intelligence. Generative models can still produce shallow, repetitive images \cite{10.1093/jaac/kpag027} or collapse toward a narrow set of generic motifs \cite{HINTZE2026101451}. An artistically intelligent model must preserve room for invention across composition, subject matter, texture, atmosphere, and affect. This is realized through mixtures of learned generative pathways spanning artistic, regional, and visual dimensions.

During generation, the active mixture determines which pathways are emphasized and how they are combined, allowing the model to resolve uncertainty through different expressive outcomes rather than converge on a narrow visual formula. We hypothesize that, as a model evolves, this capacity may lead to the emergence of new artistic styles as a manifestation of Artistic Intelligence. The exploration remains bounded by the \emph{Artistic Frame}, which preserves the semantic, artistic, and compositional structure needed to keep output coherent and meaningful. \emph{World Realization} therefore defines what the model is free to invent, while the \emph{Artistic Frame} preserves what must remain true.


Formally, given an Artistic Frame
\(F=\phi(H_{\mathrm{sem}},H_{\mathrm{art}},H_{\mathrm{comp}})\),
we model World Realization as

\begin{equation}
p_{\mathrm{WR}}(\mathbf{x}\mid F)
=
\sum_{m=1}^{K}
p_{\eta}(m\mid F)\,
p_{\theta}(\mathbf{x}\mid F,m),
\end{equation}

where \(p_{\eta}(m\mid F)\) determines the active mixture induced by
the Artistic Frame, and \(p_{\theta}(\mathbf{x}\mid F,m)\) captures the
space of expressive visual realizations generated through that mixture.

\section{Lunara Architecture: Diffusion Mixture Transformer}

\begin{figure}[]
    \centering
    \includegraphics[width=\textwidth]{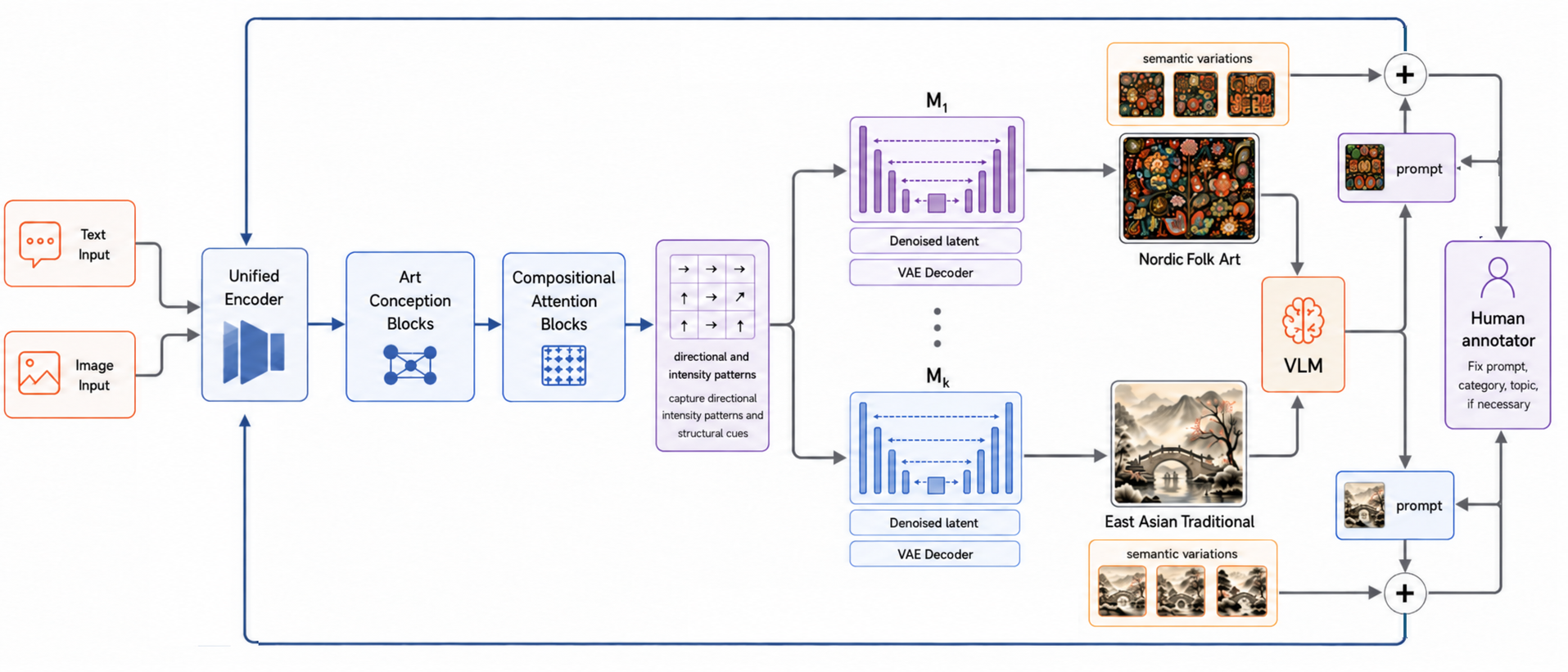}
    \caption{Overview of the Lunara architecture and iterative training data acquisition pipeline. Text and optional image input are encoded and transformed into semantic, artistic, and compositional representations, which guide active Mixture for image generation. A VLM identifies informative samples for semantic variation, which are added to the training set alongside selective human-created artwork.}
    \label{fig:architecture}
\end{figure}

Lunara utilizes a novel Diffusion Mixture Transformer architecture to implement the Artistic Intelligence framework. See Figure~\ref{fig:architecture} for the overall architecture.  

\subsection{Artistic Frame Blocks}
\label{subsec:Constraint Formation}

\emph{Artistic Frame Blocks} construct the semantic, artistic, and compositional representations that define what must remain true. 

\textbf{Semantic and Art Conception Blocks}
Given an image--text pair $({\mathbf{x}}_{i}, {p}_{i})$, a unified encoder encodes $x_i,p_i$ and projects it into $H_{sem,i}$. Art Conception blocks, ${A}_{\psi}$, map $H_{sem,i}$ into an artistic representation space, yielding ${H}_{\mathrm{art},i} = {A}_{\psi}\left({H}_{\mathrm{sem},i}\right)$. 

\textbf{Compositional Attention Blocks}
Compositional Attention blocks \(G_{\omega}\) model ${H}_{\mathrm{sem},i}$, learning what content must be preserved. During training, a visual structure extractor \(F_{\mathrm{comp}}\) extracts a structural composition representation from each image \(\mathbf{x}_i\), $H_{\mathrm{comp},i}=F_{\mathrm{comp}}(\mathbf{x}_i)$. The extracted composition is then encoded by the VAE into a latent compositional representation, $\mathbf{z}_{\mathrm{comp},i}$. A continuous training timestep $\tau$ is sampled by drawing $u \sim \mathcal{N}(\mu,\sigma^2)$ and mapping it to $(0,1)$ through the logistic function,
\begin{equation}
\tau = \operatorname{sigmoid}(u) = \frac{1}{1+e^{-u}}.
\end{equation}
We then sample Gaussian noise $\boldsymbol{\epsilon}_i \sim \mathcal{N}(\mathbf{0},\mathbf{I})$ and construct the noisy latent as 
\begin{equation}
\mathbf{z}_{i,\tau} = \tau \mathbf{z}_{comp, i} + (1-\tau)\boldsymbol{\epsilon}_i,
\end{equation}

, where $\tau=1$ corresponds to the target latent and $\tau=0$ to pure Gaussian noise. The corresponding target velocity is

\begin{equation}
\mathbf{v}_{i}
=\frac{d\mathbf{z}_{i,\tau}}{d\tau}
=\mathbf{z}_{comp, i}
- \boldsymbol{\epsilon}_i
\end{equation}

Given the noisy compositional latent $\mathbf{z}_{i,\tau}$, timestep $\tau$, semantic $H_{sem, i}$, the compositional attention blocks predict the velocity field
\begin{equation}
\hat{\mathbf{v}}_{i,\tau} = G_{\omega}\left(\mathbf{z}_{i,\tau}, {H_{sem, i}}, \tau\right).
\end{equation}
The blocks are then trained by minimizing the velocity-matching objective
\begin{equation}
\mathcal{L}_{\mathrm{comp}} = \mathbb{E}_{i,\tau,\boldsymbol{\epsilon}} \left[ \left\| \hat{\mathbf{v}}_{i,\tau} - \mathbf{v}_{i} \right\|_2^2 \right].
\end{equation}

The Artistic Frame \(F= \phi(H_{\mathrm{sem}}, H_{\mathrm{art}}, H_{\mathrm{comp}})\) serves as the final conditioning representation for generation, encoding the semantic, artistic, and compositional constraints established above.

\subsection{Latent Mixtures}
\label{subsec:Diffusion Mixture Transformer}
The Artistic Frame specifies the constraints of the world and active mixture selects the generative pathways needed for World Realization to bring it to life. Across the generative trajectory, the Artistic Frame remains as a persistent conditioning state, while the active mixture iteratively resolves the unconstrained degrees of freedom into an expressive visual world. Given the Artistic Frame $F$, mixture activation is modeled as
$p_{\eta}(m \mid F)$. 

During training for each mixture, a VAE encodes each image \(\mathbf{x}_i\) into a clean latent representation \(\mathbf{z}_{i,0}\). We then sample a discrete timestep \(t \sim p(t)\), with
\(p(t)=\mathrm{Uniform}(\{1,\ldots,T\})\), together with Gaussian noise
\(\boldsymbol{\epsilon}_i \sim \mathcal{N}(\mathbf{0},\mathbf{I})\). Given the predefined noise schedule \(\alpha_t\), with \(\bar{\alpha}_t=\prod_{s=1}^{t}\alpha_s\), we construct the noisy latent \(\mathbf{z}_{i,t}\) as:
\begin{equation}
\mathbf{z}_{i,t} = \sqrt{\bar{\alpha}_t} \mathbf{z}_{i,0} + \sqrt{1 - \bar{\alpha}_t} \boldsymbol{\epsilon}_i.
\end{equation}


Each mixture is instantiated with learnable parameters $\theta$, jointly modeling transformations across artistic, regional, and cultural representation spaces. Conditioned on artistic frame $F_i$, the noisy latent $\mathbf{z}_{i,t}$, and timestep $t$, the denoising network  $\boldsymbol{\epsilon}_{\theta}$ predicts the noise added to the clean latent. Each mixture is optimized using a mean squared error objective:

\begin{equation}
\mathcal{L}_{\mathrm{noise}} = \mathbb{E}_{i, t, \boldsymbol{\epsilon}_i} \left[ \frac{1}{d}\left| \boldsymbol{\epsilon}_i - \boldsymbol{\epsilon}_\theta\left(\mathbf{z}_{i,t}, t, F_i\right) \right|_2^2 \right].
\end{equation}

\section{Iterative Training and Semantic Variations}

An initial general training stage follows a standard text-to-image training objective using broad image-text pairs from our training dataset. For images without accompanying captions, we use a vision-language model to generate text descriptions. This stage establishes a broad generative foundation, which is subsequently reshaped through iterative training.

\subsection{Iterative Training}
An iterative training stage concentrates learning on highly informative artistic and semantically controlled samples, enabling the model to develop stronger artistic representations without sacrificing content integrity. 
Traditional active learning ranks unlabeled samples by predictive uncertainty, commonly using entropy,
\(x^* = \arg\max_{x \in \mathcal{U}} \left[-\sum_{k=1}^{K} p_\theta(y_k \mid x)\log p_\theta(y_k \mid x)\right]\).
For generative models, entropy over the full output space is often intractable and may remain poorly aligned with the informativeness of the sample. To address this gap, a regulated-attribute-conditioned score has been introduced to replace entropy-based acquisition \cite{hassan-etal-2025-active}: 

\begin{equation}
x_E^*
=
\arg\max_{x \in \mathcal{U}}
\left[
-\operatorname{Softmax}\left(R(G(x),H)\right)
\right],
\end{equation}

where \(G(x)\) is the learner model output, \(H\) is the regulated attribute, and \(R\) is an auxiliary transformer conditioned on \(H\). The corresponding informativeness score is
\(E_i=-\operatorname{Softmax}(R(G(x_i),H))\).

A subsequent formulation \cite{hassan2025coherence} generalizes this from a single regulated attribute to multiple attributes:

\begin{equation}
x_E^*
=
\arg\max_x
\left[
-\sum_k
Z_\theta\left(p_k \mid G(x)\right)
\right],
\end{equation}

where \(Z_\theta\) is the external transformer and
\(p_k \in \mathcal{P}=\{p_1,\ldots,p_K\}\) denotes a task-relevant generation attribute. 

Moonworks has recast active learning into a training paradigm spanning pretraining and post-training. Lunara pushes this paradigm into generative image modeling, where acquisition is reformulated around the properties and objectives of the output space. Following iteration $k$, the training loop surfaces underrepresented capabilities and failures. These shape the acquisition and composition of the next training mixture across human-created, human-refined, and targeted synthetic samples. Training therefore progressively concentrates on underrepresented regions, converting gaps in iteration $k$ into acquisition priorities for iteration $k+1$. 

We define the training set $\mathcal{D}^{(k)}$ partitioned into clusters $\mathcal{C}^{(k)} = \{C_{1}^{(k)}, \ldots, C_{J}^{(k)}\}$ based on style, region, and category at acquisition round $k$ as clustering can effectively organize data for segmentation and augmentation \cite{sicilia2023isabel,asano2025contextual}. For each candidate example $x$ with its prompt $P$, a VLM is used to assign a composite score $s_{composite}(x,P)$ across {N} dimensions, where each component evaluates a different aspect of the generated image relative to the prompt.
 
We select the top $M$ examples with the highest acquisition scores to form a target anchor set $\mathcal{A}_{\mathrm{target}}^{(k)} = \{a_1, \dots, a_M\}$. Each anchor $a_i=(\mathbf{x}_i,p_i,F_i)$ contains a image $\mathbf{x}_i$, its prompt $p_i$, and corresponding Artistic Frame ${F}_i$ computed by the Compositional Attention blocks. 

Each selected image is then refined by an edit model $R$ conditioned on its prompt: $\mathbf{x}_i' = \mathrm{R}_{\mathrm{edit}}\left(\mathbf{x}_i, p_i\right) $. The refined anchor becomes $a_i'=(\mathbf{x}_i',p_i, F_i)$, yielding the updated target set $\mathcal{A}_{\mathrm{target}}^{(k)} = \{a_1', \dots, a_M'\}$.

To ground this targeted refinement in human artistic practice, we additionally draw from a pool of human-created artwork. We encode these works into the same representation space and use a VLM to identify a subset, $\mathcal{A}_{\mathrm{human}}^{(k)}$, that lies near the selected target anchors. The final anchor set, $\mathcal{A}^{(k)}=\mathcal{A}_{\mathrm{target}}^{(k)}\cup\mathcal{A}_{\mathrm{human}}^{(k)}$, combines model-targeted refinement with human artistic references. We then expand each anchor into a controlled neighborhood of semantic variations for further training.



\subsection{Semantic Variations}

\begin{figure}[]
\centering

\begin{subfigure}[t]{0.48\textwidth}
    \centering
    \begin{minipage}[t]{0.48\textwidth}
        \vspace{0pt}
        \centering
        \includegraphics[height=2.8cm,keepaspectratio]{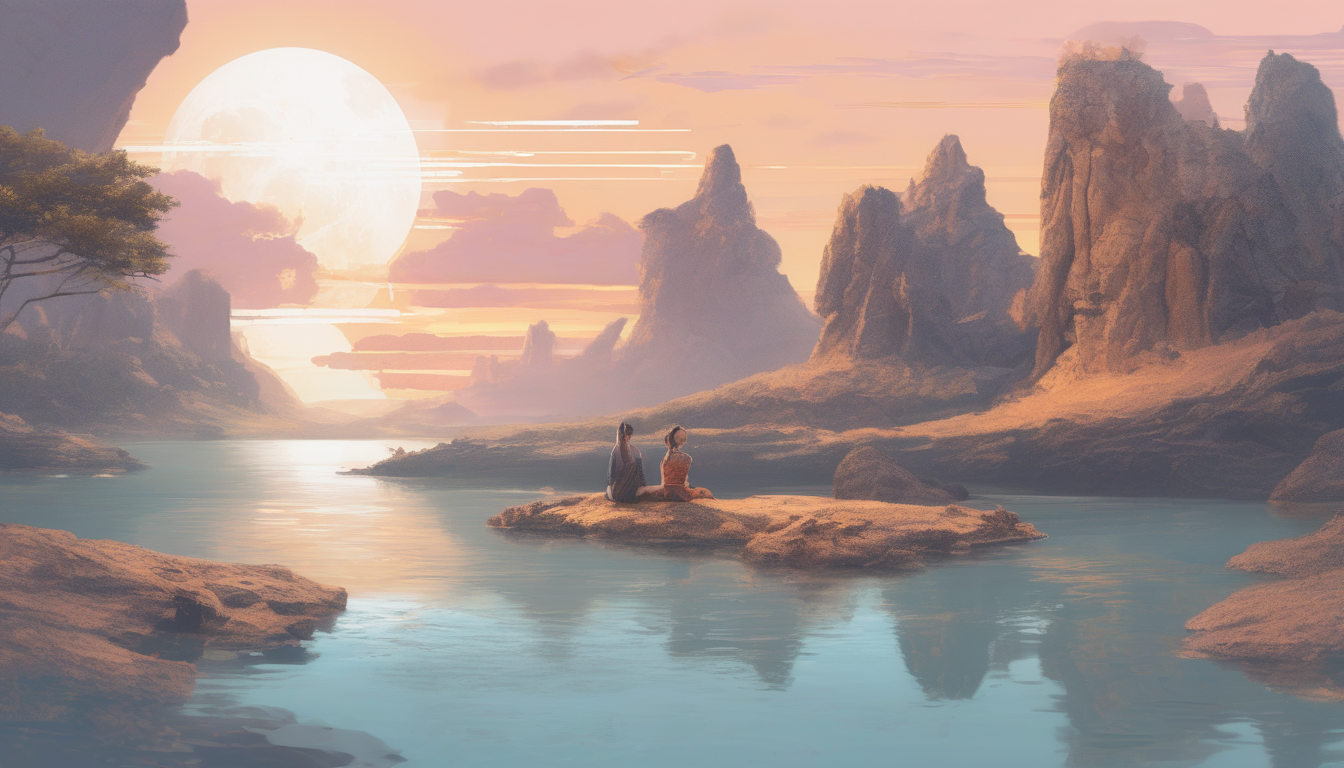}
        \vspace{-0.4em}

    \end{minipage}
    \hfill
    \begin{minipage}[t]{0.48\textwidth}
        \vspace{0pt}
        \centering
        \includegraphics[height=2.8cm,keepaspectratio]{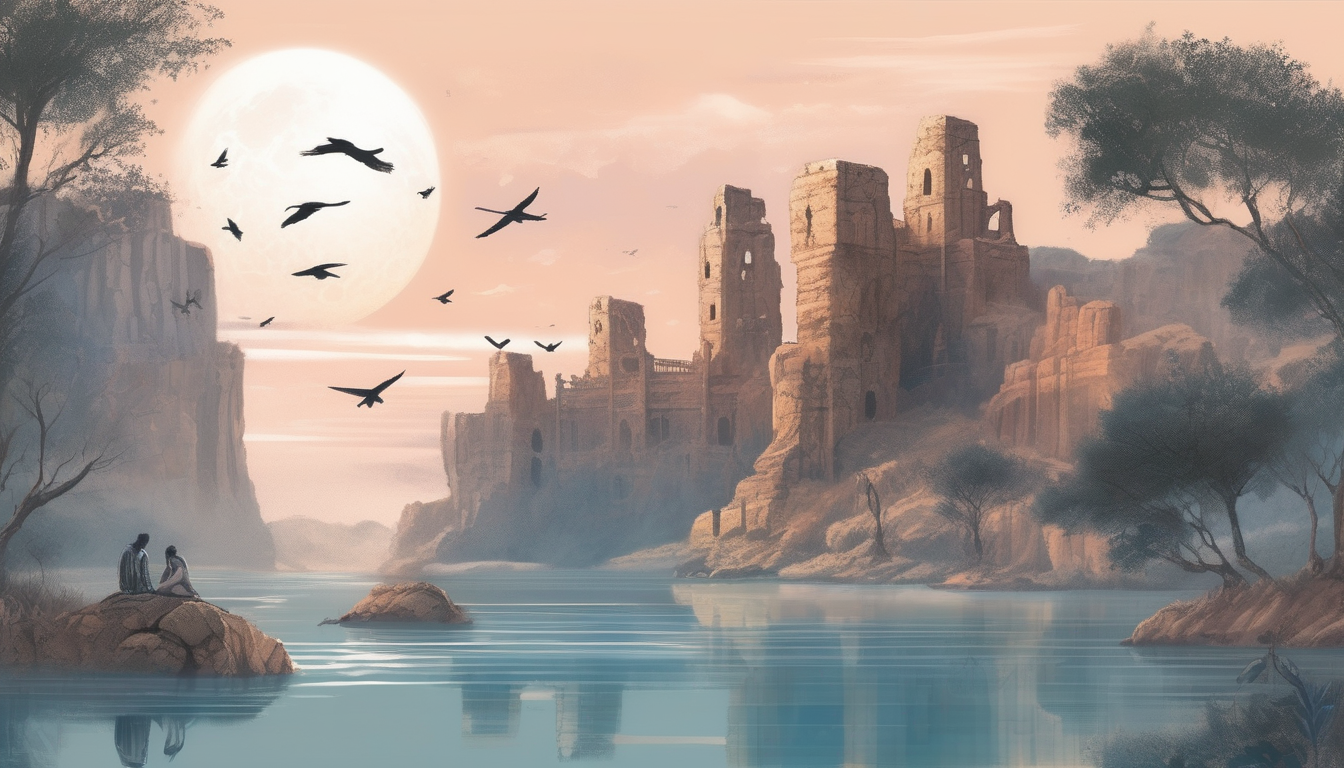}
        \vspace{-0.4em}

    \end{minipage}
    \caption{Change in scene composition.}
\end{subfigure}
\hfill
\begin{subfigure}[t]{0.48\textwidth}
    \centering
    \begin{minipage}[t]{0.48\textwidth}
        \vspace{0pt}
        \centering
        \includegraphics[height=2.8cm,keepaspectratio]{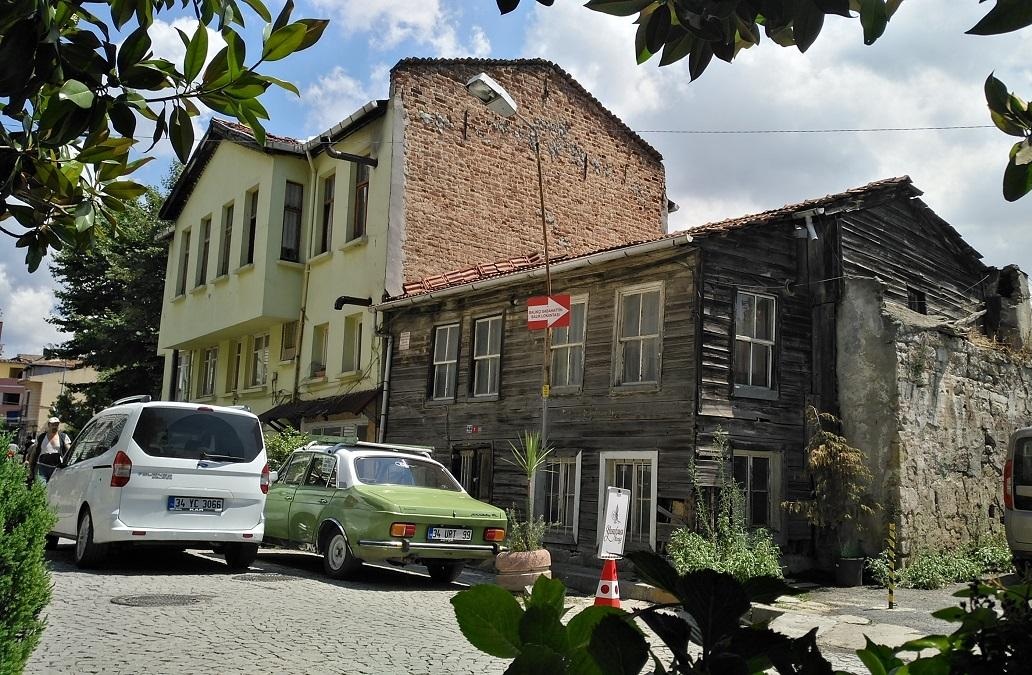}
        \vspace{-0.4em}

    \end{minipage}
    \hfill
    \begin{minipage}[t]{0.48\textwidth}
        \vspace{0pt}
        \centering
        \includegraphics[height=2.8cm,keepaspectratio]{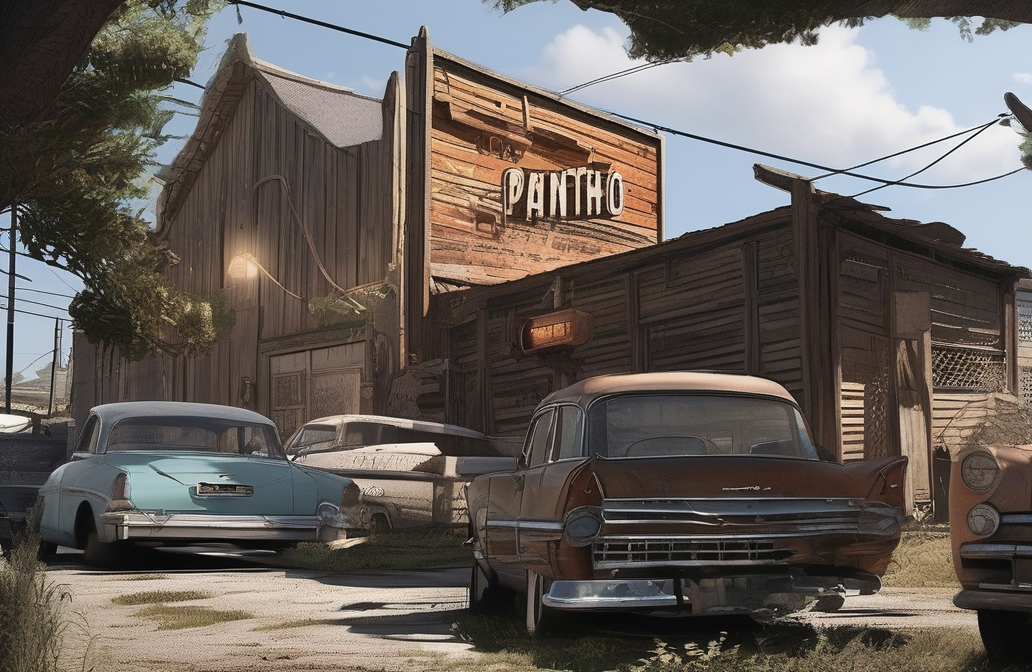}
        \vspace{-0.4em}

    \end{minipage}
    \caption{Change in scene composition.}
\end{subfigure}

\begin{subfigure}[t]{0.48\textwidth}
    \centering
    \begin{minipage}[t]{0.48\textwidth}
        \vspace{0pt}
        \centering
        \includegraphics[height=2.8cm,keepaspectratio]{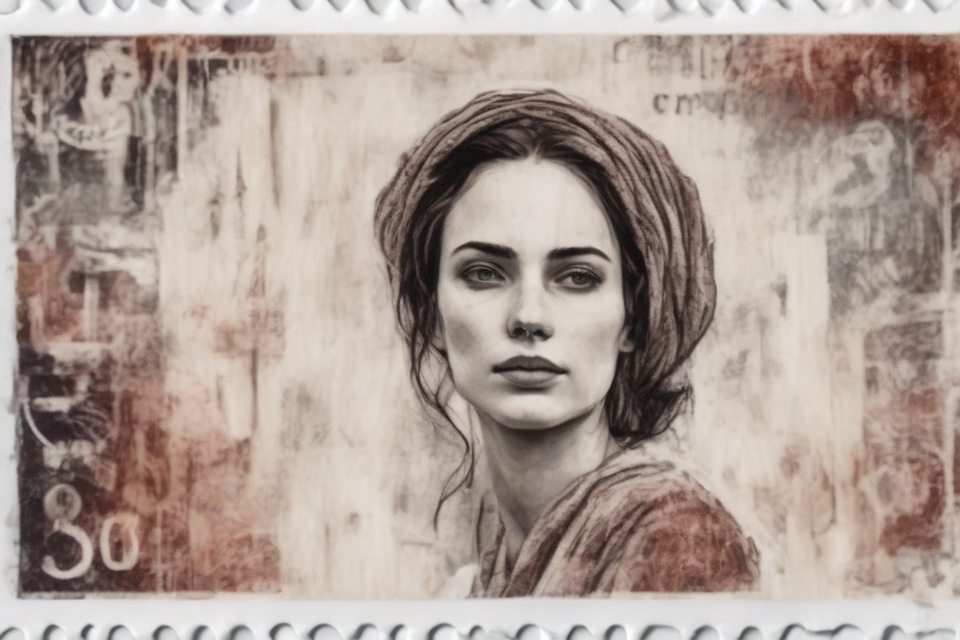}
        \vspace{-0.4em}

    \end{minipage}
    \hfill
    \begin{minipage}[t]{0.48\textwidth}
        \vspace{0pt}
        \centering
        \includegraphics[height=2.8cm,keepaspectratio]{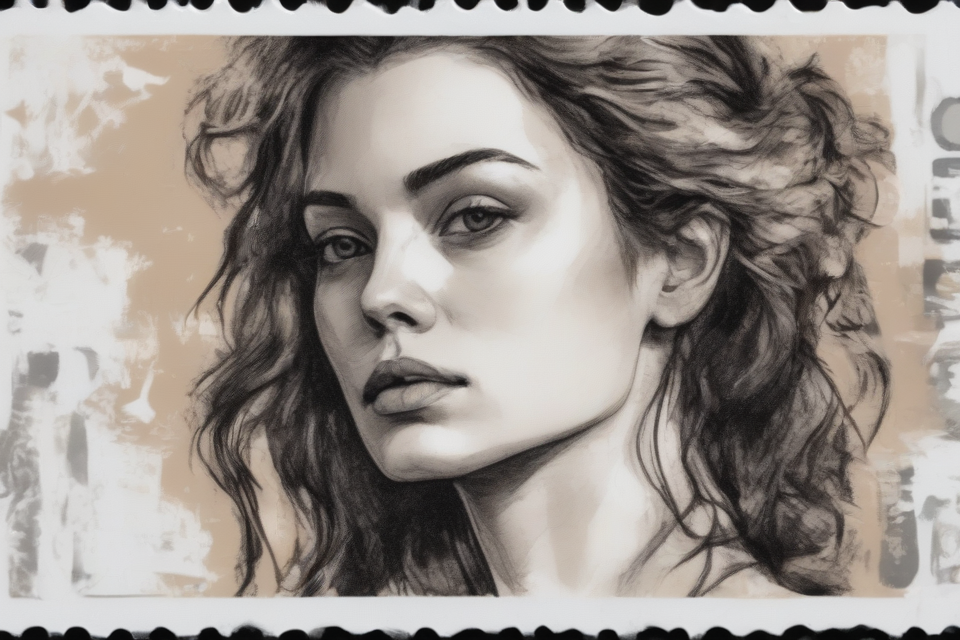}
        \vspace{-0.4em}

    \end{minipage}
    \caption{Change in viewpoint and color tone.}
\end{subfigure}
\hfill
\begin{subfigure}[t]{0.48\textwidth}
    \centering
    \begin{minipage}[t]{0.48\textwidth}
        \vspace{0pt}
        \centering
        \includegraphics[height=2.8cm,keepaspectratio]{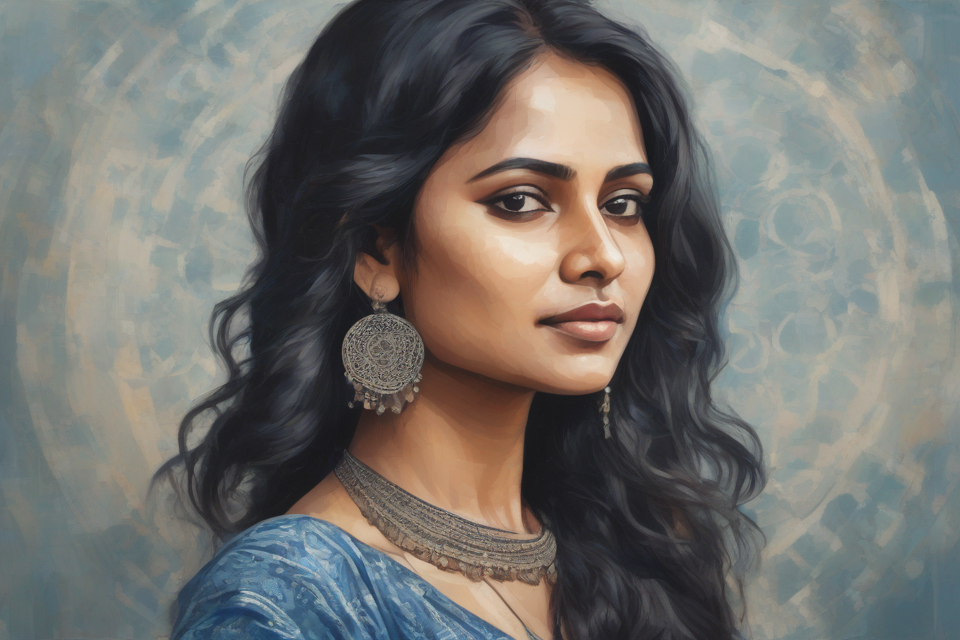}
        \vspace{-0.4em}

    \end{minipage}
    \hfill
    \begin{minipage}[t]{0.48\textwidth}
        \vspace{0pt}
        \centering
        \includegraphics[height=2.8cm,keepaspectratio]{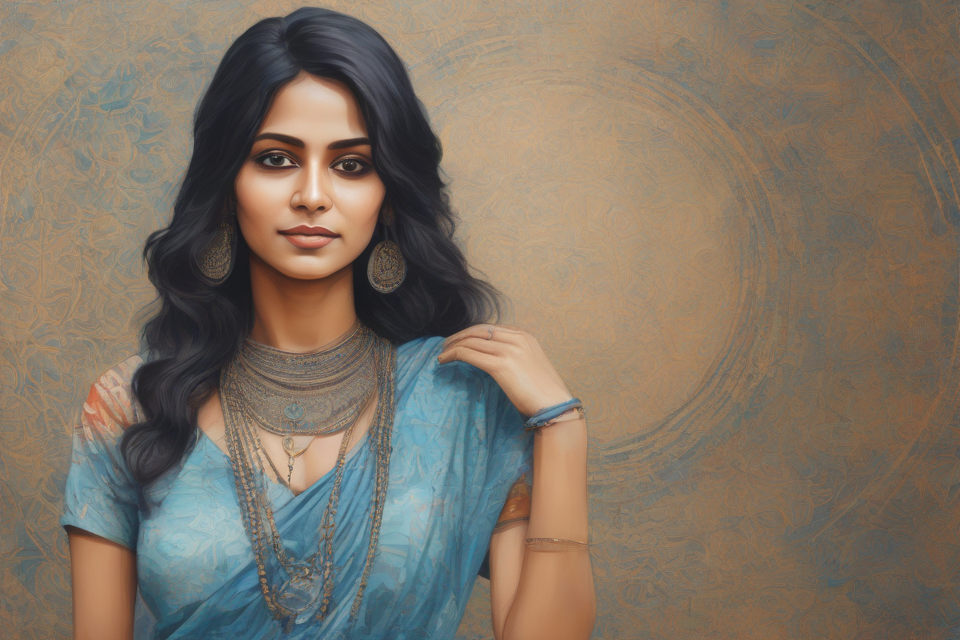}
        \vspace{-0.4em}

    \end{minipage}
    \caption{Change in viewpoint and color tone.}
\end{subfigure}

\caption{Examples of semantic variations in original artwork. Each pair shows an anchor image (left) and a semantically modified variant (right).}
\label{fig:contextual-variations-art}
\end{figure}

Next, for each anchor $a_i \in \mathcal{A}^{(k)}$, we use a VLM to construct a collection of semantic variations $\mathcal{V}_i = \{v_{i,1}, \dots, v_{i,R_i}\}$. Each variation modifies one or more meaningful visual attributes while
preserving the core content of the anchor. Typical modifications include
illumination, weather, season, viewpoint, camera distance, and scene composition. Figure ~\ref{fig:contextual-variations-art} shows examples of semantic variations created.

For each anchor $a_i=(\mathbf{x}_i,p_i,F_i)$, a VLM generates semantic variations by modifying the prompt to produce $p_{i,r}$. Lunara then constructs the corresponding Artistic Frame $F_{i,r}$, which conditions the generation of the modified image $\mathbf{x}_{i,r}$. Each variation is represented as $v_{i,r}=(\mathbf{x}_{i,r},p_{i,r},F_{i,r})$, and the resulting images are reviewed by human annotators.

Each anchor and its associated variants define a local neighborhood of semantically and artistically related samples, rather than a single isolated image–prompt pair. At acquisition round $k$, these neighborhoods are aggregated with the selected anchors to form the acquisition set $\mathcal{N}^{(k)}=\mathcal{A}^{(k)}\cup\mathcal{V}^{(k)}$, where $\mathcal{V}^{(k)}=\bigcup_{i=1}^{M}\mathcal{V}_i$. The training corpus is then expanded as $\mathcal{D}^{(k+1)}=\mathcal{D}^{(k)}\cup\mathcal{N}^{(k)}$, providing the data for the subsequent training round. Algorithm \ref{alg:generation} formalizes Iterative Training with Semantic Variations. 

\begin{algorithm}[]
\caption{Iterative Training with Semantic Variations}
\label{alg:generation}
\small
\begin{algorithmic}

\State $\mathcal{D}^{(1)}
\gets \mathrm{Caption}(\mathcal{D}_{\mathrm{broad}},f_{\mathrm{VLM}})$
\State $\theta^{(1)} \gets \mathrm{TrainT2I}(\mathcal{D}^{(1)})$

\For{$k \in \{1,2,\dots,K\}$}

    \State $\mathcal{C}^{(k)}
    \gets \mathrm{Cluster}(\mathcal{D}^{(k)};
    \mathrm{style},\mathrm{region},\mathrm{category})$

    \For{$(\mathbf{x}_i,p_i)\in\mathcal{C}^{(k)}$}
        \State $F_i \gets \phi(H_{\mathrm{sem},i}, H_{\mathrm{art},i}, H_{\mathrm{comp},i})$         \Comment{Artistic Frames}

        \State $s_{\mathrm{composite}}(\mathbf{x}_i,p_i)
        \gets f_{\mathrm{VLM}}^{\mathrm{score}}(\mathbf{x}_i,p_i)$
        \Comment{Composite score over $N$ dimensions}
    \EndFor

    \State $\widetilde{\mathcal{A}}_{\mathrm{target}}^{(k)}
    \gets \operatorname{Top}_{M}
    (\mathcal{C}^{(k)},s_{\mathrm{composite}})$

    \State $\mathcal{A}_{\mathrm{target}}^{(k)}
    \gets
    \left\{
    \bigl(R_{\mathrm{edit}}(\mathbf{x}_i,p_i),p_i,F_i\bigr)
    :
    (\mathbf{x}_i,p_i,F_i)
    \in\widetilde{\mathcal{A}}_{\mathrm{target}}^{(k)}
    \right\}$

    \State $\mathcal{A}_{\mathrm{human}}^{(k)}
    \gets
    \mathrm{RetrieveHumanArt}
    (\mathcal{A}_{\mathrm{target}}^{(k)},\mathcal{H})$

    \State $\mathcal{A}^{(k)}
    \gets
    \mathcal{A}_{\mathrm{target}}^{(k)}
    \cup\mathcal{A}_{\mathrm{human}}^{(k)}$\Comment{Combined anchor set}

    \State $\mathcal{V}^{(k)} \gets \emptyset$

    \For{$a_i=(\mathbf{x}_i,p_i,F_i)\in\mathcal{A}^{(k)}$}
        \State $\{p_{i,r}\}_{r=1}^{R_i}
        \gets
        f_{\mathrm{VLM}}^{\mathrm{variation}}
        (\mathbf{x}_i,p_i,\mathcal{S})$

        \For{$r\in\{1,\dots,R_i\}$}\Comment{create variations}
            \State $F_{i,r}\gets\phi(H_{\mathrm{sem},i,r}, H_{\mathrm{art},i,r}, H_{\mathrm{comp},i,r})$
            \State $\mathbf{x}_{i,r}
            \gets G_{\theta^{(k)}}(p_{i,r},F_{i,r})$
            \State $v_{i,r}
            \gets(\mathbf{x}_{i,r},p_{i,r},F_{i,r})$

            \If{$f_{\mathrm{Human}}(v_{i,r},a_i)=\mathrm{accept}$}
                \State $\mathcal{V}^{(k)}
                \gets\mathcal{V}^{(k)}\cup\{v_{i,r}\}$
            \EndIf
        \EndFor
    \EndFor

    \State $\mathcal{N}^{(k)}
    \gets\mathcal{A}^{(k)}\cup\mathcal{V}^{(k)}$\Comment{New neighborhood set}
    \State $\mathcal{D}^{(k+1)}
    \gets\mathcal{D}^{(k)}\cup\mathcal{N}^{(k)}$\Comment{Updated training set}
    \State $\theta^{(k+1)}
    \gets\mathrm{TrainT2I}(\theta^{(k)},\mathcal{D}^{(k+1)})$

\EndFor

\end{algorithmic}
\end{algorithm}

\subsection{Dataset} 

Lunara's mix training corpus brings together public-domain images, open-source, permissively licensed (Apache 2.0/MiT), synthetic data, proprietary artwork and photographs and iterative human-reviewed variations. The proprietary artwork and photographs were developed in collaboration with affiliated practicing artists and photographers, incorporating their perspective on what makes an image artistically successful. Key public domain and open-source data portions include PD12M \cite{meyer2024publicdomain12mhighly}, permissive WikiMedia Commons \footnote{\url{https://commons.wikimedia.org/wiki/Main_Page}}, CMA dataset \footnote{\url{https://huggingface.co/datasets/links-ads/cma-dataset?}} and a mix of generated images by Qwen-Image, Z-Image Turbo \cite{2025zimageturbo}, and SDXL \cite{podell2023sdxl}. The training corpus was further expanded through iterative targeted sampling and selective injection of human-created art, each followed by generation of corresponding semantic variations. The variations expose Lunara to multiple controlled modifications of the same underlying content. This leads to Lunara learning to respond to changes in the conditioning signal while retaining the shared semantic structure. Compared with isolated image--prompt pairs, anchor-linked variations increase control and local coverage of the training distribution.

\section{Automated Evaluation}

We benchmark Lunara against seven image-generation models spanning frontier, open-weight, and efficiency-oriented ones to evaluate content integrity and artistic performance. We also compare Lunara with 16 image-generation models on GenEval to assess semantic and compositional fidelity. 

\subsection{Evaluation setup}
Using a suite of 8,000 images generated by 8 models (1000 each), we evaluate three dimensions of artistic intelligence: Content Integrity, Aesthetic Quality, and Emotional Resonance. The suite comprises 1,000 prompts: 500 spanning 14 artistic styles and 500 open-ended art-generation prompts, balancing stylistic coverage with broader creative variation. All prompts were generated by GPT-5.6 Sol without human intervention.

We then use GPT-5.6 Sol as an evaluator to assess the content, aesthetic, and emotional resonance properties. The evaluated models include Lunara, GPT-Image-1 mini, Qwen Image, AuraFlow \cite{auraflow2024v03}, SD 3.5 Turbo \cite{sd35largeturbo}, HiDream-I1 Fast \cite{2025hidreami1}, FLUX-Klein-4B \cite{flux2klein4b2025}, and Z-Image-Turbo. Each model generates 1,000 images at 1024×1024 resolution, using default generation parameters where applicable. All models are evaluated on the same prompt under blinded model identity, with the evaluation prompt provided in Appendix \ref{app:evaluation_prompt}. 

We also evaluate the same eight models with conventional automated metrics. Image–text alignment is measured using CLIP cosine similarity with the ViT-L/14 backbone \cite{radford2021learning}, providing an embedding-based measure of prompt–image semantic alignment. Visual aesthetics are assessed using the LAION aesthetics predictor, which maps normalized CLIP ViT-L/14 image embeddings to predicted human aesthetic preference. These metrics provide conventional measures of semantic alignment and visual appeal, but offer limited coverage of higher-order artistic and affective properties.

Finally, we use GenEval to measure compositional and attribute-level correctness, testing whether models reliably realize specified objects, counts, and colors. Sixteen leading image-generation models are included in the analysis: Lunara, Nano Banana 2.0, Seedream-4.0, GPT Image 2, GPT Image 1 [High], FLUX.2 [Dev], Qwen-Image, HiDream-O1-Image \cite{hidream2026o1image}, Z-Image-Turbo, Stable Diffusion 3.5 Large, Stable Diffusion 3 Medium \cite{sd3medium}, PixArt-$\alpha$ \cite{pixart}, Show-o \cite{showo}, Emu3-Gen \cite{emu3gen}, JanusFlow \cite{janusflow}, FLUX.1 [Dev], Janus-Pro-7B \cite{januspro}. 


\begin{table*}[]
\centering
\caption{
Overall model performance across 1,000 prompts.
Scores and win rates are reported independently for Aesthetic Quality, Emotion,
Resonance, and Content Integrity. CLIPScore and LAION Aesthetic are reported as
automatic evaluation metrics. Best results are shown in \textbf{bold},
and second-best results are \underline{underlined}.
}
\label{tab:overall_independent}

\small
\setlength{\tabcolsep}{5pt}
\renewcommand{\arraystretch}{1.05}

\begin{tabular}{lcccccc||cc}
\toprule
& \multicolumn{2}{c}{Aesthetic Quality}
& \multicolumn{2}{c}{Emotional Resonance}
& \multicolumn{2}{c}{Content Integrity}
& \multicolumn{2}{c}{Automatic Metrics} \\
\cmidrule(lr){2-3}
\cmidrule(lr){4-5}
\cmidrule(lr){6-7}
\cmidrule(lr){8-9}

Model
& Score $\uparrow$ & Win Rate
& Score $\uparrow$ & Win Rate
& Score $\uparrow$ & Win Rate
& CLIP $\uparrow$ & LAION $\uparrow$ \\
& & & & & & & Score & Aesthetic\\
\midrule

Moonworks Lunara
& \textbf{8.473} & \textbf{23.3}\%
& \underline{8.119} & \underline{17.5}\%
& 7.944 & 4.7\%
& \textbf{0.792} & \underline{6.721} \\

GPT 1 Image Mini
& \underline{8.457} & \underline{18.9}\%
& \textbf{8.363} & \textbf{31.7}\%
& \textbf{8.609} & \underline{28.8}\%
& 0.777 & 6.661 \\

Qwen Image
& 8.366 & 15.9\%
& 7.979 & 10.7\%
& \underline{8.345} & \textbf{29.6}\%
& \underline{0.789} & 6.561 \\

AuraFlow
& 8.313 & \underline{18.9}\%
& 8.011 & 15.9\%
& 6.662 & 1.1\%
& 0.781 & \textbf{6.745} \\

SD 3.5 Turbo
& 8.081 & 11.4\%
& 7.949 & 17.4\%
& 7.238 & 3.2\%
& 0.743 & 6.625 \\

HiDream-I1 Fast
& 8.035 & 5.1\%
& 7.568 & 3.5\%
& 7.794 & 9.2\%
& 0.764 & 6.710 \\

FLUX-Klein-4B
& 8.023 & 5.4\%
& 7.579 & 3.2\%
& 7.438 & 9.0\%
& 0.773 & 6.706 \\

Z-Image-Turbo
& 7.230 & 1.2\%
& 6.721 & 0.7\%
& 7.875 & 17.8\%
& 0.719 & 6.033 \\

\bottomrule
\end{tabular}
\end{table*}

\subsection{GPT-5.6 Sol Evaluation}

Lunara leads in Aesthetic Quality, achieving the highest mean score (8.473) and the highest win rate (23.3\%). GPT-Image-1-Mini follows closely in mean aesthetic score (8.457), with Qwen-Image third (8.366), while GPT-Image-1-Mini and AuraFlow share the second-highest win rate at 18.9\%. Lunara also performs strongly in Emotional Resonance, achieving a mean score of 8.119 and a win rate of 17.5\%, placing it among the leading models in the evaluation. GPT-Image-1-Mini records the highest mean Emotional Resonance score at 8.363. In Content Integrity, Lunara achieves a mean score of 7.944, remaining competitive with the frontier models. GPT-Image-1-Mini leads in mean Content Integrity with 8.609, followed by Qwen-Image at 8.345, while Qwen-Image records the highest win rate at 29.6\%. Table ~\ref{tab:overall_independent} and Figure~\ref{fig:Baseline_performance} show overall performance across evaluated models. 

Taken together, Lunara emerges as one of the most balanced generalists across the artistic intelligence frontier, while many competing models exhibit more specialized capability profiles. Notably, Lunara achieves this breadth with sub-10B active parameters and fast inference. Yet comparable scale does not imply comparable capabilities. AuraFlow, also at sub-10B scale, approaches Qwen-Image (20B) in Aesthetic Quality and Emotional Resonance while ranking very low in Content Integrity (6.662). Its strong artistic scores are consistent with its additional aesthetic-oriented models, but the divergence suggests a broader decoupling between visual appeal and content control, where aesthetically coherent or emotionally compelling images can still violate prompt constraints. Table ~\ref{tab:style_breakdown} breaks down performance across styles, regions, and general art. 

\subsection{Conventional Text-Image Alignment and Aesthetic Preference}
On conventional automated benchmarks, Lunara exhibits a balanced profile across content control and aesthetic preference. Lunara achieves the highest CLIPscore (0.792), followed by Qwen-Image (0.789) and AuraFlow (0.781), indicating strong prompt–image correspondence under a standard embedding-based measure. Lunara also achieves the second-highest LAION Aesthetics score (6.721), narrowly behind AuraFlow (6.745) and ahead of HiDream-I1-Fast (6.710) and FLUX.2-Klein-4B (6.706). Table ~\ref{tab:overall_independent} shows baseline models performance in ClipScore and LAION Aesthetic predictor.

\begin{table*}[]
\centering
\caption{
Performance across style categories for \emph{Aesthetic Quality},
\emph{Emotional Resonance}, and \emph{Content Integrity}.
Best results within each category are shown in \textbf{bold}, and
second-best results are \underline{underlined}.
}
\label{tab:style_breakdown}

\small
\setlength{\tabcolsep}{2.6pt}
\renewcommand{\arraystretch}{1.05}

\begin{tabular}{lcccccccc}
\toprule
Style
& GPT1mini & Lunara & Qwen & AuraFlow
& SD3.5 & HiDream-I1 & FLUX-Klein-4 & Z-Image-T \\
\midrule

\multicolumn{9}{c}{\textbf{Aesthetic Quality}} \\
\midrule

Arabic Portrait
& \underline{8.383}
& 7.961
& 8.128
& \textbf{8.528}
& 7.825
& 7.725
& 7.722
& 6.722 \\

Arabic Painting
& \textbf{8.464}
& 8.161
& 8.342
& \underline{8.411}
& 7.869
& 8.011
& 8.161
& 7.025 \\

NodiBangla
& \underline{8.267}
& \textbf{8.736}
& 7.953
& 7.975
& 7.397
& 7.750
& 7.758
& 7.153 \\

Digital Art
& 8.419
& \textbf{8.650}
& \underline{8.553}
& 8.389
& 8.164
& 8.108
& 8.278
& 7.283 \\

Chinese Oil Painting
& \underline{8.380}
& \textbf{8.540}
& 8.066
& 8.194
& 8.243
& 7.940
& 8.123
& 7.814 \\

Chinese Sketch
& 7.847
& \underline{8.486}
& 7.989
& 7.208
& \textbf{8.703}
& 7.958
& 7.819
& 7.694 \\

Korean Digital Art
& 8.178
& \underline{8.561}
& \textbf{8.686}
& 8.181
& 8.350
& 7.975
& 8.228
& 7.228 \\

Korean Portrait
& 8.103
& \underline{8.314}
& 7.811
& \textbf{8.377}
& 8.151
& 7.474
& 7.591
& 6.943 \\

Korean Traditional Art
& 8.017
& \textbf{8.557}
& 7.851
& 8.094
& 8.020
& 7.811
& \underline{8.140}
& 7.183 \\

Mixed Media Art
& 8.158
& 8.422
& 8.042
& \textbf{8.589}
& \underline{8.475}
& 7.619
& 8.072
& 7.189 \\

Nordic Aurora
& \textbf{8.814}
& 8.636
& \underline{8.644}
& 8.250
& 8.164
& 7.972
& 8.408
& 7.697 \\

Nordic Folk Art
& 8.411
& \textbf{8.706}
& \underline{8.422}
& 8.097
& 7.769
& 8.361
& 8.303
& 7.039 \\

Nordic Painting
& \textbf{8.656}
& 8.414
& \underline{8.581}
& 8.147
& 8.183
& 7.861
& 7.819
& 7.650 \\

Stamp Art
& 8.329
& \textbf{8.609}
& 8.331
& 7.857
& \underline{8.509}
& 8.137
& 8.089
& 7.060 \\

General Art
& \textbf{8.596}
& 8.463
& \underline{8.487}
& 8.461
& 8.033
& 8.162
& 8.008
& 7.197 \\

\midrule
\multicolumn{9}{c}{\textbf{Emotion Resonance}} \\
\midrule

Arabic Portrait
& \textbf{8.175}
& 7.397
& 7.697
& \underline{8.003}
& 7.697
& 7.342
& 7.189
& 6.186 \\

Arabic Painting
& \textbf{8.497}
& 7.736
& 7.894
& \underline{7.939}
& 7.656
& 7.411
& 7.522
& 6.367 \\

NodiBangla
& \underline{8.069}
& \textbf{8.289}
& 7.592
& 7.622
& 7.206
& 7.181
& 7.397
& 6.581 \\

Digital Art
& \textbf{8.522}
& \underline{8.400}
& 8.217
& 8.139
& 8.036
& 7.722
& 7.889
& 6.744 \\

Chinese Oil Painting
& \textbf{8.457}
& \underline{8.200}
& 7.669
& 7.877
& 8.094
& 7.440
& 7.651
& 7.363 \\

Chinese Sketch
& 7.658
& \underline{8.289}
& 7.467
& 6.619
& \textbf{8.664}
& 7.447
& 7.219
& 7.297 \\

Korean Digital Art
& 8.100
& 8.175
& \textbf{8.303}
& 7.956
& \underline{8.239}
& 7.533
& 7.808
& 6.692 \\

Korean Portrait
& 7.751
& \textbf{7.869}
& 7.200
& \underline{7.851}
& 7.774
& 6.811
& 7.071
& 6.357 \\

Korean Traditional Art
& 7.603
& \textbf{8.083}
& 7.431
& 7.566
& \underline{7.826}
& 7.194
& 7.583
& 6.649 \\

Mixed Media Art
& 7.875
& 8.000
& 7.514
& \underline{8.133}
& \textbf{8.292}
& 6.989
& 7.647
& 6.506 \\

Nordic Aurora
& \textbf{8.761}
& 8.444
& \underline{8.539}
& 8.042
& 8.200
& 7.739
& 8.122
& 7.481 \\

Nordic Folk Art
& 7.853
& \textbf{8.547}
& \underline{7.914}
& 7.483
& 7.708
& 7.858
& 7.831
& 6.531 \\

Nordic Painting
& \textbf{8.764}
& 8.086
& \underline{8.281}
& 7.789
& 8.167
& 7.417
& 7.381
& 7.386 \\

Stamp Art
& 7.617
& \underline{8.206}
& 7.626
& 7.417
& \textbf{8.260}
& 7.386
& 7.471
& 6.320 \\

General Art
& \textbf{8.602}
& 8.116
& 8.145
& \underline{8.277}
& 7.911
& 7.743
& 7.601
& 6.695 \\

\midrule
\multicolumn{9}{c}{\textbf{Content Integrity}} \\
\midrule

Arabic Portrait
& \textbf{8.797}
& 7.892
& \underline{8.631}
& 7.144
& 7.253
& 8.047
& 7.869
& 8.228 \\

Arabic Painting
& \textbf{8.306}
& 7.917
& \underline{8.044}
& 7.164
& 7.344
& 7.675
& 7.781
& 7.631 \\

NodiBangla
& \textbf{8.519}
& 7.733
& \underline{8.125}
& 6.236
& 6.767
& 7.639
& 7.750
& 7.986 \\

Digital Art
& \textbf{8.583}
& 7.900
& \underline{8.283}
& 7.369
& 7.203
& 7.906
& 7.369
& 7.036 \\

Chinese Oil Painting
& \textbf{8.491}
& 7.900
& 7.954
& 6.931
& 7.360
& 7.520
& 6.909
& \underline{8.260} \\

Chinese Sketch
& \textbf{8.558}
& 8.389
& \underline{8.442}
& 6.592
& 7.728
& 7.883
& 7.803
& 8.317 \\

Korean Digital Art
& \underline{8.481}
& 8.133
& \textbf{8.739}
& 6.511
& 7.275
& 8.053
& 7.353
& 7.214 \\

Korean Portrait
& \textbf{8.837}
& 7.871
& \underline{8.343}
& 6.646
& 7.526
& 7.686
& 7.574
& 8.077 \\

Korean Traditional Art
& \textbf{8.657}
& 7.900
& \underline{8.180}
& 6.449
& 6.571
& 7.334
& 7.569
& 7.686 \\

Mixed Media Art
& \textbf{8.617}
& 8.025
& \underline{8.581}
& 7.719
& 7.422
& 7.861
& 8.092
& 8.100 \\

Nordic Aurora
& \textbf{8.742}
& 8.219
& \underline{8.567}
& 7.097
& 7.667
& 7.836
& 7.481
& 8.197 \\

Nordic Folk Art
& \textbf{8.869}
& 8.281
& \underline{8.583}
& 6.483
& 7.208
& 7.944
& 7.906
& 7.675 \\

Nordic Painting
& \textbf{8.758}
& 8.169
& 8.050
& 7.236
& 8.156
& 7.592
& 7.533
& \underline{8.333} \\

Stamp Art
& \textbf{9.000}
& 7.743
& \underline{7.977}
& 6.174
& 7.431
& 7.906
& 7.817
& 7.694 \\

General Art
& \textbf{8.560}
& 7.881
& \underline{8.367}
& 6.483
& 7.124
& 7.809
& 7.245
& 7.861 \\

\bottomrule
\end{tabular}

\end{table*}


\subsection{GenEval image-generation evaluation}
We compare Lunara with sixteen leading image-generation models on the original GenEval benchmark for object-level prompt adherence. As shown in Table~\ref{tab:model-evaluation}, Lunara achieves perfect single-object accuracy, a color score of 0.92, nearly matching GPT-Image-2 at 0.93, and an overall score of 0.90, matching Z-Image-Turbo. It also matches GPT Image 1 [High] on two-object and color accuracy and outperforms SD3.5 Large across all measures. These results support that Lunara’s artistic gains do not come at the expense of content control, preserving strong Content Integrity while extending Aesthetic Quality and Emotional Resonance.


\begin{table}[htbp]
    \centering
    \caption{Image generation model evaluation results. Lunara achieves a color
    score just behind GPT Image 2 and Seedream-4.0. Dense ranks are shown in
    parentheses; lower ranks are better.}
    \label{tab:model-evaluation}
    \begin{tabular}{lccccc}
        \toprule
        \textbf{Model}
        & \textbf{Single-Obj}
        & \textbf{Two-Obj (\%)}
        & \textbf{Count}
        & \textbf{Color}
        & \textbf{Avg} \\
        \midrule
        Nano Banana 2.0
        & 1.00 (1)
        & 0.96 (3)
        & 0.71 (8)
        & 0.84 (6)
        & 0.88 (6) \\

        Seedream-4.0
        & 1.00 (1)
        & 0.92 (5)
        & 0.71 (8)
        & 0.93 (1)
        & 0.89 (5) \\

        GPT Image 1 [High]
        & 0.99 (2)
        & 0.92 (5)
        & 0.85 (2)
        & 0.92 (2)
        & 0.92 (3) \\

        GPT Image 2
        & 0.99 (2)
        & 0.98 (2)
        & 0.85 (2)
        & 0.93 (1)
        & 0.94 (1) \\

        PixArt
        & 0.98 (3)
        & 0.50 (12)
        & 0.44 (13)
        & 0.80 (10)
        & 0.68 (13) \\

        Show-o
        & 0.95 (5)
        & 0.52 (11)
        & 0.49 (11)
        & 0.82 (8)
        & 0.70 (12) \\

        Emu3-Gen
        & 0.98 (3)
        & 0.71 (9)
        & 0.34 (14)
        & 0.81 (9)
        & 0.71 (11) \\

        SD3-Medium
        & 0.98 (3)
        & 0.74 (8)
        & 0.63 (9)
        & 0.67 (12)
        & 0.76 (10) \\

        JanusFlow
        & 0.97 (4)
        & 0.59 (10)
        & 0.45 (12)
        & 0.83 (7)
        & 0.71 (11) \\

        FLUX.1 [Dev]
        & 0.98 (3)
        & 0.81 (7)
        & 0.74 (6)
        & 0.79 (11)
        & 0.83 (9) \\

        SD3.5 Large
        & 0.98 (3)
        & 0.89 (6)
        & 0.73 (7)
        & 0.83 (7)
        & 0.86 (7) \\

        \addlinespace

        Janus-Pro-7B
        & 0.99 (2)
        & 0.89 (6)
        & 0.59 (10)
        & 0.90 (3)
        & 0.84 (8) \\

        Z-Image-Turbo
        & 1.00 (1)
        & 0.95 (4)
        & 0.77 (4)
        & 0.89 (4)
        & 0.90 (4) \\

        FLUX.2 [Dev]
        & 1.00 (1)
        & 0.99 (1)
        & 0.79 (3)
        & 0.93 (1)
        & 0.93 (2) \\

        Qwen-Image
        & 0.99 (2)
        & 0.92 (5)
        & 0.89 (1)
        & 0.88 (5)
        & 0.92 (3) \\

        HiDream-O1-Image
        & 1.00 (1)
        & 0.99 (1)
        & 0.79 (3)
        & 0.89 (4)
        & 0.92 (3) \\

        \textbf{Lunara}
        & \textbf{1.00 (1)}
        & 0.92 (5)
        & 0.76 (5)
        & \textbf{0.92 (2)}
        & \textbf{0.90 (4)} \\
        \bottomrule
    \end{tabular}
\end{table}

%

\section{Human Preference}

To further assess the dimensions of artistic intelligence, we conduct a blinded human-preference study using the same 1,000 prompts and evaluation dimensions as the GPT 5.6 Sol evaluation (Appendix~\ref{app:evaluation_prompt}).

\subsection{Evaluation Setup}

 Six evaluators assess anonymized, randomly ordered image pairs comparing Lunara against the same seven baseline models. Lunara generates one image for each of the 1,000 evaluation prompts, while comparison prompts are randomly assigned across baseline models and evaluators without replacement.

For each image pair, evaluators assess \emph{Content Integrity}, \emph{Aesthetic Quality}, and \emph{Emotional Resonance}. Each measure is rated on a 1--10 scale for both images, followed by a forced-choice preference between the pair. Left--right image positions are randomized to mitigate presentation-order effects. Ratings and pairwise preferences are then aggregated to estimate relative performance and preference strength across models.

\begin{table*}[]
\centering
\caption{
Average human-evaluation scores for Aesthetic Quality, Emotional Resonance, and Content Integrity. Lunara is fixed in every pairwise comparison, with each prompt randomly assigned to a comparison model as the opponent. Scores are reported on a 10-point scale. The best result in each category is shown in \textbf{bold}, and the
second-best result is \underline{underlined}.
}
\label{tab:dimension_scores}

\small
\setlength{\tabcolsep}{10pt}
\renewcommand{\arraystretch}{1.1}

\begin{tabular}{lcccc}
\toprule
Model
& Evaluations
& Aesthetic Quality $\uparrow$
& Emotional Resonance $\uparrow$
& Content Integrity $\uparrow$ \\
\midrule

Moonworks Lunara
& 1000
& \textbf{7.70}
& \textbf{7.71}
& \textbf{7.89} \\

HiDream-I1 Fast
& 143
& \underline{7.34}
& 7.22
& 7.70 \\

AuraFlow
& 143
& 7.31
& \underline{7.26}
& 7.10 \\

GPT 1 Image Mini
& 143
& 7.14
& 7.22
& \underline{7.71} \\

Qwen Image
& 142
& 7.10
& 7.01
& 7.70 \\

SD 3.5 Turbo
& 143
& 7.08
& 6.94
& 7.17 \\

FLUX-Klein-4B
& 143
& 6.99
& 6.93
& 7.31 \\

Z-Image-Turbo
& 143
& 6.77
& 6.66
& 7.17 \\

\bottomrule
\end{tabular}
\end{table*}

\begin{figure*}[t]
    \centering
    \includegraphics[
        width=\textwidth
    ]{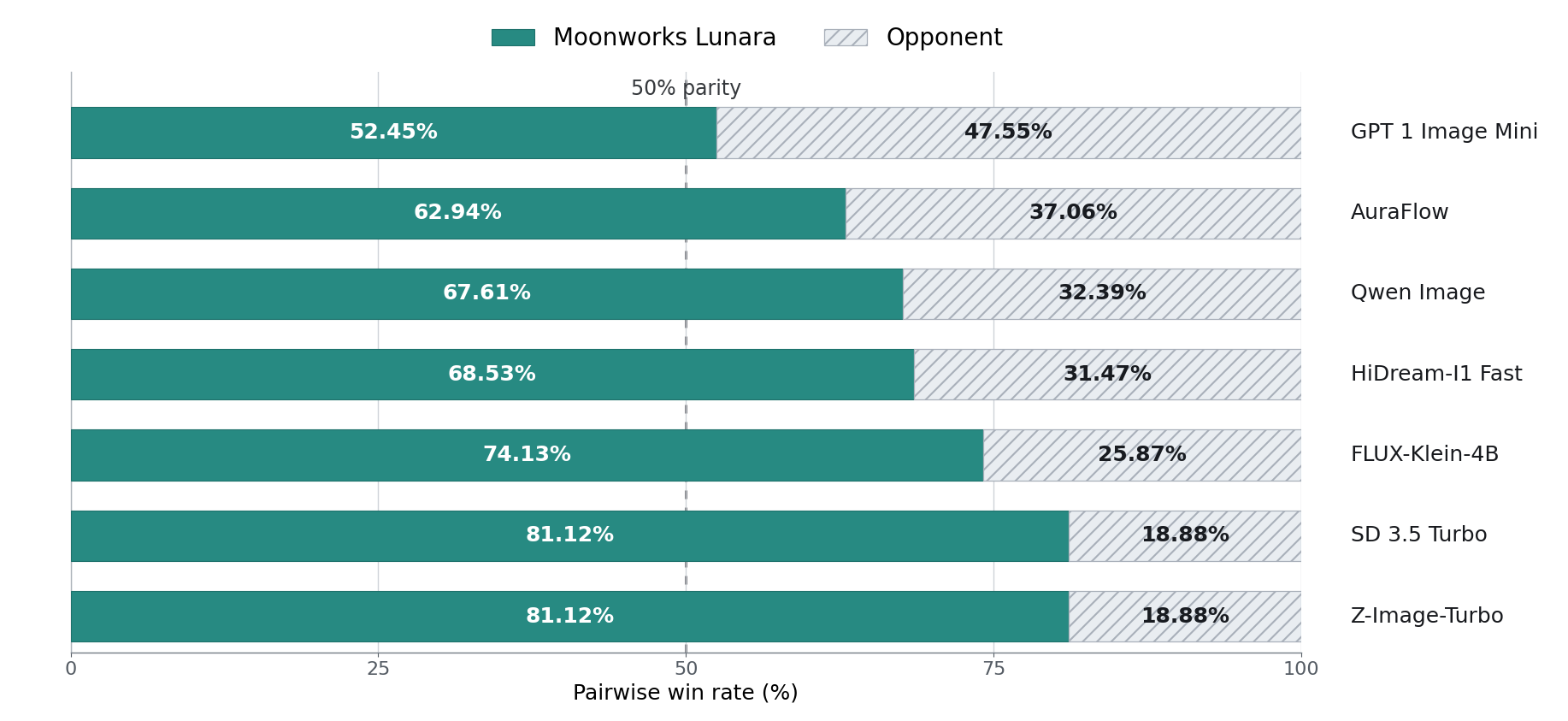}

    \caption{
    Derived head-to-head human preference win rates. Lunara outperforms all comparison models. Each opponent’s reported overall win rate is treated as its pairwise win rate against Lunara, with Lunara’s win rate calculated as the reminder to 100\%.
    }
    \label{fig:lunara_head_to_head}
\end{figure*}

\begin{figure*}[p]
    \centering

    \begin{minipage}{0.49\textwidth}
        \centering
        \includegraphics[height=0.245\textheight,keepaspectratio]{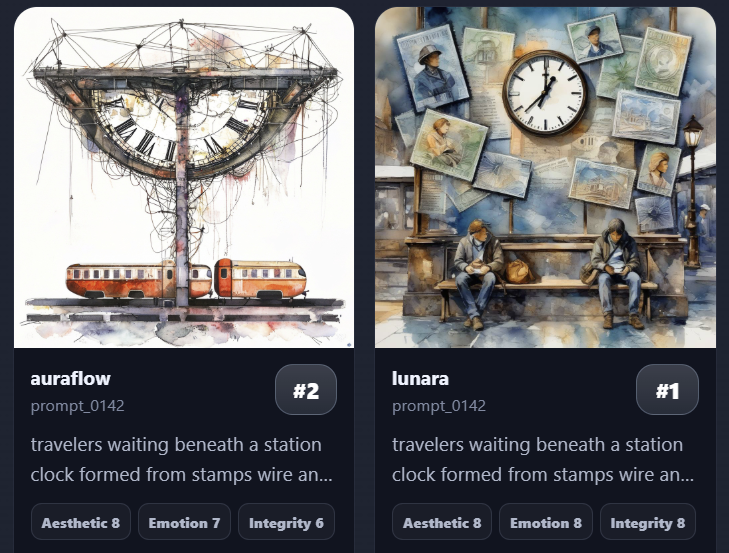}
    \end{minipage}
    \hfill
    \begin{minipage}{0.49\textwidth}
        \centering
        \includegraphics[height=0.245\textheight,keepaspectratio]{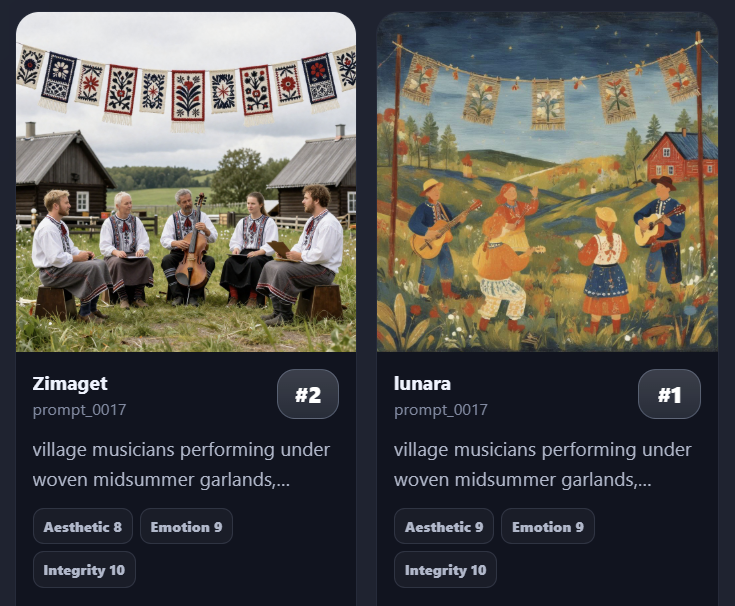}
    \end{minipage}

    \vspace{0.5em}

    \begin{minipage}{0.49\textwidth}
        \centering
        \includegraphics[height=0.245\textheight,keepaspectratio]{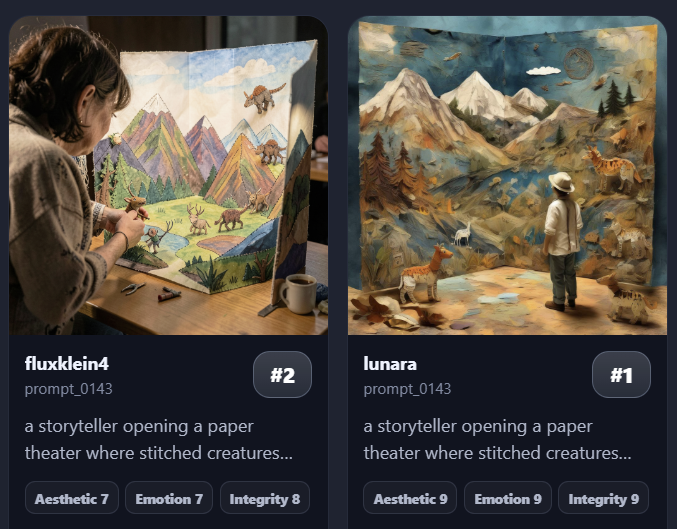}
    \end{minipage}
    \hfill
    \begin{minipage}{0.49\textwidth}
        \centering
        \includegraphics[height=0.245\textheight,keepaspectratio]{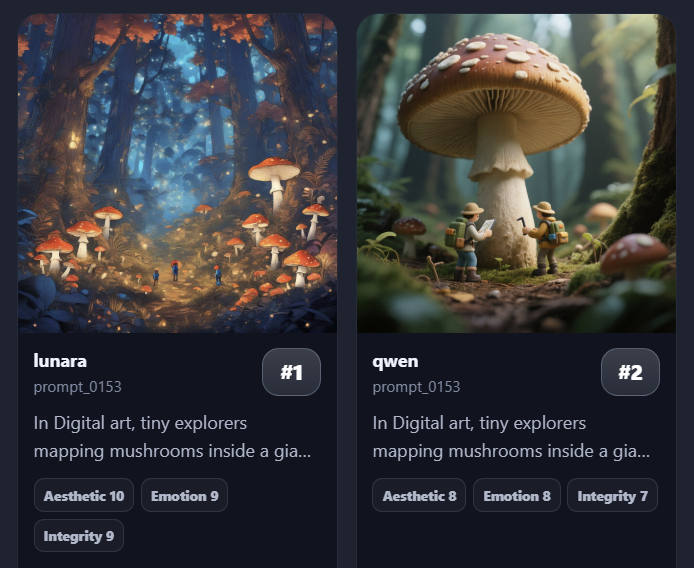}
    \end{minipage}

    \vspace{0.5em}

    \begin{minipage}{0.49\textwidth}
        \centering
        \includegraphics[height=0.245\textheight,keepaspectratio]{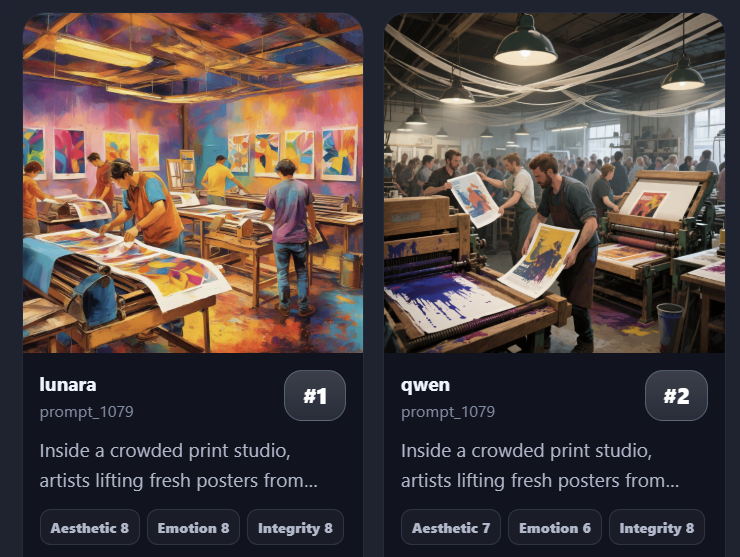}
    \end{minipage}
    \hfill
    \begin{minipage}{0.49\textwidth}
        \centering
        \includegraphics[height=0.245\textheight,keepaspectratio]{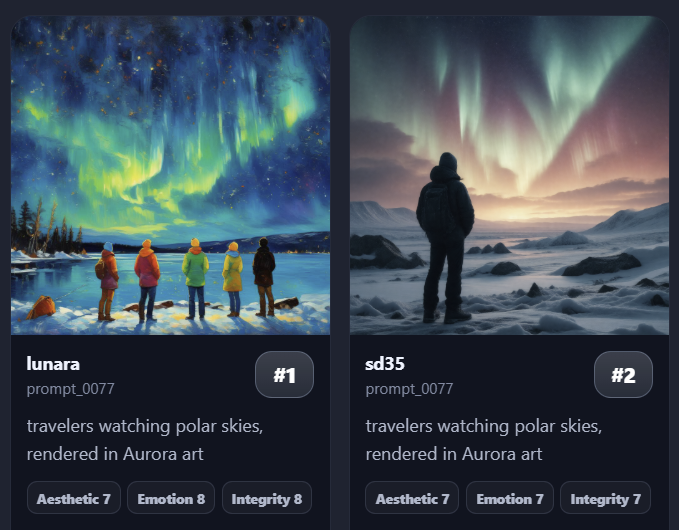}
    \end{minipage}

    \caption{Blind human evaluation examples. Names of the models were hidden during the evaluation.}
    \label{fig:human_evaluation_results}
\end{figure*}

Annotations are collected through Moonworks Nightmark, a general-purpose human-annotation platform that supports configurable evaluation interfaces, including scalar ratings, pairwise comparisons, and comparison of multiple model outputs. Nightmark is used to present blinded model generations and record structured evaluator judgments under a consistent evaluation prompt.     
Figure~\ref{fig:human_evaluation_results} shows representative anonymized image pairs evaluated by annotators on Nightmark.

\subsection{Human Validation of Artistic Intelligence}

Human preference broadly agrees with the automated results at the extremes, with Lunara consistently leading and Z-Image-Turbo remaining near the bottom,  but substantially reorders the middle of the field. Lunara achieves the highest human-rated \emph{Aesthetic Quality} (7.70) and \emph{Emotional Resonance} (7.71), leading on both dimensions across machine and human evaluation. Table~\ref{tab:dimension_scores} summarizes the results across all baseline models. HiDream-I1-Fast shows the largest evaluator shift, rising from sixth to second in Aesthetic Quality (7.34) and from near the bottom to a tie for third with GPT-Image-1-Mini in Emotional Resonance (7.22).  In contrast, GPT-Image-1-Mini and Qwen-Image decline slightly in relative standing under human judgment. These shifts expose a gap between machine scoring and human experience that defines a frontier for Artistic Intelligence. It is the difference between producing a successful image and constructing an expressive world with enough coherence and depth to be entered and experienced.

Lunara also leads human-rated \emph{Content Integrity} at 7.89, rising from third under automated evaluation to first under human judgment. HiDream-I1-Fast similarly moves up, while Z-Image-Turbo drops from fourth to near the bottom. This reordering suggests that fidelity is more than local constraint satisfaction. Human judgment is especially sensitive to whether the meaning of a request survives artistic transformation, while automated evaluators can still catch subtle attributes, secondary objects, or obscured content that people may miss \cite{hassan2025coherence}. 

Figure~\ref{fig:lunara_head_to_head} reports head-to-head win rates between Lunara and each baseline model. Lunara is preferred over every competing model, with the narrowest margin against GPT-Image-1-Mini and progressively larger advantages across the remaining models. Even against AuraFlow, Qwen-Image, and HiDream-I1-Fast, each of which is strong on particular human-rated dimensions, Lunara wins approximately 63--69\% of pairwise comparisons. This also reveals what aggregate benchmarks can miss. Models with similar measured capability can still produce worlds that feel fundamentally different to human evaluators. More representative examples from the human evaluation on Nightmark are shown in Appendix ~\ref{app:additional examples}. 

Across GPT 5.6 Sol evaluation, conventional benchmarks, GenEval, and blinded human judgment, Lunara consistently combines strong content control with high-order artistic performance. The results support the central premise of \emph{Artistic Intelligence}: preserving what must remain true while giving that meaning enough aesthetic and emotional force to become a coherent expressive world. Lunara reaches this balance with a compact model and fast inference, positioning Artistic Intelligence as a distinct axis of progress toward general visual intelligence.



	

\section*{Conclusion}

We formulate \emph{Artistic Intelligence} as a computational direction for image generation and implement it in Lunara, a sub-10B active-parameter model. Across GPT 5.6 Sol evaluation, conventional benchmarks, GenEval, and blind human evaluation, Lunara combines strong \emph{Content Integrity} with leading \emph{Aesthetic Quality} and \emph{Emotional Resonance}. Pairwise preference further reveals that models close on standard metrics can occupy different positions under human judgment.

Lunara reaches artistic intelligence through its novel Diffusion Mixture Transformer, iterative training algorithm, and informative data acquisition concentrating computation and learning on the semantic, artistic, and compositional constraints that govern generation. We hypothesize that, as this capability develops, new artistic styles may emerge with \emph{Artistic Intelligence} as a distinct axis of progress at the frontier of general visual intelligence. The broader implication is that progress in image generation is not exhausted by scale or correctness. Better architecture, algorithms, and compute allocation can open new dimensions of visual intelligence.


\bibliographystyle{acl}
\bibliography{references}

\appendix
\section{Evaluation Prompt}
\label{app:evaluation_prompt}
\begin{tcolorbox}[
  enhanced,
  breakable,
  colback=CriticBackground,
  colframe=CriticBlue,
  boxrule=0.8pt,
  arc=3mm,
  left=2mm,
  right=2mm,
  top=2mm,
  bottom=2mm,
  fonttitle=\large\bfseries,
  coltitle=white,
  colbacktitle=CriticBlue
]
\small
You are an expert art critic judging AI-generated images as visual
artworks. Compare all candidate images created from the same source
prompt side by side. The candidates are anonymous. Never infer or mention model or vendor
identities. Do not automatically reward any particular model-family
style. Judge primarily from an \textbf{artistic} point of view: consider what
is compelling, pleasing, expressive, and memorable to look at. Do not
reward technical cleanliness, sharpness, crisp edges, dense detail,
high resolution, or photorealism for their own sake. A painterly image
with expressive color, stylistic fusion, or reinterpretation may score
higher when these qualities strengthen the visual experience. Do not
assume any single interpretation of a style.

\medskip
Give each candidate a \textbf{relative score from 1.0 to 10.0} for
every criterion:

\begin{enumerate}[leftmargin=*, label=\textbf{\arabic*.}]
  \item \textbf{Aesthetic quality}

  Evaluate color relationships, vibrance, mood, atmospheric lighting,
  compositional flow and rhythm, expressive mark-making, and the
  coherent integration of materials and blended textures. Do not
  automatically reward formulaic, commercial, or hollow styling.
  Penalize malformed figures and unintended artifacts when they weaken
  the overall visual experience.

  \item \textbf{Emotional resonance}

  Evaluate how color, light, atmosphere, rhythm, space, gesture, and
  expressive marks create a convincing, multilayered emotional
  experience and leave a meaningful emotional aftereffect.

  \item \textbf{Content integrity}

  Evaluate proportional anatomy and the reasonable completeness of
  intended objects and their parts in relation to the shared prompt.
  The image should remain free from unintended artifacts, including
  malformed anatomy, watermarks, signatures, and structural
  inconsistencies.
\end{enumerate}

\medskip
\textbf{Calibration}

\begin{description}[leftmargin=2.7cm, style=multiline]
  \item[\textbf{9.0--10.0}] Exceptional performance on the criterion
  among these candidates.
  \item[\textbf{7.0--8.9}] Strong and convincing.
  \item[\textbf{5.0--6.9}] Mixed, ordinary, or moderately effective.
  \item[\textbf{3.0--4.9}] Weak.
  \item[\textbf{1.0--2.9}] Severely lacking.
\end{description}

Use the full scale when justified. Ties are allowed; do not force a
winner. Compare the candidates with one another while keeping the scale
meaningful across prompt sets. Judge only what is actually visible.

\medskip
\textbf{Output requirements}

Return exactly one score object for every supplied candidate ID, using
each candidate ID exactly as written. Every score object must contain
only the following fields:

\begin{quote}
\ttfamily
candidate ID\\
aesthetic quality\\
emotion resonance\\
content integrity
\end{quote}

Do not return a rationale or introduce additional scoring criteria.
After the score objects, return one concise comparative summary for the
full candidate set.

\end{tcolorbox}

\section{Additional examples from the human evaluation on Nightmark platform}
\label{app:additional examples}
\begin{figure*}[p]
    \centering

    \begin{minipage}{0.49\textwidth}
        \centering
        \includegraphics[height=0.245\textheight,keepaspectratio]{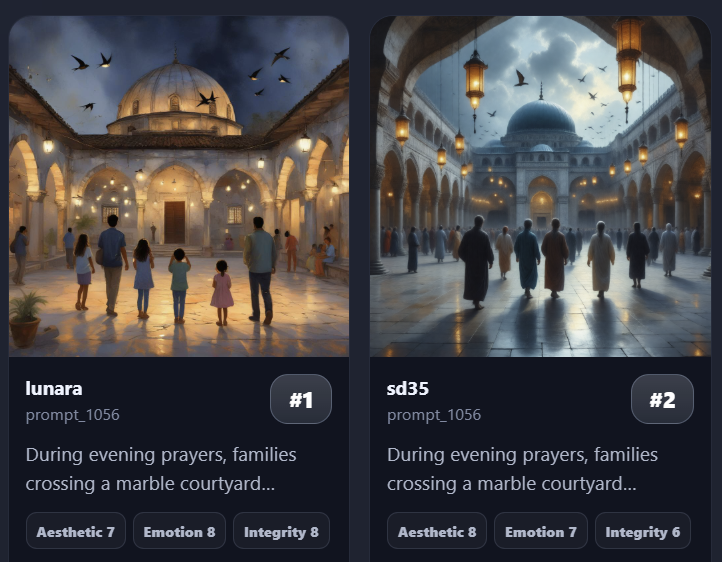}
    \end{minipage}
    \hfill
    \begin{minipage}{0.49\textwidth}
        \centering
        \includegraphics[height=0.245\textheight,keepaspectratio]{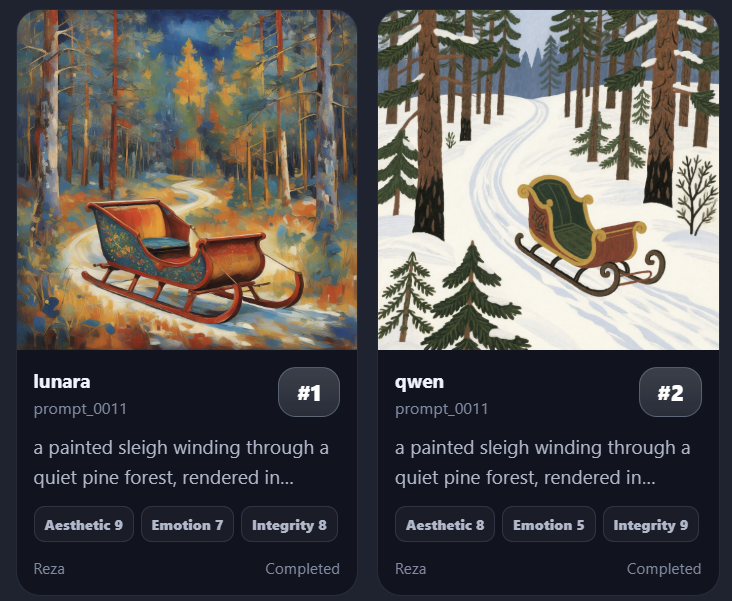}
    \end{minipage}

    \vspace{0.5em}

    \begin{minipage}{0.49\textwidth}
        \centering
        \includegraphics[height=0.245\textheight,keepaspectratio]{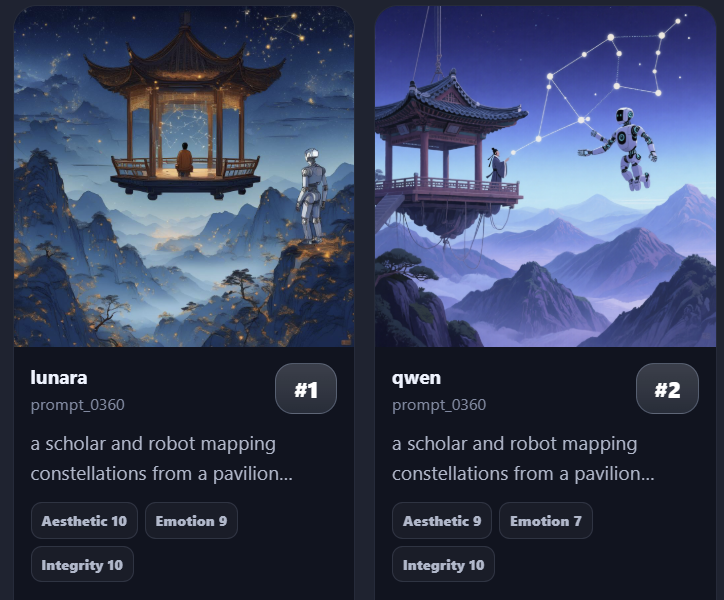}
    \end{minipage}
    \hfill
    \begin{minipage}{0.49\textwidth}
        \centering
        \includegraphics[height=0.245\textheight,keepaspectratio]{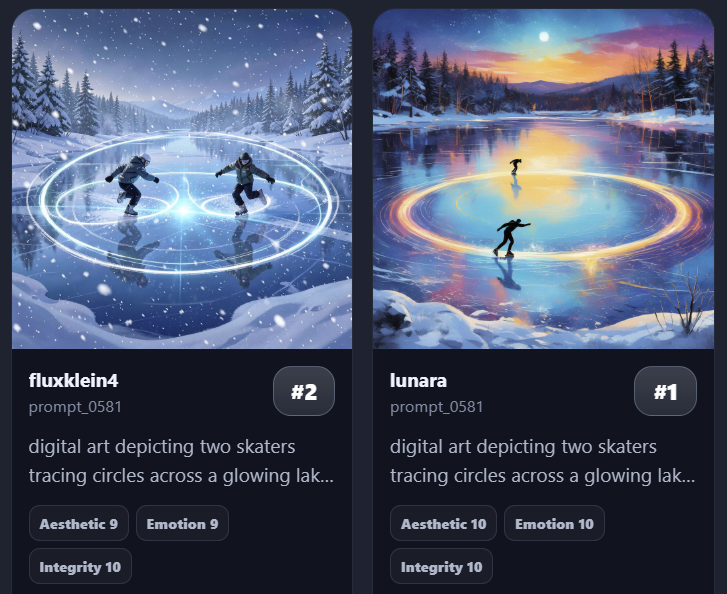}
    \end{minipage}

    \vspace{0.5em}

    \begin{minipage}{0.49\textwidth}
        \centering
        \includegraphics[height=0.245\textheight,keepaspectratio]{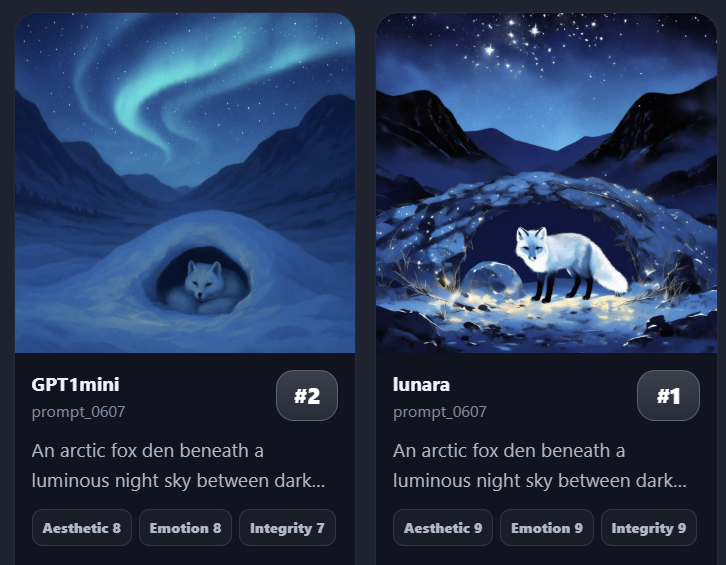}
    \end{minipage}
    \hfill
    \begin{minipage}{0.49\textwidth}
        \centering
        \includegraphics[height=0.245\textheight,keepaspectratio]{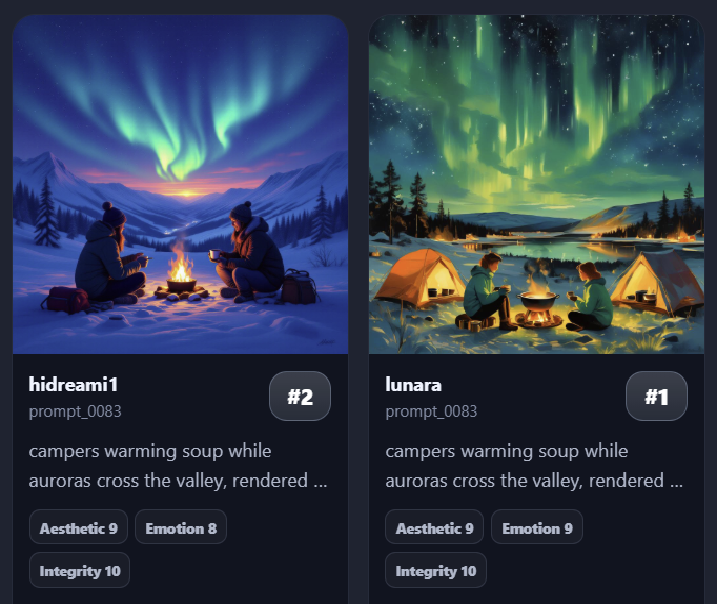}
    \end{minipage}

    \caption{Additional blinded human evaluation results. Names of the models were hidden during the evaluation.}
    \label{fig:human_evaluation_results_1}
\end{figure*}

\begin{figure*}[p]
    \centering

    \begin{minipage}{0.49\textwidth}
        \centering
        \includegraphics[height=0.245\textheight,keepaspectratio]{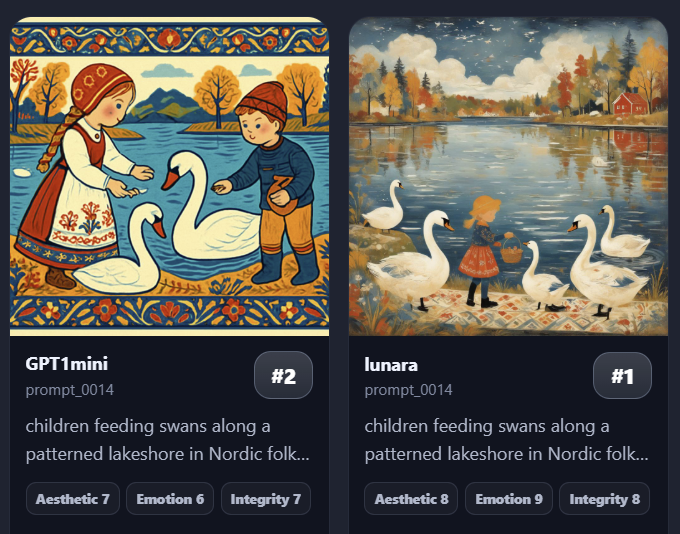}
    \end{minipage}
    \hfill
    \begin{minipage}{0.49\textwidth}
        \centering
        \includegraphics[height=0.245\textheight,keepaspectratio]{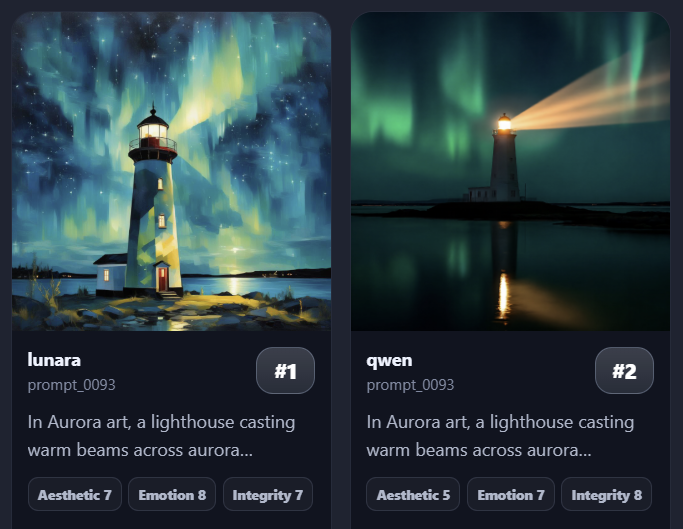}
    \end{minipage}

    \vspace{0.5em}

    \begin{minipage}{0.49\textwidth}
        \centering
        \includegraphics[height=0.245\textheight,keepaspectratio]{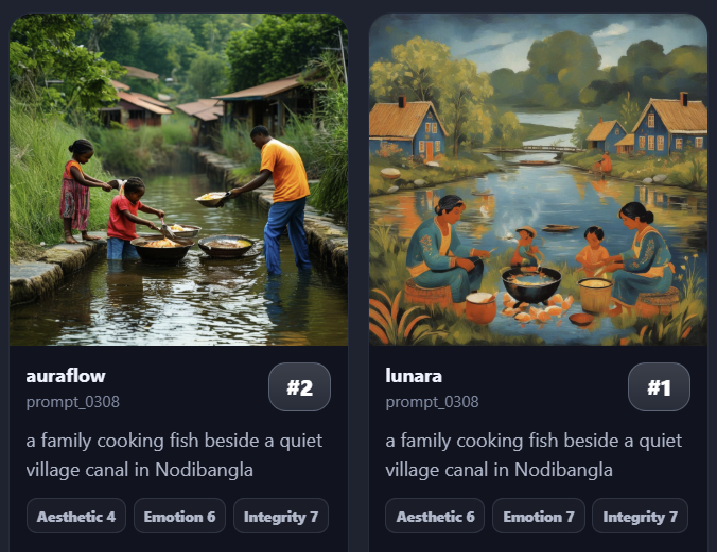}
    \end{minipage}
    \hfill
    \begin{minipage}{0.49\textwidth}
        \centering
        \includegraphics[height=0.245\textheight,keepaspectratio]{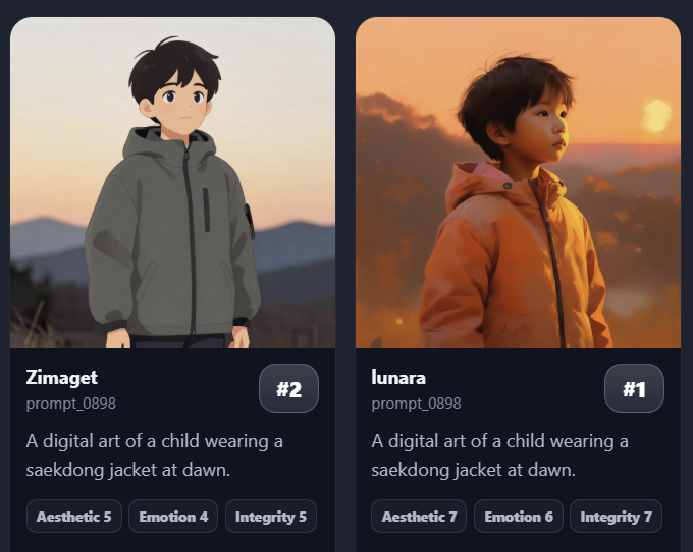}
    \end{minipage}

    \vspace{0.5em}

    \begin{minipage}{0.49\textwidth}
        \centering
        \includegraphics[height=0.245\textheight,keepaspectratio]{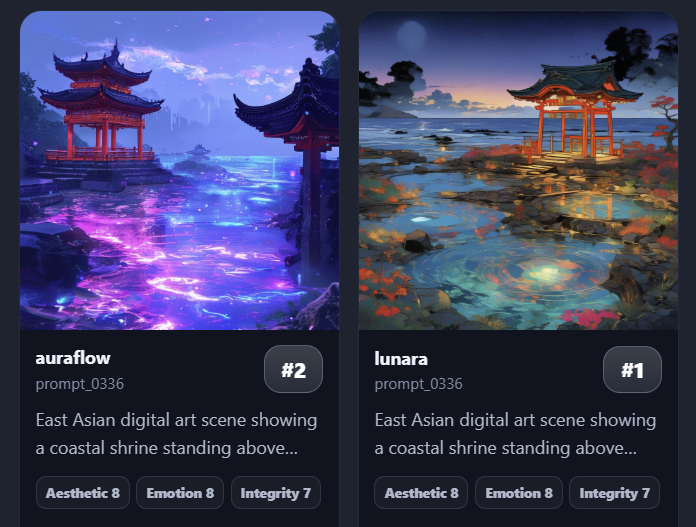}
    \end{minipage}
    \hfill
    \begin{minipage}{0.49\textwidth}
        \centering
        \includegraphics[height=0.245\textheight,keepaspectratio]{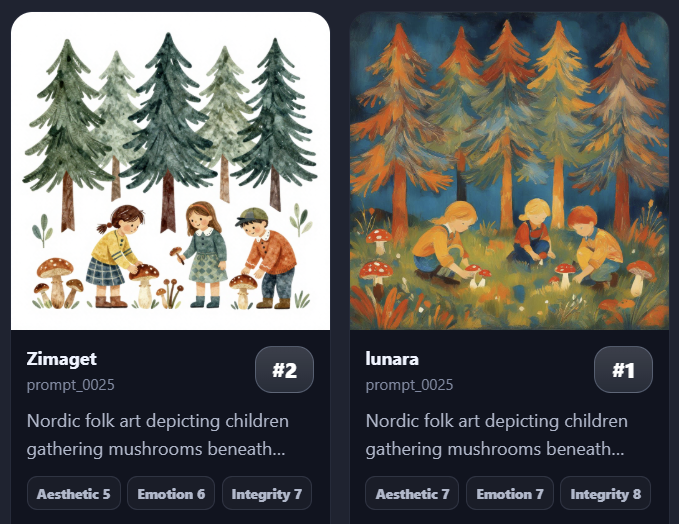}
    \end{minipage}

    \caption{Additional blinded human evaluation results. Names of the models were hidden during the evaluation.}
    \label{fig:human_evaluation_results_2}
\end{figure*}

\end{document}